\documentclass[letterpaper]{article} 
\usepackage{aaai2026}  
\usepackage{times}  
\usepackage{helvet}  
\usepackage{courier}  
\usepackage[hyphens]{url}  
\usepackage{graphicx} 
\usepackage{natbib}  
\usepackage{caption} 
\usepackage{algorithm}
\usepackage{algorithmic}
\usepackage{multirow}
\usepackage{amssymb}
\usepackage{amsmath}
\usepackage{amsthm}
\usepackage{booktabs}
\usepackage{enumitem}
\usepackage{graphicx}
\usepackage{color}
\usepackage{subcaption}
\usepackage{newfloat}
\usepackage{listings}
\usepackage{algorithm}
\usepackage{algorithmic}
\usepackage{tabularx}
\usepackage{xcolor}
\usepackage{float}
\usepackage{newfloat}
\usepackage{listings}
\DeclareCaptionStyle{ruled}{labelfont=normalfont,labelsep=colon,strut=off} 
\floatstyle{ruled}
\newfloat{listing}{tb}{lst}{}
\floatname{listing}{Listing}
\usepackage{mdframed}
\mdfdefinestyle{KeyTakeaway}{
    linecolor=black,
    outerlinewidth=1pt,
    roundcorner=10pt,
    innertopmargin=5pt,
    innerbottommargin=5pt,
    innerrightmargin=5pt,
    innerleftmargin=5pt,
}

\mdfdefinestyle{Box1}{style=KeyTakeaway}
\mdfdefinestyle{Box2}{style=KeyTakeaway}
\mdfdefinestyle{Box3}{style=KeyTakeaway}
\mdfdefinestyle{Box4}{style=KeyTakeaway}
\mdfdefinestyle{Box5}{style=KeyTakeaway}
\mdfdefinestyle{Box6}{style=KeyTakeaway}
\mdfdefinestyle{Box7}{style=KeyTakeaway}
\mdfdefinestyle{Box8}{style=KeyTakeaway}

\title{Language Models Coupled with Metacognition\\ Can Outperform Reasoning Models}
\author{
Vedant Khandelwal\textsuperscript{\rm 1,2}, Francesca Rossi\textsuperscript{\rm 1}, Keerthiram Murugesan\textsuperscript{\rm 1}, Erik Miehling\textsuperscript{\rm 1},\\ Murray Campbell\textsuperscript{\rm 1}, Karthikeyan Natesan Ramamurthy\textsuperscript{\rm 1}, Lior Horesh\textsuperscript{\rm 1} \\
}
\affiliations{
    \textsuperscript{\rm 1}IBM Research, NY, USA\\
    \textsuperscript{\rm 2}University of South Carolina, SC, USA\\

    vedant@email.sc.edu, francesca.rossi2@ibm.com, Keerthiram.Murugesan@ibm.com, Erik.Miehling@ibm.com, mcam@us.ibm.com, knatesa@us.ibm.com, lhoresh@us.ibm.com
}

\usepackage{bibentry}

\begin{document}

\maketitle

\begin{abstract}
Large language models (LLMs) excel in speed and adaptability across various reasoning tasks, but they often struggle when strict logic or constraint enforcement is required. In contrast, Large Reasoning Models (LRMs) are specifically designed for complex, step-by-step reasoning, although they come with significant computational costs and slower inference times. To address these trade-offs, we employ and generalize the SOFAI (Slow and Fast AI) cognitive architecture into SOFAI-LM, which coordinates a fast LLM with a slower but more powerful LRM through metacognition. The metacognitive module actively monitors the LLM's performance and provides targeted, iterative feedback with relevant examples. This enables the LLM to progressively refine its solutions without requiring the need for additional model fine-tuning. Extensive experiments on graph coloring and code debugging problems demonstrate that our feedback-driven approach significantly enhances the problem-solving capabilities of the LLM. In many instances, it achieves performance levels that match or even exceed those of standalone LRMs while requiring considerably less time. Additionally, when the LLM and feedback mechanism alone are insufficient, we engage the LRM by providing appropriate information collected during the LLM's feedback loop, tailored to the specific characteristics of the problem domain and leads to improved overall performance. Evaluations on two contrasting domains—graph coloring, requiring globally consistent solutions, and code debugging, demanding localized fixes—demonstrate that SOFAI-LM enables LLMs to match or outperform standalone LRMs in accuracy while maintaining significantly lower inference time. 
\end{abstract}

\section{Introduction}

Reasoning tasks, from classical constraint satisfaction problems to complex debugging scenarios in software development, continue to pose significant challenges for artificial intelligence (AI) systems. Large language models (LLMs) have shown remarkable flexibility and speed, quickly generalizing across a wide array of reasoning domains \cite{brown2020language,kojima2022large}. However, LLMs often struggle with tasks requiring strict logical consistency and hard constraint satisfaction \cite{jiang2023followbench,valmeekam2022large}. In contrast, large reasoning models (LRMs) offer robust step-by-step deliberative reasoning, but their effectiveness comes with substantially higher computational demands and slower inference times \cite{kambhampati2024can,stechly2024self}.

Reconciling the trade-off between speed and reliability remains an open research problem. To tackle this issue, we draw on previous work related to architectures inspired by the dual-process cognitive theory of human decision-making \cite{booch2021thinking,fabiano2023plan,ganapini2022combining,lin2024swiftsage}, specifically the SOFAI architecture \cite{cacmsofai}. SOFAI employs two types of problem solvers: "fast" solvers, which are generally quicker but less reliable, and "slow" solvers, which are more reliable but typically require more resources and time. Additionally, SOFAI incorporates a metacognitive component that selects the most appropriate solver for each problem instance, based on the risk profile and the available resources. This approach has been shown to achieve better overall performance than any single solver alone, particularly when the fast solver is an LLM-based planner and the slow solver is a classical symbolic planner \cite{cacmsofai}.


In this stduy, we define a generalized version of the SOFAI architecture, which we call SOFAI-LM, by introducing a training-free metacognitive feedback loop for LLMs. Specifically, we implement a feedback mechanism connecting the fast solver — an LLM in our case — with the metacognitive component.  This empowers the LLM to self-correct and improve its outputs iteratively, without any additional fine-tuning. If the LLM fails to converge within a fixed number of iterations, the metacognitive component invokes the slower, deliberative solver — an LRM in this case —  along with the accumulated feedback and a history of past solutions for a problem.

The paper makes the following key contributions:
\begin{itemize}[leftmargin=*]
\item We introduce a generalized version of the SOFAI architecture, called SOFAI-LM. This incorporates a training-free metacognitive governance module that features iterative S1 feedback and selective fallback.
\item Through extensive experiments in graph coloring and code debugging, we demonstrate that an LLM guided by the SOFAI-LM metacognitive feedback can match or even outperform a standalone LRM in both domains.
\item Additionally, we investigate the impact of the information collected in the LLM + Metacognition loop when calling the LRM. We characterize the domains where sharing this information enhances the performance of the LRM. 
\end{itemize}
Overall, our findings reveal that the architecture invokes the slower, compute-intensive LRM only when necessary, making use of the information accumulated during the LLM’s feedback loop. This approach is tailored to domain-specific characteristics, leading to significant improvements in the LRM's computational efficiency and inference time. To achieve these contributions, we address the following research questions:

\begin{mdframed}[style=Box1]
\begin{itemize}
\item RQ1: Can a feedback-driven LLM outperform an LRM? 
\item RQ2:  How do the type of feedback and the format of episodic memory affect a feedback-driven LLM?
\item RQ3: Can the information gathered by SOFAI-LM, when used iteratively with an LLM, enhance the performance of an LRM?
\item RQ4: Does SOFAI-LM perform better than its LRM counterpart?
\end{itemize}
\end{mdframed}

Our experimental results from the two problem domains indicate a promising directions to some of these research questions. Additionally, they demonstrate that increased feedback improves outcomes when iterating with an LLM.


\section{Background}

\subsection{The SOFAI architecture}

Kahneman’s influential work Thinking, Fast and Slow popularized the distinction between System 1 (fast, intuitive thinking) and System 2 (slow, analytical thinking), providing a conceptual lens that has shaped both cognitive psychology \cite{kahneman2011thinking}. In the realm of AI, \citep{booch2021thinking} introduced the SOFAI (Slow and Fast AI) architecture, which combines a fast solver and a slow solver under the governance of a metacognition.

The SOFAI architecture draws inspiration from dual-process theories of human cognition, which differentiate between fast, intuitive reasoning (System 1) and slow, deliberate reasoning (System 2) \cite{cacmsofai}. In this framework, System 1 solvers are fast and intuitive, like LLMs, that rapidly generate candidate solutions using learned heuristics. In contrast, System 2 solvers like LRM rely on slow, deliberate reasoning to ensure correctness and reliability. A key advantage in SOFAI is its metacognitive module, which monitors the performance of the solvers and dynamically selects the most suitable solver for a given problem. This selection process balances speed and accuracy, as metacognitive module considers solution quality, available resources, and past solver performance to adjust future decisions, thus promoting adaptability, reliability, and efficient resource use.

The original SOFAI architecture successfully demonstrated this hybrid strategy in applications such as automated planning and pathfinding \cite{booch2021thinking, fabiano2023plan, ganapini2022combining, lin2024swiftsage}, showing that alternating between neural intuition and symbolic rigor can yield efficiency gains without sacrificing correctness. However, previous SOFAI-like approaches typically treated S1 and S2 as independent solvers, simply choosing between them on a per-instance basis rather than enabling iterative collaboration.

\subsection{Problem domains}

Reasoning over structured and unstructured domains remains a cornerstone of AI, underpinning both theoretical advancements and practical applications. To rigorously assess the capabilities of reasoning systems, we focus on two representative and challenging problem domains: graph coloring decision problem and automated program debugging.

\paragraph{Graph Coloring Decision Problem.}

The graph coloring decision problem is a canonical constraint satisfaction problem (CSP) with broad relevance in scheduling, allocation, and resource management. Given an undirected, unweighted graph \( G = (V, E) \) and a positive integer \( k \), the objective is to determine whether there exists an assignment of at most \( k \) colors to the vertices such that no two adjacent nodes share the same color:
\begin{itemize}
    \item \textbf{Input:} An undirected, unweighted graph \( G = (V, E) \) and a positive integer \( k \).
    \item \textbf{Goal:} Decide if there exists a function \( f: V \rightarrow \{1, 2, \dots, k\} \) such that for every edge \( (u, v) \in E \), \( f(u) \neq f(v) \).
\end{itemize}
Unlike the optimization variant of the problem, which aims to find minimum number of colors, the decision version focuses on satisfiability for a fixed \( k \), directly reflecting global consistency requirements of many real-world problems.

We utilize the DIMACS representation~\cite{johnson1996cliques} to express graph coloring instances, which provides a standardized format for benchmarking and visualization. Notably, verifying a candidate solution in the graph coloring decision problem is computationally easy. However, generating a correct solution or rectifying an incorrect one requires reasoning over the entire graph's structure. \textit{Global} consistency is crucial, as each coloring choice may influence the overall feasibility of the solution.

\begin{figure}[ht]
    \centering
    \begin{lstlisting}[language=, basicstyle=\ttfamily\scriptsize\color{black!90}, frame=shadowbox, backgroundcolor=\color{gray!5}, keywordstyle=\bfseries\color{blue!60!black}, commentstyle=\itshape\color{green!40!black}, numbers=left, numberstyle=\scriptsize\color{gray!70}, breaklines=true]
    c Example of a graph coloring problem in DIMACS format
    p edge 5 5
    c edges
    e A B
    e A C
    e B C
    e C D
    e D E
    \end{lstlisting}
    \label{fig:dimacs_graph}
\end{figure}

\paragraph{Code Debugging.}

Code debugging is a fundamental yet time-consuming task in software engineering, accounting for up to half of the overall effort involved in development~\cite{mcconnell2004code}. While the advent of LLMs has led to significant advancements in code generation,  automated debugging—the ability to identify and repair faults in code—has remained an underexplored challenge until recently \cite{tian2024debugbench}. In this paper, we focus on the DebugBench, a large-scale benchmark designed to systematically evaluate the debugging capabilities of LLMs ~\cite{tian2024debugbench}. It comprises 4,253 instances that cover Python, Java, and C++. It includes a diverse taxonomy of bug types and programming scenarios. Each example in DebugBench is generated by deliberately introducing a bug into an otherwise correct code snippet. This ensures that only a \textit{local} region of the code is erroneous while the remainder of the code remains correct. This benchmark design necessitates precisely identifying and repairing erroneous segments without introducing new errors, distinguishing it from problems like graph coloring, where every color assignment must be checked against the global structure of the problem.


\paragraph{Domain Relevance and Motivation.}

We believe that the two problem domains of graph coloring and code debugging encompass a range of reasoning challenges that AI systems often encounter. Graph coloring requires holistic and globally consistent reasoning over combinatorial structures, while code debugging, as defined in DebugBench, assesses a model's ability to make precise and targeted corrections based on the local context. By evaluating both of these domains, we offer a comprehensive assessment of the ability to tackle diverse, high-impact, and practically relevant tasks.

\section{The SOFAI-LM Architecture}
\begin{figure*}[ht!]
    \centering
    \includegraphics[width=0.74\linewidth]{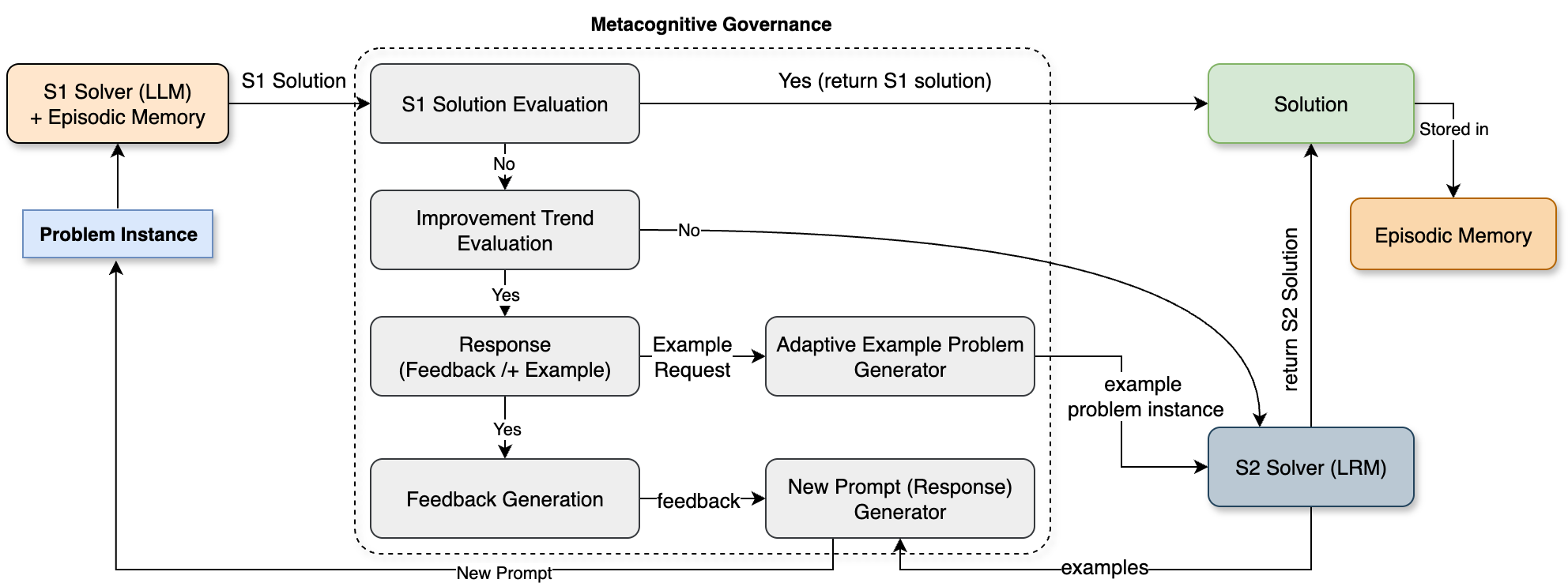}
    \caption{The SOFAI-LM architecture.}
    \label{fig:sofai-arch}
\end{figure*}
Our architecture builds upon the SOFAI architecture by incorporating a feedback loop between the System 1 solver and the metacognition component. In this paper, we focus on a specific instance of this architecture, in which the System 1 (S1) solver is a large language model (LLM) and the System 2 (S2) solver is a logical reasoning model (LRM). We refer to this instance as SOFAI-LM. Figure~\ref{fig:sofai-arch} provides an overview of this architecture.

\paragraph{System 1 solver: a Large Language Model.}
As mentioned earlier, the System 1 ($S_1$) solver in SOFAI-LM is an LLM, selected for its ability to rapidly generate initial candidate solutions for a given problem instance. Let $x$ denote the problem input and $M$ represent the episodic memory (a set of previously encountered examples and their solutions). The $S_1$ solver produces a candidate solution $y$ as follows:
\begin{equation}
S_1(x, M) \rightarrow y,
\end{equation}
where $y$ is the output generated by the LLM for input $x$.

\paragraph{Metacognitive Governance Module (MC).}
The metacognitive governance module (MC) of the architecture orchestrates the reasoning process by evaluating, guiding, and monitoring the outputs of the $S_1$ solver, and deciding when and if to invoke the $S_2$ solver. MC operates through the following sequential steps:

\begin{itemize}[leftmargin=*]
    \item \textbf{Evaluation:} After the $S_1$ solver generates a candidate solution $y$, MC evaluates its correctness using a problem-specific correctness function. For example, in graph coloring, the correctness function $C(y)$ can be defined as:
    \begin{equation}
    C(y) = \frac{\sum_{(u,v) \in E} \mathbb{I}[y(u)\neq y(v)]}{|E|},
    \end{equation}
    where $E$ is the set of edges in the graph, and $y(u)$ denotes the color assigned to vertex $u$.

    \medskip
\noindent\textbf{Code debugging (DebugBench).} 
For code-debugging tasks, we interface with the official \emph{LeetCode} API. 
Given a candidate patched code $y$, the API compiles and executes it on all hidden test cases and returns the pass ratio
\begin{equation}
    \operatorname{Pass}(y)=\frac{\text{\# tests passed by }y}{\text{\# total tests}}.
\end{equation}
MC deems $y$ correct when $\operatorname{Pass}(y)=1.0$.  
If $\operatorname{Pass}(y)<1.0$, the API also returns the last failing testcase 
$\langle \text{input},\text{expected output},\text{actual output}\rangle$. 
\medskip

    \item \textbf{Feedback Generation:} If the solution $y$ does not meet the required correctness threshold $\theta$, MC generates feedback $F(y)$ for $S_1$. This feedback may include information about specific errors, violated constraints, or example problems relevant to problem instance.
    \item \textbf{Iterative Refinement and Monitoring:} MC then iteratively refine the candidate solution by incorporating the feedback and re-invoking the $S_1$ solver:
    \begin{equation}
    y_{t+1} = S_1(x, M \cup F(y_t)), \quad t = 0, 1, \ldots, T-1,
    \end{equation}
    Where $T$ is the maximum number of allowed iterations. During this process, MC continuously monitors improvement in solution quality, e.g., by checking whether $C(y_{t+1}) > C(y_t)$ at each iteration.
    \item \textbf{Solver Selection:} If the candidate solution fails to achieve the correctness threshold after $T$ iterations, or if no further improvement is observed, MC invokes the System 2 ($S_2$) solver for final deliberation.
\end{itemize}

\paragraph{System 2 solver: Large Reasoning Model.}
The System 2 ($S_2$) solver is instantiated as a large reasoning model (LRM) chosen for its logical inference capabilities, usually stronger than those of LLMs. When invoked by MC, the $S_2$ solver receives the original problem instance $x$ as input, and may also receive the final unsuccessful attempt $y_T$ or the whole feedback history, depending on the implementation. The $S_2$ solver then produces the final solution $y^*$:
\begin{equation}
S_2(x, y_T, H) \rightarrow y^*,
\end{equation}
where $H$ denotes the available history of attempts and feedback, and $y^*$ is the output of the LRM.

\paragraph{Generalist and Specific modules in SOFAI-LM.}
The architecture’s feedback loop and metacognitive monitoring require no additional training or fine-tuning for the language models employed, enabling straightforward adaptation to diverse reasoning tasks such as graph coloring and code debugging. On the other hand, the evaluation and feedback generation modules are domain-specific, since they need to check correctness (or other properties) and generate feedback for candidate solutions in specific problem domains.

\paragraph{Workflow Overview.}
The full SOFAI-LM workflow proceeds as follows:
\begin{enumerate}
    \item The $S_1$ solver (an LLM) receives a problem instance $x$ and generates an initial candidate solution $y$.
    \item MC evaluates $y$ using the correctness function $C(y)$.
    \item If $y$ does not meet the correctness threshold, MC generates feedback $F(y)$ and calls again the $S_1$ solver, monitoring progress for up to $T$ iterations.
    \item If no satisfactory solution is found after $T$ iterations, or if improvement stagnates, MC invokes the $S_2$ solver to produce the final solution $y^*$.
\end{enumerate}

\section{Experimental Setting}

To use SOFAI-LM to address considered research questions, we design a set of experiments spanning two domains and multiple architecture configurations. Our goal is to systematically analyze influence of feedback, episodic memory, and solver initialization on both performance and computational efficiency of employes' LLM and LRM, as detailed below\footnote{Additional domain-specific details and prompts examples are provided in the Supplementary Material accompanying this paper.}
.

\subsection{Language models}

\paragraph{System 1 (LLM).}
As mentioned above, the S1 solver in SOFAI-LM is an LLM. In particular, we use \textbf{Granite 3.3 8B} (without thinking) \cite{ibm_granite_33_8b_instruct} and, in separate experiments, also \textbf{Llama3.1} \cite{dubey2024llama} as S1.

\paragraph{System 2 (LRM).}
The S2 solver in SOFAI-LM is an LRM. In particular, we use \textbf{DeepSeek R1 8B} \cite{guo2025deepseek}, \textbf{Granite 3.3 8B} (with thinking) \cite{ibm_granite_33_8b_instruct}, and \textbf{Qwen 3 8b} \cite{yang2025qwen3}.

All models' inference work was conducted using the Ollama \cite{marcondes2025using}, with decoding parameters set to \texttt{seed=12345}, \texttt{temperature=0.0}, \texttt{top\_k=1}, and \texttt{top\_p=1.0}, yielding greedy sampling and improved output stability across runs. This standardized configuration facilitates more consistent comparisons between LLMs' and LRMs' performance under matched inference conditions.

\subsection{Evaluation Domains}

\begin{itemize}[leftmargin=*]
    \item \textbf{Graph Coloring Decision Problem}: For this domain, we generate both solvable and unsolvable instances of graphs ranging from $5$ to $25$ vertices. For each size, $100$ unique instances are created, with edge probabilities sampled uniformly from $[0.1, 0.9]$ to ensure a broad spectrum of structural complexity and constraint tightness. This enables evaluation of LLM and LRM performance under varying degrees of combinatorial difficulty and solution feasibility.
    
    \item \textbf{DebugBench}: We use the Python and C++ subsets of DebugBench~\cite{tian2024debugbench}, a benchmark designed to assess the practical debugging abilities of LLMs and LRMs on realistic program repair tasks. The Python subset consists of diverse bug types, including logic errors, API misuse, and syntax errors, spanning both single- and multi-function programs. The C++ subset emphasizes language-specific pitfalls such as pointer misuse, memory management bugs, and object lifetime errors, providing a rigorous test of model reasoning in complex, stateful codebases. Each DebugBench instance contains one localized bug, requiring the solver to identify and repair the fault while preserving the correctness of the unaffected code.
\end{itemize}

\subsection{Feedback and Episodic Memory Variants}

To analyze the impact of the metacognitive module’s feedback and memory on the LLMs and LRMs, we define and test the following variants:

\paragraph{Feedback Types}
\begin{itemize}[leftmargin=*]
    \item \textbf{Multi-Line Feedback (MLF):} Feedback containing all error information and violated constraints or failing test cases, presented in a structured, hierarchical, multi-sentence format. This organization aims to make each issue explicit and easier to parse for the LLM.
    
    \item \textbf{Single-Line Feedback (SLF):} The same content as MLF, but compressed into a single line, providing concise guidance while preserving all informational content.
\end{itemize}
\begin{figure*}[!ht]
\centering
\includegraphics[width=0.88\linewidth]{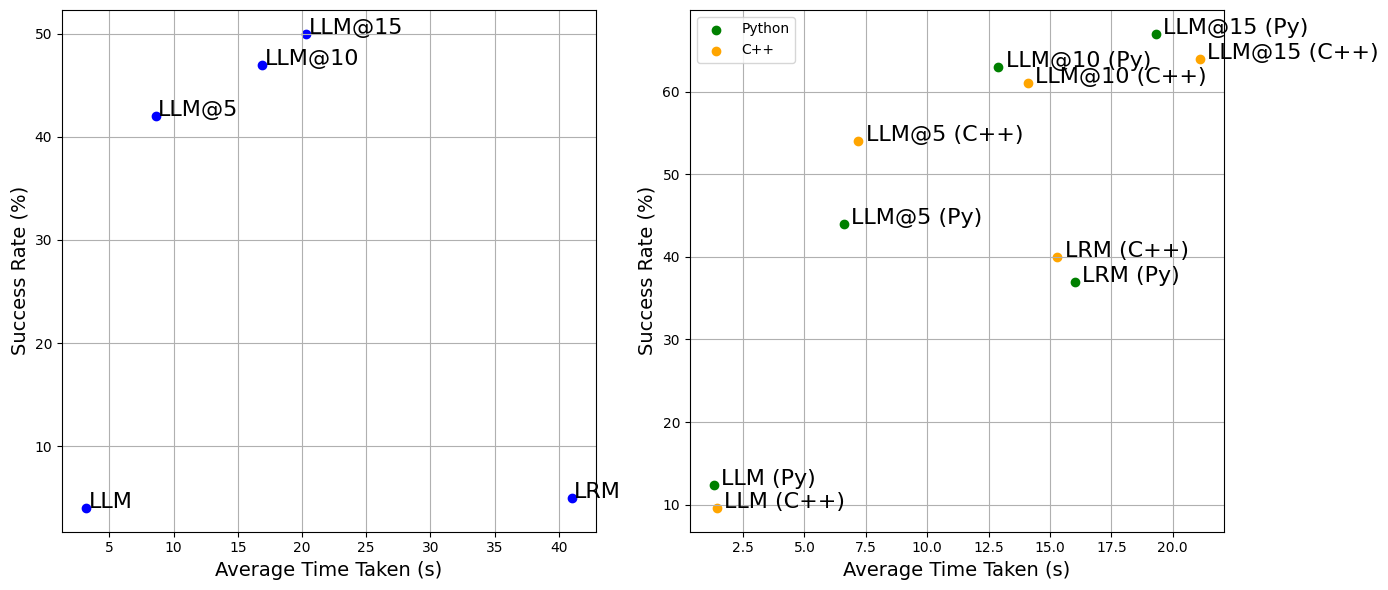}
\caption{
Each point corresponds to a configuration: \textbf{LLM}, \textbf{LLM@$5$}, \textbf{LLM@$10$}, \textbf{LLM@$15$}, and \textbf{LRM}.
Left: Graph coloring problems (Solvable, size = 25).
Right: Code debugging (Python and C++). The x-axis shows the average time per instance (in sec), while the y-axis shows the success rate (in \%). The LLM is Granite3.38b (without thinking); the LRM is Deepseek R18b.
}
\label{fig:rq1_scatter}
\end{figure*}

\paragraph{Episodic Memory Variants}
\begin{itemize}[leftmargin=*]
    \item \textbf{Minimal Episodic Memory (MEM):} Stores only pairs including a problem instance and its correct solution, with no historical context.
    \item \textbf{Extended Episodic Memory (EEM):} Stores the full sequence of LLM solution attempts and all feedback exchanges for each instance, along with the problem instances and their correct solutions.
\end{itemize}

\subsection{LRM prompting}

When the metacognitive governance module determines that the LLM cannot solve a problem after $T$ iterations, an LRM is invoked with one of the following prompts: 
\begin{itemize}[leftmargin=*]
    \item \textbf{Problem Instance-Only (PO):} the LRM receives only the original problem statement.
    \item \textbf{Best Attempt (BA):} LRM is provided with the best solution produced by LLM, along with the problem instance.
    \item \textbf{Full History (FH):} LRM is given entire history of LLM's attempts and MC feedback for the problem instance.
\end{itemize}


\section{Results}

Given the above experimental setting, we now describe the experiments we conducted to address the considered research questions and the results shown by the experiments. 


\vspace{0.5em}
\noindent {\bf RQ1: Can a feedback-driven LLM outperform an LRM?} 

To address this question, we evaluate whether the SOFAI-LM feedback loop enables an LLM (in this case, Granite3.38b without thinking) to match or surpass the performance of an LRM (in this case, DeepSeek R18B) across the two considered problem domains: graph coloring and code debugging. 

Figure \ref{fig:rq1_scatter} presents a comparative analysis of various configurations, noted as follows: LLM, LLM@$5$ (i.e, the LLM with $\le5$ iterations with feedback loop), LLM@$10$, LLM@$15$, and LRM. The figure shows that increasing the number of iterations leads to a consistent increase in solved problems, with the most significant improvements observed between LLM and LLM@$5$. While LLM@$15$ achieves the highest success rate in both domains, it does so with more compute time. Notably, the LRM, though faster on average, performs substantially worse on larger graph sizes in the graph coloring domain.
Overall, it is clear that iterating the use of the LLM dominates the LRM, since it can solve many more problem instances while using much less time\footnote{Results with other LLMs and LRMs are consistent and are included in the Supplementary Material accompanying this paper.\label{supp}}. 


\vspace{0.5em}
\noindent {\bf RQ2: How do the type of feedback and the format of episodic memory affect a feedback-driven LLM?} 

 \begin{figure}[H]
\centering
\includegraphics[width=0.9\linewidth]{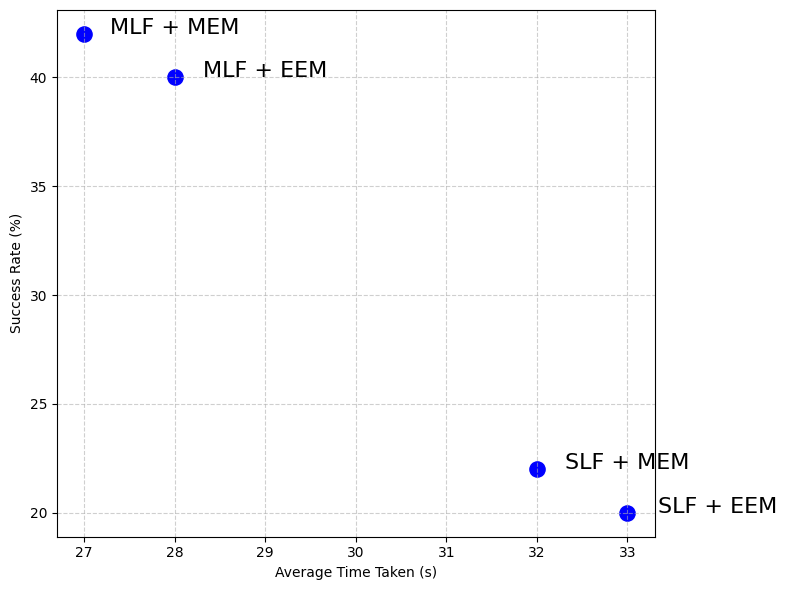}
\caption{
The x-axis shows the average time per instance (in sec), while the y-axis shows the success rate (in \%) for four metacognitive configurations, for the graph coloring domain (with graphs of size 25). Feedback types: Minimal Line Feedback (MLF) and Single Line Feedback (SLF), each paired with Minimal Episodic Memory (MEM) or Extended Episodic Memory (EEM). The LLM is Granite3.38b (without thinking) and the LRM is Deepseek R18b.}
\label{fig:rq2_gc_scatter}
\end{figure}

   \begin{figure*}[t]
  \centering
  \includegraphics[width=0.9\linewidth]{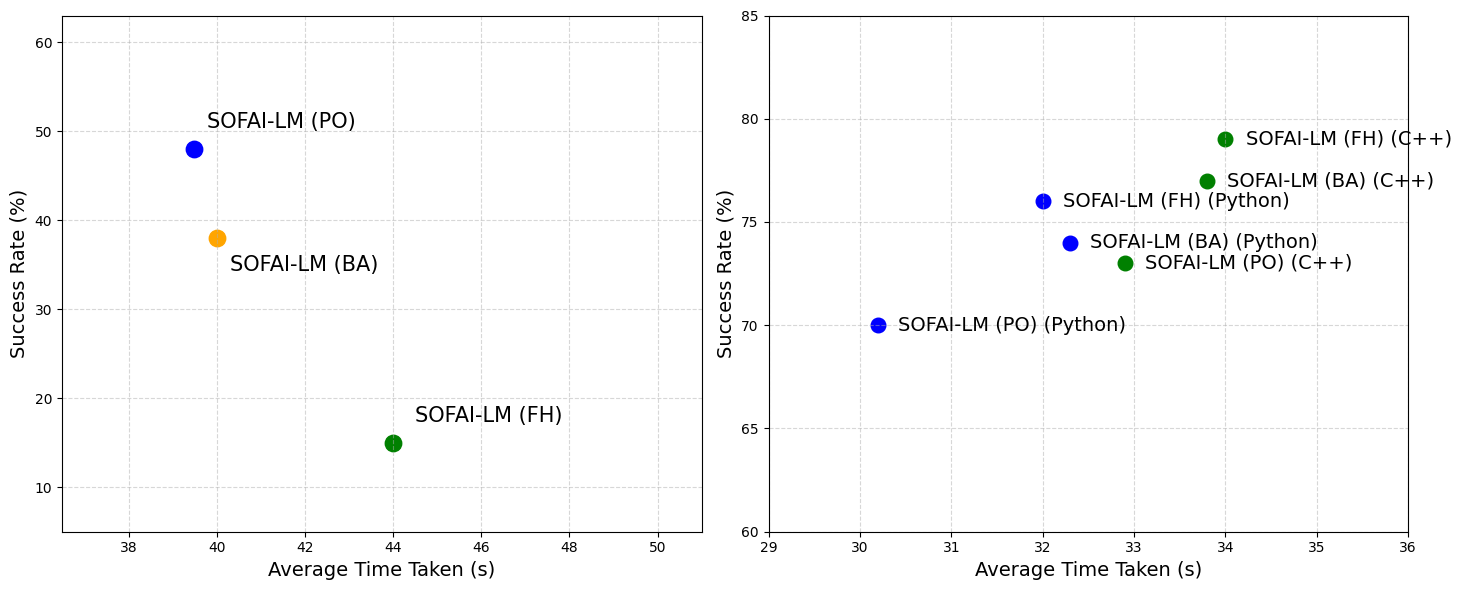}
  \caption{
  Left: graph coloring, with graph size 25.
  Right: code debugging (Python and C++). 
  The x-axis shows the average time per instance (in sec), while the y-axis shows the success rate (in \%) for SOFAI-LM using three LRM prompting strategies: PO (just the problem instance), BA (best attempt from LLM), and FH (full history of LLM@$15$ attempts). The LLM is Granite3.38b (without thinking) and the LRM is Deepseek R18b.
  }
  \label{fig:rq3_scatter}
\end{figure*}

To respond to this question, we conducted an experiment with four variants of the metacognitive loop in the graph coloring domain (graph size = 25): Minimal Line Feedback (MLF) and Single Line Feedback (SLF), each paired with either Extended Episodic Memory (EEM) or Minimal Episodic Memory (MEM).

Figure~\ref{fig:rq2_gc_scatter} shows the success rate versus time for these four configurations. We observe that using MLF consistently outperforms SLF, achieving a higher success rate with lower compute time across both episodic memory settings. Additionally, MEM yields better performance than its EEM counterpart, suggesting that limited context stored in the episodic memory improves the model's ability to focus on relevant corrections.
The best performing configuration is \textbf{MLF + MEM}, which achieves the highest success rate (42\%) at the lowest average time (~27s). In contrast, SLF + EEM lags both in success rate and time, indicating that both overly concise feedback and redundant historical context can hinder reasoning performance.
Summarizing, the combination of MLF with memory-limited feedback (MEM) offers the best trade-off between success rate and time. Because of this analysis, for the code debugging domain, we adopt the SLF + MEM setting across all experiments\footref{supp} . 


\vspace{0.5em}
\noindent {\bf RQ3: Can the information gathered by SOFAI-LM, when used iteratively with an LLM, enhance the performance of an LRM?} 

To investigate whether information generated during LLM iterations benefits the subsequent LRM solver, we compare three prompting strategies when calling the LRM: (1) PO (just problem instance), where LRM receives no prior context from LLM; (2) BA (Best Attempt), where LRM is called with most promising partial solution produced by LLM; and (3) FH (Full History), where LRM is provided full sequence of LLM’s attempts and MC feedback.

Figure~\ref{fig:rq3_scatter} shows the trade-off between success rate and time for all three LRM prompting strategies, for both the graph coloring domain (left) and the code debugging domain (right). 
In the graph coloring domain, using PO consistently yields the highest success rate and most favorable time efficiency, with the MLF+MEM configuration. While BA and FH can leverage additional context, they often introduce noise and degrade performance, especially in more complex memory and feedback settings. On the other hand, for the code debugging domain, results are the opposite: BA and FH show improvements in success rate over PO, albeit at the cost of increased time.
To justify this difference between the two domains, we conjecture that, in domains where fixing an incorrect solution requires only a local revision, such as in code debugging, providing negative examples is helpful. On the contrary, in domains where fixing an incorrect solution requires a possible revision of the whole solution, like in graph coloring, then it is not helpful, and actually detrimental, to provide negative examples and feedback on them\footref{supp} . 





\vspace{0.5em}
\noindent {\bf RQ4: Does SOFAI-LM perform better than its LRM counterpart?} 

To address this question, we compare the behavior of SOFAI-LM, where the LLM and metacognitive controller drive most reasoning and selectively invoke the LRM only when needed, versus using the LRM alone. We assess both the overall success rate and computation time per instance across the graph coloring (with graph size 25) and code debugging (both in Python and C++) domains.

    \begin{figure}[!ht]
  \centering
  \includegraphics[width=0.95\linewidth]{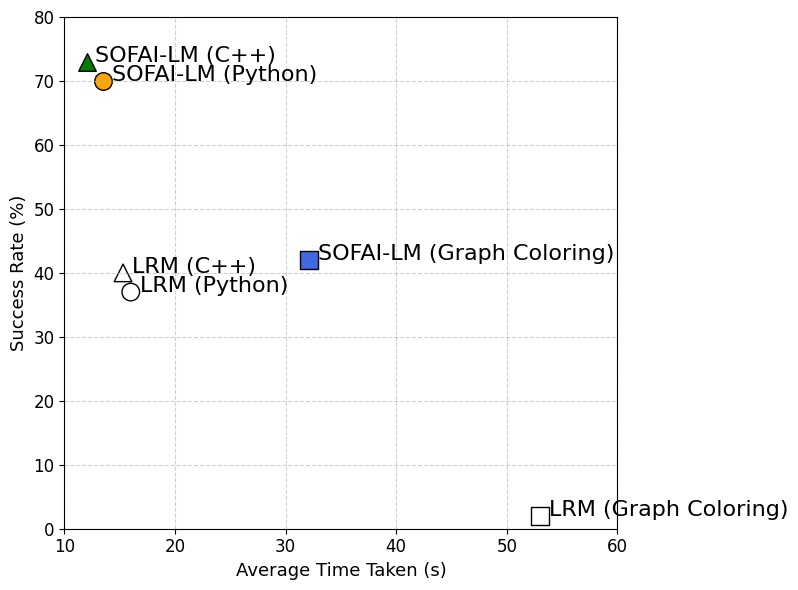}
  \caption{
  The x-axis shows the average time per instance (in sec), while the y-axis shows the success rate (in \%) for SOFAI-LM and LLM approaches across three settings: graph coloring (size 25), DebugBench-Python, and DebugBench-C++. Each point is marked by both domain and method. The LLM is Granite3.38b (without thinking) and the LRM is Deepseek R18b.}
  \label{fig:rq4_tradeoff}
\end{figure}

Figure~\ref{fig:rq4_tradeoff} presents the results for both domains. The SOFAI-LM pipeline consistently achieves a much higher success rate compared to the LRM in both domains, while also reducing the compute time, especially in the graph coloring domain. Notably, SOFAI-LM solves over 70\% of code debugging problems (Python: 70\%, C++: 73\%) in less time than the LRM, which solves only 37\% (Python) and 40\% (C++) with higher average runtimes. In the graph coloring domain, the advantage is even more pronounced: SOFAI-LM solves 42\% of the problems with a much lower time, whereas the LRM solves just 2\% of the problems and takes significantly longer. Therefore, selective fallback to LRM, as orchestrated by the SOFAI-LM pipeline, delivers a superior trade-off between success rate and efficiency compared to LRM reasoning alone, across both reasoning tasks\footref{supp}.


\section{Related Work}

\paragraph{LLM-based Reasoning and Constraint Satisfaction.}
LLMs have demonstrated emerging reasoning capabilities using prompting strategies such as chain-of-thought (CoT) reasoning \cite{Wei2022}, zero-shot reasoning \cite{kojima2022large}, scratchpads \cite{Nye2021}, and self-consistency \cite{Wang2022}. Extensions such as tool augmentation \cite{Schick2023toolformer,paranjape2023hindsight} and iterative self-refinement \cite{Shinn2023Reflexion,Madaan2023SelfRefine} enable models to tackle complex reasoning without explicit training. Iterative verification approaches, such as PiVe \cite{Han2023} and debate-based reasoning \cite{Li2023SelfConsistencyDebate}, further improve solution quality. However, empirical evaluations reveal persistent weaknesses in enforcing hard constraints and guaranteeing globally correct solutions, particularly in combinatorial tasks like graph coloring \cite{kambhampati2024can,Valmeekam2022,stechly2024self} and in program repair benchmarks like DebugBench \cite{tian2024debugbench}. These limitations motivate architectures that combine the adaptability of LLMs with the reliability of structured reasoning methods.


\paragraph{Fine-tuning and Verifier-Augmented Models.}
Several efforts aim to bridge the gap between LLMs and robust reasoning through additional training or verifier integration. Reinforcement learning–based fine-tuning, as in DeepSeek-R1 \cite{guo2025deepseek} and ThinkGPT \cite{Besta2023ThinkGPT}, biases models toward stepwise reasoning behaviors, while supervised alignment methods enhance code reasoning \cite{chen2021evaluating}. Verifier-augmented approaches explicitly incorporate symbolic or learned checkers \cite{Lightman2023LetsVerify,Paul2023ChainOfVerification}, improving accuracy on domains such as formal mathematics \cite{Polu2022Minerva} and theorem proving \cite{Welleck2022NaturalProofs}. Despite these gains, such models require substantial compute resources and often lose task generality, making them difficult to transfer between diverse problem classes such as graph-based CSPs and software debugging.

\paragraph{Neuro-Symbolic and Metacognitive Architectures.}
Neuro-symbolic architectures inspired by dual-process cognition \cite{kahneman2011thinking} combine fast heuristic reasoning (“System 1”) with slow deliberate reasoning (“System 2”). The SOFAI framework \cite{booch2021thinking} instantiated this paradigm in AI, improving planning and pathfinding performance via instance-level solver selection \cite{fabiano2023plan,ganapini2022combining,lin2024swiftsage}. Related work integrates symbolic solvers or verifiers into neural pipelines, e.g., differentiable reasoning modules \cite{Evans2021NeuralDeduction}, SAT/SMT solver integration \cite{Amizadeh2020NeuroSAT}, and theorem-proving hybrids \cite{Selsam2019NeuroSAT}. More recent approaches explore metacognition for LLMs, including self-evaluation and strategy adaptation \cite{Shinn2023Reflexion,Madaan2023SelfRefine}. However, these frameworks often rely on static solver selection or additional trained verifiers, limiting flexibility and adding engineering overhead. 

\section{Conclusions and Future Work}

This paper introduced SOFAI-LM, a training-free metacognitive architecture that couples the speed and flexibility of large language models (LLMs) with the robust reasoning capabilities of large reasoning models (LRMs). 
This work enables LLMs to iteratively refine their outputs through targeted feedback and, when necessary, selectively invoke LRMs. Across two distinct reasoning domains—global constraint satisfaction in graph coloring and localized program repair in code debugging—SOFAI-LM consistently matches or surpasses the performance of standalone LRMs while requiring significantly less computation.
A key advantage of SOFAI-LM is its model-agnostic design: any LLM can serve as the fast solver and any LRM as the slow solver, with domain-specific evaluation and feedback modules providing the only customization required. This flexibility makes SOFAI-LM applicable to a broad range of reasoning challenges beyond those explored here. Our experiments further show how differences in problem structure—global versus local fixes—affect the value of information sharing between the iterative LLM loop and LRM fallback, offering insights into how hybrid systems can be tailored to specific domains.
Future work will focus on automating and optimizing the metacognitive module itself. By learning policies for solver selection, iteration depth, and feedback and evaluation strategies, we aim to create a fully self-improving reasoning framework. Such a system could dynamically adapt across tasks, further reducing computational cost while improving reliability, bringing us closer to scalable, domain-general AI reasoning.

\bibliography{aaai2026}

\begin{thebibliography}{39}
\providecommand{\natexlab}[1]{#1}

\bibitem[{Amizadeh(2020)}]{Amizadeh2020NeuroSAT}
Amizadeh, S. e.~a. 2020.
\newblock NeuroSAT: End-to-End SAT Solver Learning.
\newblock \emph{ICLR}.

\bibitem[{Besta(2023)}]{Besta2023ThinkGPT}
Besta, M. e.~a. 2023.
\newblock ThinkGPT: Enhancing LLM Reasoning with Chain-of-Thought Fine-Tuning.
\newblock \emph{arXiv preprint arXiv:2309.02664}.

\bibitem[{Booch et~al.(2021)Booch, Fabiano, Horesh, Kate, Lenchner, Linck, Loreggia, Murugesan, Mattei, and Rossi}]{booch2021thinking}
Booch, G.; Fabiano, F.; Horesh, L.; Kate, K.; Lenchner, J.; Linck, N.; Loreggia, A.; Murugesan, K.; Mattei, N.; and Rossi, F. e.~a. 2021.
\newblock Thinking fast and slow in AI.
\newblock In \emph{Proceedings of the AAAI Conference on Artificial Intelligence}, volume~35, 15042--15046.

\bibitem[{Brown et~al.(2020)Brown, Mann, Ryder, Subbiah, Kaplan, Dhariwal, Neelakantan, Shyam, Sastry, and Askell}]{brown2020language}
Brown, T.~B.; Mann, B.; Ryder, N.; Subbiah, M.; Kaplan, J.; Dhariwal, P.; Neelakantan, A.; Shyam, P.; Sastry, G.; and Askell, A. e.~a. 2020.
\newblock Language Models are Few-Shot Learners.
\newblock \emph{arXiv preprint arXiv:2005.14165}.

\bibitem[{Chen et~al.(2021)Chen, Tworek, Jun, Yuan, Pinto, Kaplan, Edwards, Burda, Joseph, Brockman et~al.}]{chen2021evaluating}
Chen, M.; Tworek, J.; Jun, H.; Yuan, Q.; Pinto, H. P. D.~O.; Kaplan, J.; Edwards, H.; Burda, Y.; Joseph, N.; Brockman, G.; et~al. 2021.
\newblock Evaluating large language models trained on code.
\newblock \emph{arXiv preprint arXiv:2107.03374}.

\bibitem[{Dubey et~al.(2024)Dubey, Jauhri, Pandey, Kadian, Al-Dahle, Letman, Mathur, Schelten, Yang, and Fan}]{dubey2024llama}
Dubey, A.; Jauhri, A.; Pandey, A.; Kadian, A.; Al-Dahle, A.; Letman, A.; Mathur, A.; Schelten, A.; Yang, A.; and Fan, A. e.~a. 2024.
\newblock The Llama 3 Herd of Models.
\newblock \emph{CoRR}.

\bibitem[{Evans(2021)}]{Evans2021NeuralDeduction}
Evans, R. e.~a. 2021.
\newblock Making Neural Deduction Differentiable.
\newblock \emph{Nature Machine Intelligence}.

\bibitem[{Fabiano et~al.(2025)Fabiano, Ganapini, Loreggia, Mattei, Murugesan, Pallagani, Rossi, Srivastava, and Venable}]{cacmsofai}
Fabiano, F.; Ganapini, M.~B.; Loreggia, A.; Mattei, N.; Murugesan, K.; Pallagani, V.; Rossi, F.; Srivastava, B.; and Venable, K.~B. 2025.
\newblock Thinking Fast and Slow in Human and Machine Intelligence.
\newblock \emph{Commun. ACM}, 68(8): 72–79.

\bibitem[{Fabiano et~al.(2023)Fabiano, Pallagani, Ganapini, Horesh, Loreggia, Murugesan, Rossi, and Srivastava}]{fabiano2023plan}
Fabiano, F.; Pallagani, V.; Ganapini, M.~B.; Horesh, L.; Loreggia, A.; Murugesan, K.; Rossi, F.; and Srivastava, B. 2023.
\newblock Plan-SOFAI: A Neuro-Symbolic Planning Architecture.
\newblock In \emph{Neuro-Symbolic Learning and Reasoning in the era of Large Language Models}.

\bibitem[{Ganapini et~al.(2022)Ganapini, Campbell, Fabiano, Horesh, Lenchner, Loreggia, Mattei, Rossi, Srivastava, and Venable}]{ganapini2022combining}
Ganapini, M.~B.; Campbell, M.; Fabiano, F.; Horesh, L.; Lenchner, J.; Loreggia, A.; Mattei, N.; Rossi, F.; Srivastava, B.; and Venable, K. B. e.~a. 2022.
\newblock Combining Fast and Slow Thinking for Human-like and Efficient Decisions in Constrained Environments.
\newblock In \emph{NeSy}, 171--185.

\bibitem[{Guo et~al.(2025)Guo, Yang, Zhang, Song, Zhang, Xu, Zhu, Ma, Wang, and Bi}]{guo2025deepseek}
Guo, D.; Yang, D.; Zhang, H.; Song, J.; Zhang, R.; Xu, R.; Zhu, Q.; Ma, S.; Wang, P.; and Bi, X. e.~a. 2025.
\newblock Deepseek-r1: Incentivizing reasoning capability in llms via reinforcement learning.
\newblock \emph{arXiv preprint arXiv:2501.12948}.

\bibitem[{Han(2023)}]{Han2023}
Han, J. e.~a. 2023.
\newblock PiVe: Prompting with iterative verification improving graph-based generative capability of LLMs.
\newblock \emph{arXiv:2305.12392}.

\bibitem[{{IBM Research}(2025)}]{ibm_granite_33_8b_instruct}
{IBM Research}. 2025.
\newblock {Granite 3.3 8B Instruct Model Card}.
\newblock \url{https://huggingface.co/ibm-granite/granite-3.3-8b-instruct}.
\newblock Accessed on 2025-08-01.

\bibitem[{Jiang et~al.(2023)Jiang, Wang, Zeng, Zhong, Li, Mi, Shang, Jiang, Liu, and Wang}]{jiang2023followbench}
Jiang, Y.; Wang, Y.; Zeng, X.; Zhong, W.; Li, L.; Mi, F.; Shang, L.; Jiang, X.; Liu, Q.; and Wang, W. 2023.
\newblock FollowBench: A Multi-Level Fine-Grained Constraints Following Benchmark for Large Language Models.
\newblock \emph{arXiv preprint arXiv:2310.20410}.

\bibitem[{Johnson and Trick(1996)}]{johnson1996cliques}
Johnson, D.~S.; and Trick, M.~A. 1996.
\newblock \emph{Cliques, coloring, and satisfiability: second DIMACS implementation challenge, October 11-13, 1993}, volume~26.
\newblock American Mathematical Soc.

\bibitem[{Kahneman(2011)}]{kahneman2011thinking}
Kahneman, D. 2011.
\newblock Thinking, fast and slow.
\newblock \emph{Farrar, Straus and Giroux}.

\bibitem[{Kambhampati(2024)}]{kambhampati2024can}
Kambhampati, S. 2024.
\newblock Can large language models reason and plan?
\newblock \emph{Annals of the New York Academy of Sciences}, 1534(1): 15--18.

\bibitem[{Kojima et~al.(2022)Kojima, Gu, Reid, Matsuo, and Iwasawa}]{kojima2022large}
Kojima, T.; Gu, S.~S.; Reid, M.; Matsuo, Y.; and Iwasawa, Y. 2022.
\newblock Large language models are zero-shot reasoners.
\newblock \emph{Advances in neural information processing systems}, 35: 22199--22213.

\bibitem[{Li(2023)}]{Li2023SelfConsistencyDebate}
Li, Y. e.~a. 2023.
\newblock Improving Factuality via Multi-Agent Debate and Self-Consistency.
\newblock \emph{arXiv preprint arXiv:2305.14325}.

\bibitem[{Lightman(2023)}]{Lightman2023LetsVerify}
Lightman, A. e.~a. 2023.
\newblock Let's Verify Step by Step.
\newblock \emph{arXiv preprint arXiv:2305.20050}.

\bibitem[{Lin et~al.(2024)Lin, Fu, Yang, Brahman, Huang, Bhagavatula, Ammanabrolu, Choi, and Ren}]{lin2024swiftsage}
Lin, B.~Y.; Fu, Y.; Yang, K.; Brahman, F.; Huang, S.; Bhagavatula, C.; Ammanabrolu, P.; Choi, Y.; and Ren, X. 2024.
\newblock Swiftsage: A generative agent with fast and slow thinking for complex interactive tasks.
\newblock \emph{Advances in Neural Information Processing Systems}, 36.

\bibitem[{Madaan(2023)}]{Madaan2023SelfRefine}
Madaan, A. e.~a. 2023.
\newblock Self-Refine: Iterative Refinement with Self-Feedback.
\newblock \emph{arXiv preprint arXiv:2303.17651}.

\bibitem[{Marcondes et~al.(2025)Marcondes, Gala, Magalh{\~a}es, Perez~de Britto, Dur{\~a}es, and Novais}]{marcondes2025using}
Marcondes, F.~S.; Gala, A.; Magalh{\~a}es, R.; Perez~de Britto, F.; Dur{\~a}es, D.; and Novais, P. 2025.
\newblock Using ollama.
\newblock In \emph{Natural Language Analytics with Generative Large-Language Models: A Practical Approach with Ollama and Open-Source LLMs}, 23--35. Springer.

\bibitem[{McConnell(2004)}]{mcconnell2004code}
McConnell, S. 2004.
\newblock \emph{Code complete}.
\newblock Pearson Education.

\bibitem[{Nye(2021)}]{Nye2021}
Nye, M. e.~a. 2021.
\newblock Show your work: Scratchpads for intermediate computation with language models.
\newblock \emph{arXiv:2112.00114}.

\bibitem[{Paranjape and Chen(2023)}]{paranjape2023hindsight}
Paranjape, A.; and Chen, X. e.~a. 2023.
\newblock Hindsight Chain-of-Thought Reasoning.
\newblock In \emph{NeurIPS}.

\bibitem[{Paul(2023)}]{Paul2023ChainOfVerification}
Paul, D. e.~a. 2023.
\newblock Chain-of-Verification Reduces Hallucination in LLM Reasoning.
\newblock \emph{arXiv preprint arXiv:2309.11495}.

\bibitem[{Polu and Sutskever(2022)}]{Polu2022Minerva}
Polu, S.; and Sutskever, I. 2022.
\newblock Minerva: Solving Quantitative Reasoning Problems with Language Models.
\newblock \emph{arXiv preprint arXiv:2206.14858}.

\bibitem[{Schick, Dwivedi-Yu, and Lazaridou(2023)}]{Schick2023toolformer}
Schick, T.; Dwivedi-Yu, J.; and Lazaridou, A. e.~a. 2023.
\newblock Toolformer: Language Models Can Teach Themselves to Use Tools.
\newblock \emph{arXiv preprint arXiv:2302.04761}.

\bibitem[{Selsam(2019)}]{Selsam2019NeuroSAT}
Selsam, D. e.~a. 2019.
\newblock Learning a SAT Solver from Single-Bit Supervision.
\newblock \emph{ICML}.

\bibitem[{Shinn and Cassano(2023)}]{Shinn2023Reflexion}
Shinn, N.; and Cassano, F. e.~a. 2023.
\newblock Reflexion: An Autonomous Agent with Dynamic Memory and Self-Reflection.
\newblock \emph{arXiv preprint arXiv:2303.11366}.

\bibitem[{Stechly, Valmeekam, and Kambhampati(2024)}]{stechly2024self}
Stechly, K.; Valmeekam, K.; and Kambhampati, S. 2024.
\newblock On the self-verification limitations of large language models on reasoning and planning tasks.
\newblock \emph{arXiv preprint arXiv:2402.08115}.

\bibitem[{Tian et~al.(2024)Tian, Ye, Qin, Cong, Lin, Pan, Wu, Haotian, Weichuan, and Liu}]{tian2024debugbench}
Tian, R.; Ye, Y.; Qin, Y.; Cong, X.; Lin, Y.; Pan, Y.; Wu, Y.; Haotian, H.; Weichuan, L.; and Liu, Z. e.~a. 2024.
\newblock DebugBench: Evaluating Debugging Capability of Large Language Models.
\newblock In \emph{Findings of the Association for Computational Linguistics ACL 2024}, 4173--4198.

\bibitem[{Valmeekam et~al.(2022)Valmeekam, Olmo, Sreedharan, and Kambhampati}]{valmeekam2022large}
Valmeekam, K.; Olmo, A.; Sreedharan, S.; and Kambhampati, S. 2022.
\newblock Large Language Models Still Can’t Plan: A Benchmark for LLMs on Planning and Reasoning about Change.
\newblock In \emph{NeurIPS 2022 Foundation Models for Decision Making Workshop}.

\bibitem[{Valmeekam(2022)}]{Valmeekam2022}
Valmeekam, K. e.~a. 2022.
\newblock Large language models still can’t plan: A benchmark for LLMs on planning and reasoning about change.
\newblock In \emph{NeurIPS Foundation Models for Decision Making Workshop}.

\bibitem[{Wang(2022)}]{Wang2022}
Wang, X. e.~a. 2022.
\newblock Self-consistency improves chain-of-thought reasoning in language models.
\newblock \emph{arXiv:2203.11171}.

\bibitem[{Wei(2022)}]{Wei2022}
Wei, J. e.~a. 2022.
\newblock Chain-of-thought prompting elicits reasoning in large language models.
\newblock In \emph{NeurIPS}.

\bibitem[{Welleck(2022)}]{Welleck2022NaturalProofs}
Welleck, S. e.~a. 2022.
\newblock NaturalProofs: Formal Theorem Proving with LLMs.
\newblock In \emph{ICML}.

\bibitem[{Yang et~al.(2025)Yang, Li, Yang, Zhang, Hui, Zheng, Yu, Gao, Huang, and Lv}]{yang2025qwen3}
Yang, A.; Li, A.; Yang, B.; Zhang, B.; Hui, B.; Zheng, B.; Yu, B.; Gao, C.; Huang, C.; and Lv, C. e.~a. 2025.
\newblock Qwen3 technical report.
\newblock \emph{arXiv preprint arXiv:2505.09388}.

\end{thebibliography}

\newpage
\clearpage
\section*{Supplementary Material}

\subsection*{Algorithm 1: SOFAI-LM for Graph Coloring}
\begin{algorithm}[h]
\caption{Feedback-Driven LLM with LRM Fallback (Graph Coloring)}
\begin{algorithmic}[1]
\REQUIRE Undirected graph $G=(V,E)$, color limit $k$, max iterations $T$
\ENSURE Coloring assignment or ``NOT SOLVABLE''
\STATE $history \gets []$                \COMMENT{stores $(assign, feedback)$}
\STATE $trends \gets []$                 \COMMENT{stores feedback and mistakes}
\FOR{$t = 1$ to $T$}
  \STATE $prompt \gets$ GenerateGCPrompt$(G,k,history)$
  \STATE $resp \gets$ CallLLM$(prompt)$
  \STATE $assign \gets$ ParseColoring$(resp)$
  \STATE $C \gets \frac{|\{(u,v)\in E : assign[u]\neq assign[v]\}|}{|E|}$
  \IF{$C = 1.0$}
    \STATE \RETURN $assign$              \COMMENT{all edges valid → solved}
  \ENDIF
  \STATE $conflicts \gets \{(u,v)\in E : assign[u] = assign[v]\}$
  \STATE $subG \gets$ InducedSubgraph$(G,$ vertices in $conflicts)$
  \STATE $fb \gets$ FormatFeedbackMLF$(conflicts, subG)$
  \STATE $history$.append$(assign, fb)$
  \STATE $trends$.append($fb$)         
\ENDFOR
\STATE 
  \COMMENT{Three prompting variants:}
  \STATE // (1) PO: problem instance only
  \STATE $p_{PO} \gets$ GenerateLRMPromptPO$(G,k)$
  \STATE $r_{PO} \gets$ CallLRM$(p_{PO})$
  \STATE // (2) BA: best LLM attempt + problem
  \STATE $best \gets$ argmax$(history, \text{highest }C)$
  \STATE $p_{BA} \gets$ GenerateLRMPromptBA$(G,k,best)$
  \STATE $r_{BA} \gets$ CallLRM$(p_{BA})$
  \STATE // (3) FH: full LLM feedback history
  \STATE $p_{FH} \gets$ GenerateLRMPromptFH$(G,k,history)$
  \STATE $r_{FH} \gets$ CallLRM$(p_{FH})$
  \STATE Choose final $r_{*}$ based on selected variant
  \STATE $assign^* \gets$ ParseColoring$(r_{*})$
  \STATE \RETURN $assign^*$
\end{algorithmic}
\end{algorithm}

\vspace{0.5em}
\noindent\textbf{Description (Graph Coloring):}
\begin{itemize}
  \item \textbf{Evaluation Function $C(assign)$}: Computes the fraction of edges properly colored (Eq. \(2\) in paper):
    $C=\frac{|\{(u,v)\in E: assign[u]\neq assign[v]\}|}{|E|}$. A value of 1.0 means a valid coloring.
  \item \textbf{Improvement Trend}: Track $fb$ each iteration to measure progress and detect stagnation. If stagnation detected, it invokes LRM
  \item \textbf{Adaptive Feedback}: On conflicts, extract the induced subgraph of miscolored edges and generate multi-line feedback (MLF) listing conflicting pairs and the subgraph structure.
  \item \textbf{LRM Prompting Levels}: 
    \begin{enumerate}
      \item PO (Problem-Only): Only the original graph and $k$.
      \item BA (Best Attempt): Graph, $k$, and the single best LLM assignment.
      \item FH (Full History): Graph, $k$, and the entire $(assign,feedback)$ history.
    \end{enumerate}
  \item \textbf{Fallback Selection}: Empirically, PO yields highest success for graph coloring (Fig. 4 left). The module selects the variant that maximizes solve rate vs. time.
\end{itemize}


\subsection*{Algorithm 2: SOFAI-LM for Code Debugging}
\begin{algorithm}[h]
\caption{Feedback-Driven LLM with LRM Fallback (Code Debugging)}
\begin{algorithmic}[1]
\REQUIRE Buggy snippet $C_b$, description $x$, test suite $\mathcal{T}$, max iterations $T$
\ENSURE Patched code $C^*$
\STATE $history \gets []$              \COMMENT{stores $(code, feedback)$}
\STATE $trends \gets []$               \COMMENT{stores feedback and mistakes}
\FOR{$t = 1$ to $T$}
  \STATE $prompt \gets$ GenerateCDPrompt$(C_b,x,history)$
  \STATE $resp \gets$ CallLLM$(prompt)$
  \STATE $code \gets$ ParseCode$(resp)$
  \STATE $(passed, total) \gets$ LeetCodeAPI.Test$(code,\mathcal{T})$
  \STATE $P \gets passed/total$
  \IF{$P = 1.0$}
    \STATE \RETURN $code$              \COMMENT{all tests passed → solved}
  \ENDIF
  \STATE $fails \gets$ IdentifyFailures$(code,\mathcal{T})$
  \STATE $fb \gets$ FormatFeedbackSLF$(fails, passed, total)$
  \STATE $history$.append$(code, fb)$
  \STATE $trends$.append($fb$)        
\ENDFOR
\STATE 
  \COMMENT{Three prompting variants:}
  \STATE // PO: only problem description
  \STATE $p_{PO} \gets$ GenerateLRMPromptPO$(C_b,x)$
  \STATE $r_{PO} \gets$ CallLRM$(p_{PO})$
  \STATE // BA: problem + last LLM code
  \STATE $last \gets$ history[-1].code
  \STATE $p_{BA} \gets$ GenerateLRMPromptBA$(C_b,x,last)$
  \STATE $r_{BA} \gets$ CallLRM$(p_{BA})$
  \STATE // FH: full LLM feedback history
  \STATE $p_{FH} \gets$ GenerateLRMPromptFH$(C_b,x,history)$
  \STATE $r_{FH} \gets$ CallLRM$(p_{FH})$
  \STATE Select final $r_{*}$ based on variant trade-offs
  \STATE $code^* \gets$ ParseCode$(r_{*})$
  \STATE \RETURN $code^*$
\end{algorithmic}
\end{algorithm}

\vspace{0.5em}
\noindent\textbf{Description (Code Debugging):}
\begin{itemize}
  \item \textbf{LeetCode API Evaluation $P(code)$}: Returns pass ratio $P=\frac{\#passed}{\#total}$; a value of 1.0 indicates full correctness.
  \item \textbf{Improvement Trend}: Records feedback and mistakes each iteration to monitor learning. If it identifies stagnation, LRM is invoked.
  \item \textbf{Feedback Generation}: Use single-line feedback (SLF) listing failing testcase IDs and pass/total for concise guidance.
  \item \textbf{LRM Prompting Levels}: Same three variants (PO, BA, FH). For code debugging, BA and FH often improve solve rate at cost of time (Fig. 4 right).
  \item \textbf{Fallback Selection}: The metacognitive module chooses the prompting level that best balances accuracy and inference time for the domain.
\end{itemize}

\subsection{ Sample Input Prompts}

\begin{figure*}[t]
  \centering
  \footnotesize

  \textbf{Graph Coloring ($|V|=10$, $k=4$)}\\
  \begin{verbatim}
### Task: Graph Coloring Decision Problem (< 5 colors)
You must assign an integer color to every vertex of the undirected graph below such that 
no two adjacent vertices share the same color. If no coloring exists using at most 4 
colors, respond exactly with: NOT SOLVABLE
Do not output anything else or use quotes.
### Input Graph 
 p edges 10 12
 c edges
 a b
 a c
 b d
 b e
 c f
 d g
 e h
 f i
 g j
 h i
 h j
 i j
### Output Format
Provide one (vertex color) pair per line, sorted lexicographically:
(a 1)
(b 2)
...
Where `color` is an integer in the inclusive range [1, 4].
Return only this list, with no prose, headings, or extra punctuation.

### Constraints Recap
- Use < 5 distinct colors.
- Vertex identifiers are case-sensitive and must match those in the input.
- Output must be either the ordered list above or the token `NOT SOLVABLE`.
  \end{verbatim}

  \vspace{1em}

  \textbf{Code Debugging (`the-kth-factor-of-n`)}\\
  \begin{verbatim}
You are an expert programmer specializing in code debugging. Your task is to analyze and 
fix the provided code snippet based on the problem description. IMPORTANT: You MUST return 
the complete, corrected code enclosed within <code> and </code> tags. Do not include any 
other explanatory text in your response.
### Problem Description
You are given two positive integers n and k. A factor of an integer n is defined as an 
integer i where n % i == 0. Consider a list of all factors of n sorted in ascending order, 
return the kth factor in this list or return -1 if n has less than k factors.
### Buggy Code
```
class Solution:
    def kthFactor(self, n: int, k: int) -> int:
        j = 0
        for i in range(1, n + 1):
            if n % i == 0:
                num = i
                j += 1
            if j == k:
                break
        return num if j == k+1 else -1
```
### Correct Code:
  \end{verbatim}

  \caption{Sample input prompts for the Graph Coloring and Code Debugging domains.}
\end{figure*}

\clearpage

\subsection{Detailed Walk-through: Graph Coloring}

This section provides a step-by-step illustration of the SOFAI-LM architecture's workflow on a representative graph coloring problem, with a maximum of two iterations using LLM. We demonstrate the iterative feedback loop, the generation of adaptive examples, successful and unsuccessful resolution paths by the LLM, and the final structure of the episodic memory.

\subsection{Problem Instance}
The architecture is tasked with solving a 4-coloring problem for a graph with 10 vertices and 12 edges. The prompt is provided to the System 1 (LLM) solver (Figure \ref{fig:prompt_instance}).

\begin{figure*}[t!]
\begin{verbatim}
### Task: Graph Coloring Decision Problem (< 5 colors)
You must assign an integer color to every vertex of the undirected graph 
below such that no two adjacent vertices share the same color. If no 
coloring exists using at most 4 colors, respond exactly with: NOT SOLVABLE
Do not output anything else or use quotes.

### Input Graph 
 p edges 10 12
 c edges
 a b, a c, b d, b e, c f, d g, e h, f i, g j, h i, h j, i j

### Output Format
Provide one (vertex color) pair per line, sorted lexicographically:
(a 1)
(b 2)
...
Where `color` is an integer in the inclusive range [1, 4].
Return only this list, with no prose, headings, or extra punctuation.
\end{verbatim}
\caption{The initial prompt given to the LLM for the graph coloring task.}
\label{fig:prompt_instance}
\end{figure*}

\subsection{Scenario 1: Initial Attempt and Feedback Generation}
The LLM processes the prompt and produces its first candidate solution (Figure \ref{fig:sol_attempt1}). This solution contains several constraint violations. Specifically, it assigns the same color to adjacent vertices in the subgraph formed by nodes \texttt{\{h, i, j\}}.

\begin{figure*}[h!]
\centering
\begin{verbatim}
(a 1)  (b 2)  (c 2)  (d 1)  (e 1)
(f 3)  (g 2)  (h 3)  (i 3)  (j 3)
\end{verbatim}
\caption{The LLM's first incorrect candidate solution.}
\label{fig:sol_attempt1}
\end{figure*}

\begin{figure}[H]
    \centering
    \includegraphics[width=0.8\columnwidth]{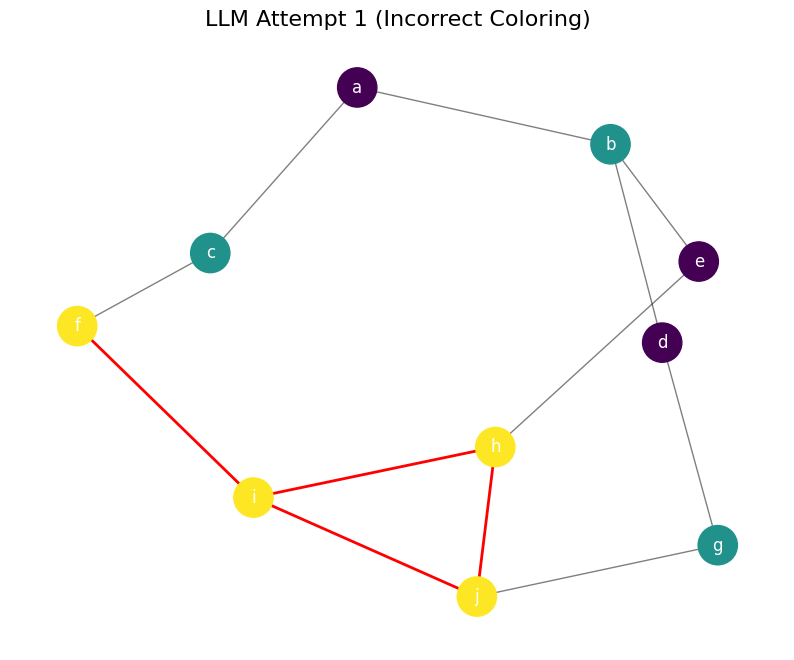}
    \caption{Visualization of the LLM's first incorrect attempt. Conflicting edges are highlighted in red.}
    \label{fig:vis_attempt1}
\end{figure}

\subsubsection{Metacognitive Feedback}
The Metacognitive Governance (MC) module evaluates the solution, identifies the errors, and generates feedback in multiple formats to guide the LLM (Figure \ref{fig:feedback1}).

\begin{figure*}[t!]
\paragraph{Single-Line Feedback (SLF):}
\begin{verbatim}
That was incorrect. The coloring is invalid: 
adjacent-conflict(s) on pairs: (h,i), (h,j), (i,j).
\end{verbatim}

\paragraph{Multi-Line Feedback (MLF):}
\begin{verbatim}
That was incorrect. The coloring is invalid for the following reason(s):
  1. adjacent-conflict: vertices h and i share color 3
  2. adjacent-conflict: vertices h and j share color 3
  3. adjacent-conflict: vertices i and j share color 3
\end{verbatim}

\paragraph{Adaptive Example:}
\begin{verbatim}
---
### Adaptive Example
To help you, here is a smaller, related subproblem:

Subproblem Graph:
p edge 3 3
e h i 
e h j 
e i j

Coloring:
(h 1)  (i 2)  (j 3)
---
\end{verbatim}
\caption{Feedback generated by the MC module, including concise (SLF), structured (MLF), and adaptive example formats.}
\label{fig:feedback1}
\end{figure*}

\subsection{Scenario 2: Iterative Failure and LRM Fallback}
This scenario illustrates the case where the LLM fails to find a solution within $T=2$ iterations, triggering a fallback to the System 2 (LRM) solver. After receiving feedback, the LLM tries again but introduces a new error (Figure \ref{fig:sol_attempt2_incorrect}).

\begin{figure*}[h!]
\centering
\begin{verbatim}
(a 2)  (b 1)  (c 2)  (d 3)  (e 4)
(f 1)  (g 2)  (h 1)  (i 2)  (j 3)
\end{verbatim}
\caption{The LLM's second incorrect candidate solution.}
\label{fig:sol_attempt2_incorrect}
\end{figure*}

\begin{figure}[H]
    \centering
    \includegraphics[width=0.8\columnwidth]{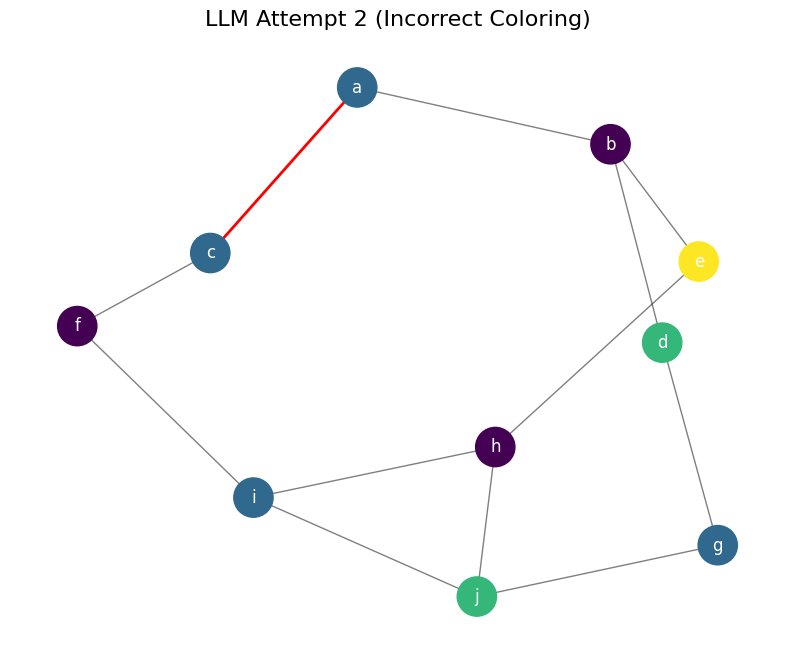}
    \caption{Visualization of the LLM's second incorrect attempt. The new conflicting edge (a,c) is highlighted.}
    \label{fig:vis_attempt2_incorrect}
\end{figure}

\subsubsection{Fallback to LRM}
Having failed to converge, the MC module invokes the LRM using the \textbf{Problem-Only (PO)} prompting strategy.

\subsection{Scenario 3: Iterative Success}
This path demonstrates a successful outcome. Using the feedback from its first attempt, the LLM generates a new, valid coloring (Figure \ref{fig:sol_attempt2_correct}).

\begin{figure*}[h!]
\centering
\begin{verbatim}
(a 1)  (b 2)  (c 3)  (d 1)  (e 3)
(f 1)  (g 2)  (h 2)  (i 4)  (j 1)
\end{verbatim}
\caption{The LLM's second, and correct, candidate solution.}
\label{fig:sol_attempt2_correct}
\end{figure*}

\begin{figure}[H]
    \centering
    \includegraphics[width=0.8\columnwidth]{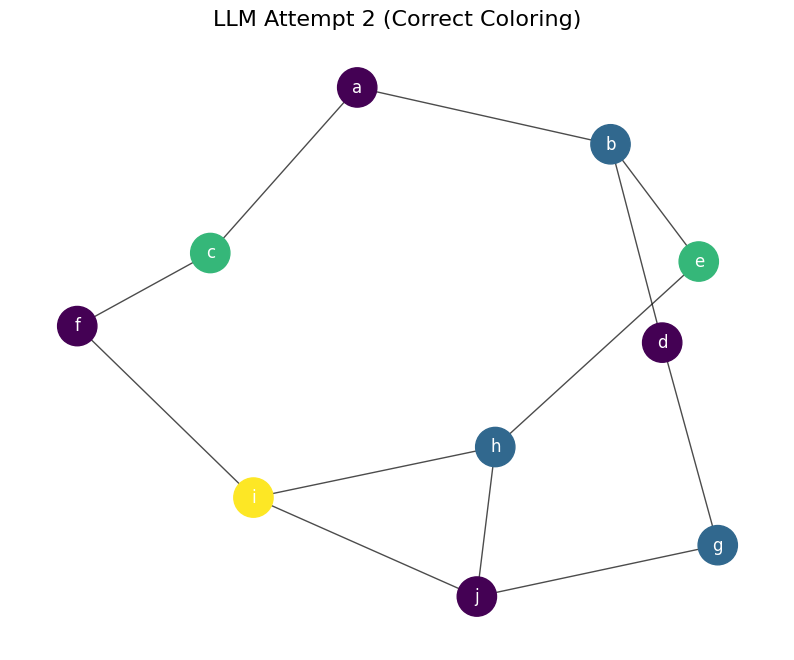}
    \caption{Visualization of the LLM's successful attempt. The coloring is valid.}
    \label{fig:vis_attempt2_correct}
\end{figure}

\subsection{Scenario 4: Episodic Memory Storage}
Following the successful resolution, the MC module updates the episodic memory. The content stored depends on the chosen memory variant (Figure \ref{fig:memory_variants}).

\begin{figure*}[t!]
\subsubsection*{Minimal Episodic Memory (MEM)}
\begin{verbatim}
{
  "problem_instance": {
    "task": "Graph Coloring Decision Problem (< 5 colors)",
    "graph": "p edges 10 12\n c edges\n a b\n ...",
    "k": 4
  },
  "correct_solution": "(a 1)\n(b 2)\n(c 3)\n(d 1)\n(e 3)\n..."
}
\end{verbatim}

\subsubsection*{Extended Episodic Memory (EEM)}
\begin{verbatim}
{
  "problem_instance": { ... },
  "interaction_history": [
    {
      "attempt": 1,
      "candidate_solution": "(a 1)\n(b 2)\n(c 2)\n...",
      "feedback_received": "That was incorrect. ..."
    },
    {
      "attempt": 2,
      "candidate_solution": "(a 1)\n(b 2)\n(c 3)\n...",
      "feedback_received": "Correct"
    }
  ],
  "correct_solution": "(a 1)\n(b 2)\n(c 3)\n..."
}
\end{verbatim}
\caption{Comparison of memory storage variants. MEM stores the final solution, while EEM stores the full interaction history.}
\label{fig:memory_variants}
\end{figure*}

\clearpage

\subsection{Detailed Walk-through: Code Debugging}

This section details the SOFAI-LM workflow for a code debugging task from the DebugBench benchmark, with a maximum of two iterations using LLM. This domain requires localized fixes and benefits from a different feedback strategy than graph coloring.

\subsection{Problem Instance}
The task is to fix a bug in a Python function named \texttt{kthFactor}. The initial prompt provided to the LLM is shown in Figure \ref{fig:code_prompt_instance}.

\begin{figure*}[t!]
\begin{verbatim}
You are an expert programmer specializing in code debugging. Your task is to 
analyze and fix the provided code snippet based on the problem description. 
IMPORTANT: You MUST return the complete, corrected code enclosed within 
<code> and </code> tags. Do not include any other explanatory text in your response.

### Problem Description
You are given two positive integers n and k. A factor of an integer n is defined 
as an integer i where n % i == 0. Consider a list of all factors of n sorted 
in ascending order, return the kth factor in this list or return -1 if n has 
less than k factors.

### Buggy Code
```python
class Solution:
    def kthFactor(self, n: int, k: int) -> int:
        j = 0
        for i in range(1, n + 1):
            if n % i == 0:
                num = i
                j += 1
                if j == k:
                    break
        return num if j == k+1 else -1
```
\end{verbatim}
\caption{The initial prompt for the code debugging task 'the-kth-factor-of-n'.}
\label{fig:code_prompt_instance}
\end{figure*}

\subsection{Scenario 1: Iterative Success}
This scenario shows the LLM successfully debugging the code after receiving targeted feedback from the testing environment.

\subsubsection{LLM's First Attempt (Incorrect)}
The LLM attempts a fix but introduces a new logical error. The loop condition is now incorrect, causing it to miss factors for some inputs. The proposed code is shown in Figure \ref{fig:code_attempt1}.

\begin{figure*}[t!]
\begin{verbatim}
<code>
class Solution:
    def kthFactor(self, n: int, k: int) -> int:
        factors = []
        for i in range(1, n // 2 + 1):  # Incorrect loop range
            if n % i == 0:
                factors.append(i)
        factors.append(n)
        
        if k <= len(factors):
            return factors[k - 1]
        else:
            return -1
</code>
\end{verbatim}
\caption{The LLM's first, incorrect attempt to fix the buggy code.}
\label{fig:code_attempt1}
\end{figure*}

\subsubsection{Metacognitive Feedback from Test Harness}
The submitted code is executed against a hidden test suite. The MC module receives the failure result and formats it as feedback (Figure \ref{fig:code_feedback1}). For code debugging, a concise feedback string (SLF) is effective.

\begin{figure*}[t!]
\begin{verbatim}
Your code failed.
- Failing Test Case Input: n = 4, k = 3
- Expected Output: 4
- Your Code's Output: -1
- Reason: Your code's logic for finding factors is incorrect.
Please re-examine the code and provide a new, corrected version.
\end{verbatim}
\caption{Feedback generated after the first attempt failed a test case.}
\label{fig:code_feedback1}
\end{figure*}

\subsubsection{LLM's Second Attempt (Correct)}
Using the feedback, the LLM corrects its logic and produces the correct code (Figure \ref{fig:code_attempt2_correct}). This version correctly iterates through all possible factors up to `n`.

\begin{figure*}[t!]
\begin{verbatim}
<code>
class Solution:
    def kthFactor(self, n: int, k: int) -> int:
        count = 0
        for i in range(1, n + 1):
            if n % i == 0:
                count += 1
                if count == k:
                    return i
        return -1
</code>
\end{verbatim}
\caption{The LLM's second, successful attempt at fixing the bug.}
\label{fig:code_attempt2_correct}
\end{figure*}

This corrected code passes all test cases, and the MC module returns it as the final solution.

\subsection{Scenario 2: LLM Fails \& LRM is Invoked}
In this scenario, the LLM fails to produce a correct solution after the maximum number of iterations.

\textbf{Attempt 1} is the same as in the successful scenario (Figure \ref{fig:code_attempt1}), resulting in the feedback shown in Figure \ref{fig:code_feedback1}.

\textbf{Attempt 2 (Incorrect):} The LLM makes a different error, incorrectly handling the 1-based vs. 0-based indexing of `k`.
\begin{figure*}[t!]
\begin{verbatim}
<code>
class Solution:
    def kthFactor(self, n: int, k: int) -> int:
        factors = []
        for i in range(1, n + 1):
            if n % i == 0:
                factors.append(i)
        
        if k < len(factors): # Off-by-one error
            return factors[k]
        else:
            return -1
</code>
\end{verbatim}
\caption{A second, different incorrect attempt from the LLM.}
\label{fig:code_attempt2_incorrect}
\end{figure*}

\subsubsection{Fallback to LRM}
After this second failure, the MC module would invoke the LRM. As our results show, providing the LRM with the \textbf{Best Attempt (BA)} or \textbf{Full History (FH)} is beneficial in this domain. The LRM would receive the original problem and either the last code attempt or the full sequence of attempts and feedback to produce the final, correct solution.

\clearpage

\subsection{Experimental Results : Graph Coloring}
 We detail the performance of various model combinations across different problem sizes for the graph coloring domain.

\subsection{RQ1: Can a feedback-driven LLM outperform an LRM?}
The following figures illustrate the performance comparison between a feedback-driven LLM and a standalone LRM across various model combinations and graph sizes for solvable graph coloring problems.

\subsubsection{Model Combination 1: Granite 3.3B (w/o thinking) vs. DeepSeek R1 8B}
\begin{figure}[h!]
    \centering
    \includegraphics[width=0.95\linewidth]{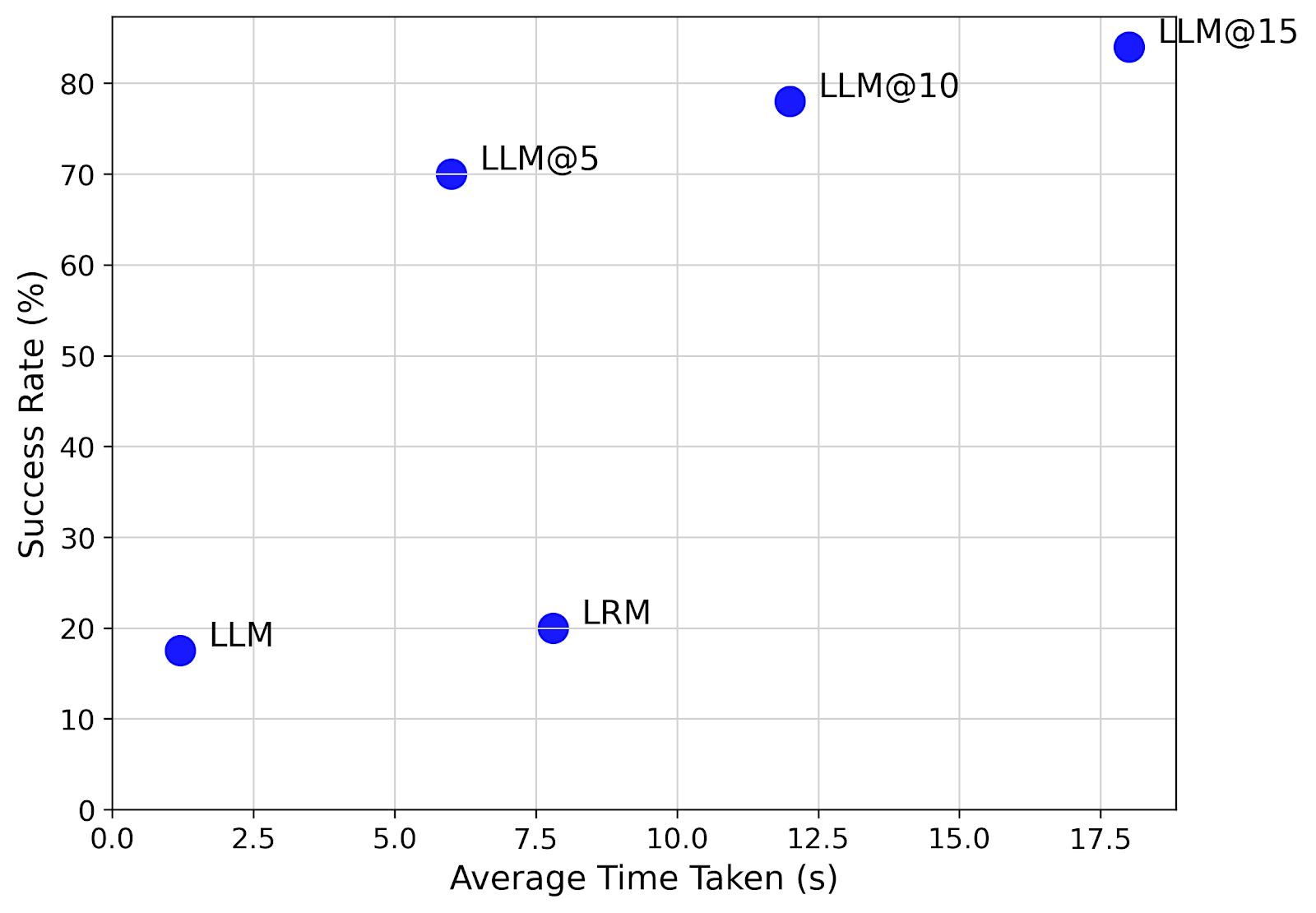}
    \caption{RQ1 - Model Combination 1, Size 5: Success rate versus average time for an iterative LLM (Granite 3.3B w/o thinking) and a standalone LRM (DeepSeek R1 8B) on solvable graph coloring problems of size 5.}
    \label{fig:rq1_mc1_5}
\end{figure}

\begin{figure}[h!]
    \centering
    \includegraphics[width=0.95\linewidth]{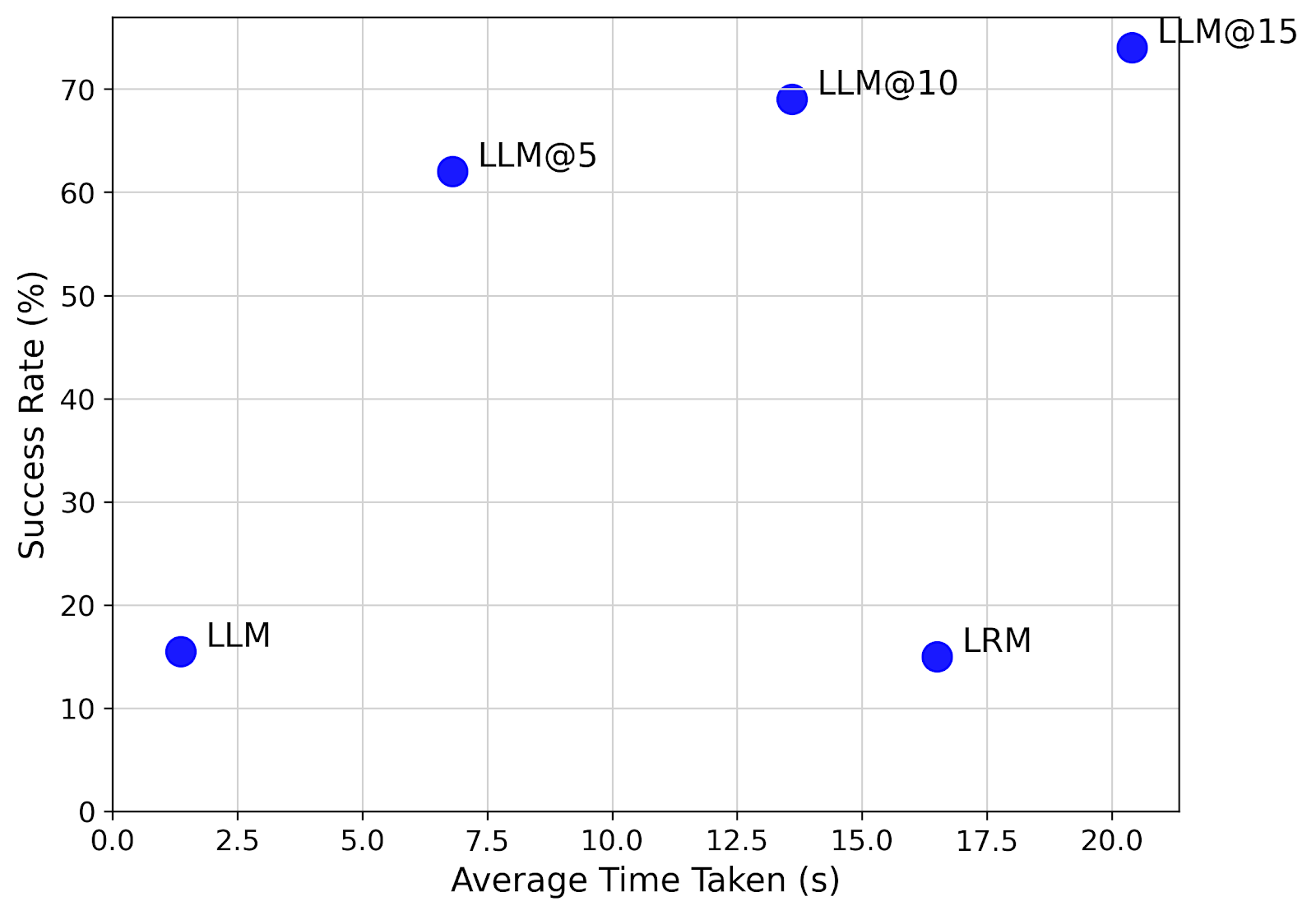}
    \caption{RQ1 - Model Combination 1, Size 10: Success rate versus average time for an iterative LLM (Granite 3.3B w/o thinking) and a standalone LRM (DeepSeek R1 8B) on solvable graph coloring problems of size 10.}
    \label{fig:rq1_mc1_10}
\end{figure}

\begin{figure}[h!]
    \centering
    \includegraphics[width=0.95\linewidth]{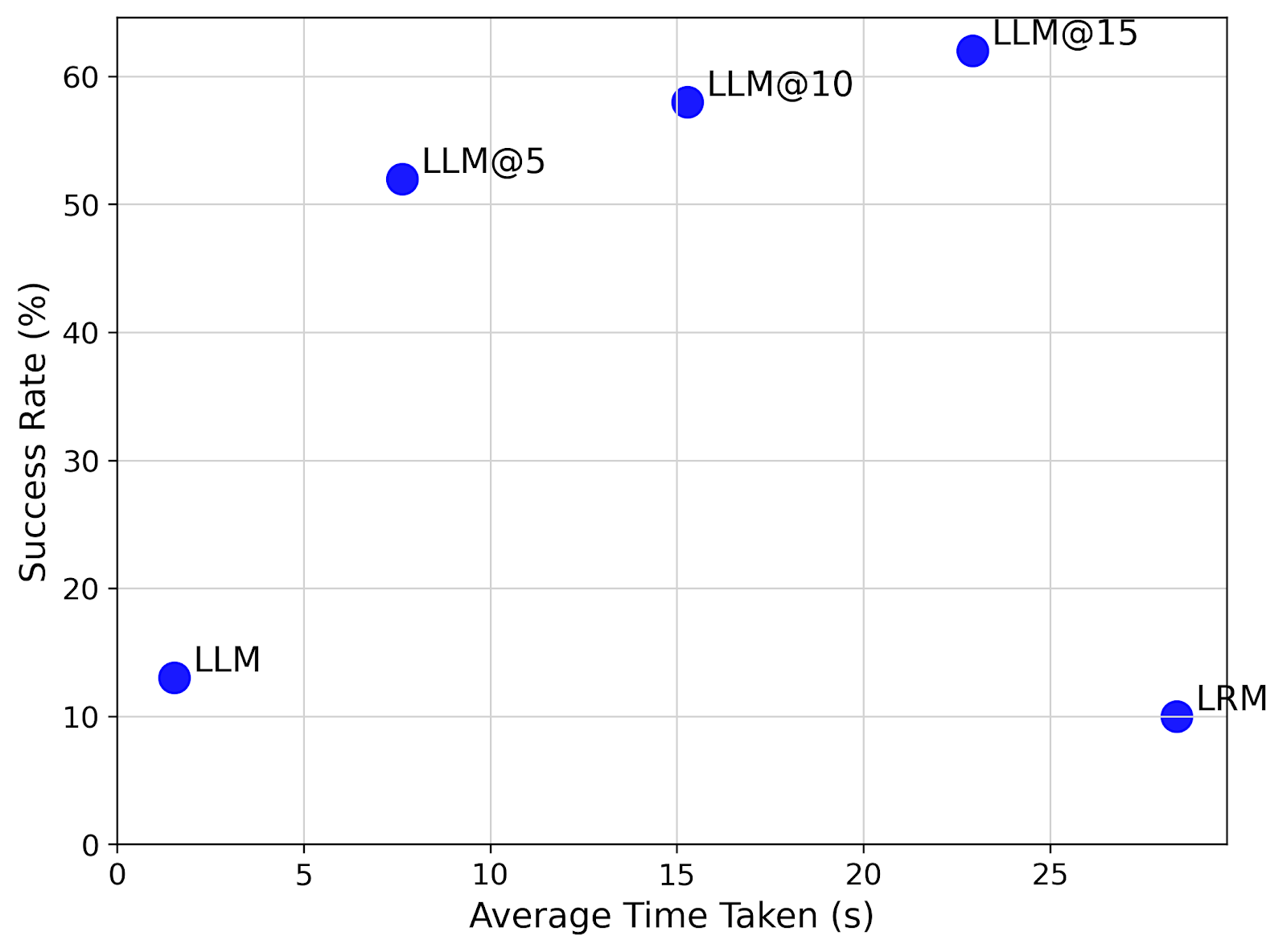}
    \caption{RQ1 - Model Combination 1, Size 15: Success rate versus average time for an iterative LLM (Granite 3.3B w/o thinking) and a standalone LRM (DeepSeek R1 8B) on solvable graph coloring problems of size 15.}
    \label{fig:rq1_mc1_15}
\end{figure}

\begin{figure}[h!]
    \centering
    \includegraphics[width=0.95\linewidth]{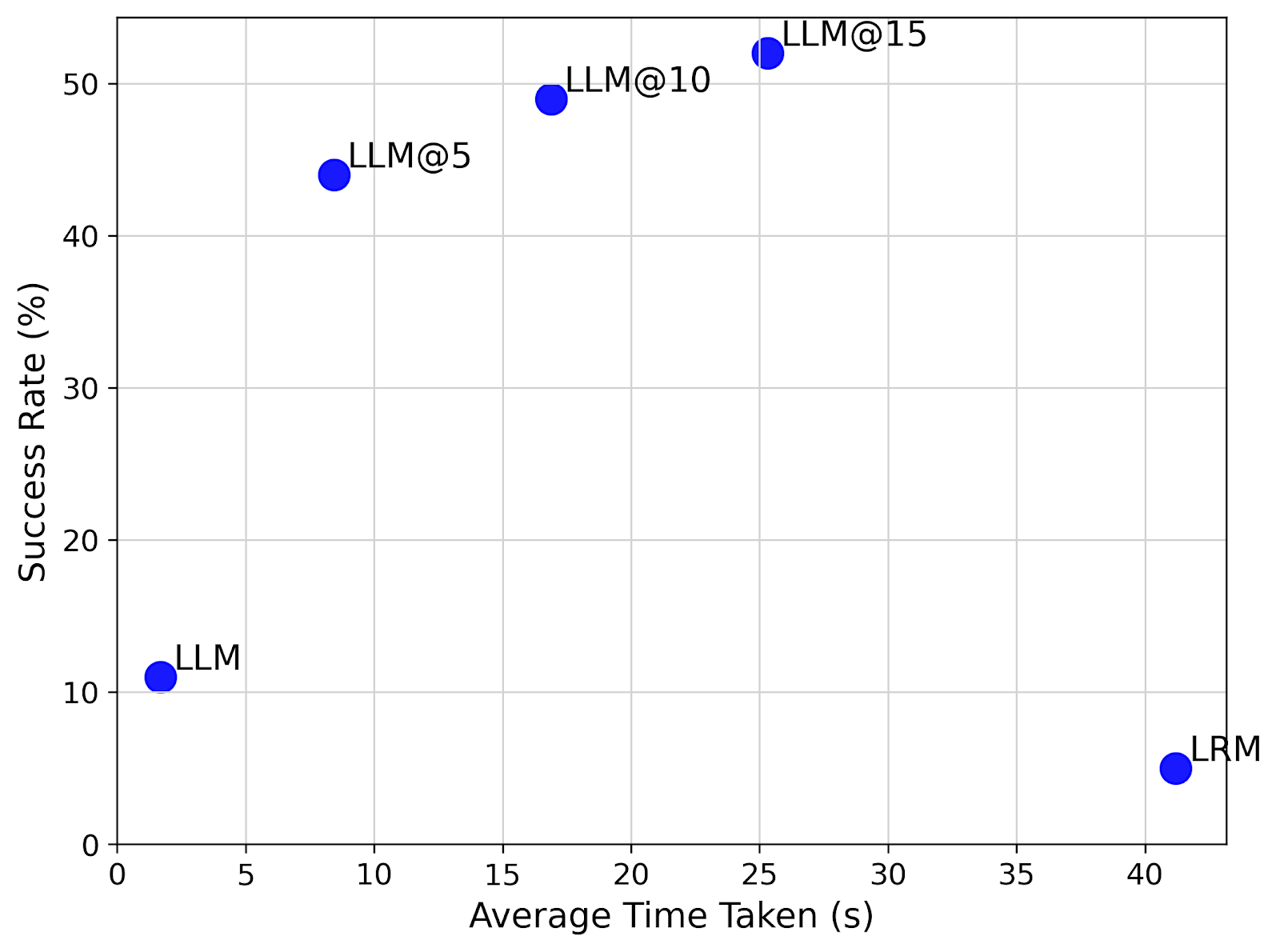}
    \caption{RQ1 - Model Combination 1, Size 20: Success rate versus average time for an iterative LLM (Granite 3.3B w/o thinking) and a standalone LRM (DeepSeek R1 8B) on solvable graph coloring problems of size 20.}
    \label{fig:rq1_mc1_20}
\end{figure}

\paragraph{Description}
The set of figures (\ref{fig:rq1_mc1_5} through \ref{fig:rq1_mc1_20}) compares the success rate and average time taken for two different approaches to solving graph coloring problems. The first approach uses Granite 3.3B as a fast LLM, showing its performance as a single-pass model (LLM) and with 5, 10, and 15 iterations of feedback (LLM@5, LLM@10, LLM@15). The second approach uses DeepSeek R1 8B as a standalone slow-thinking LRM. Across all graph sizes, the iterative LLM configurations consistently achieve a higher success rate than the standalone LRM. As the number of iterations increases from 5 to 15, the LLM's success rate improves, albeit with a corresponding increase in time. Notably, as the graph size increases, the performance of the standalone LRM diminishes rapidly, whereas the iterative LLM maintains a significantly higher success rate, highlighting the efficacy of the feedback loop.

\paragraph{Summary Finding}
A feedback-driven LLM consistently and substantially outperforms a standalone LRM in both success rate and efficiency for solving graph coloring problems, with the performance gap widening as graph size increases.

\clearpage

\subsubsection{Model Combination 2: Granite 3.3B (w/o thinking) vs. Granite 3.3B (w/ thinking)}
\begin{figure}[h!]
    \centering
    \includegraphics[width=0.95\linewidth]{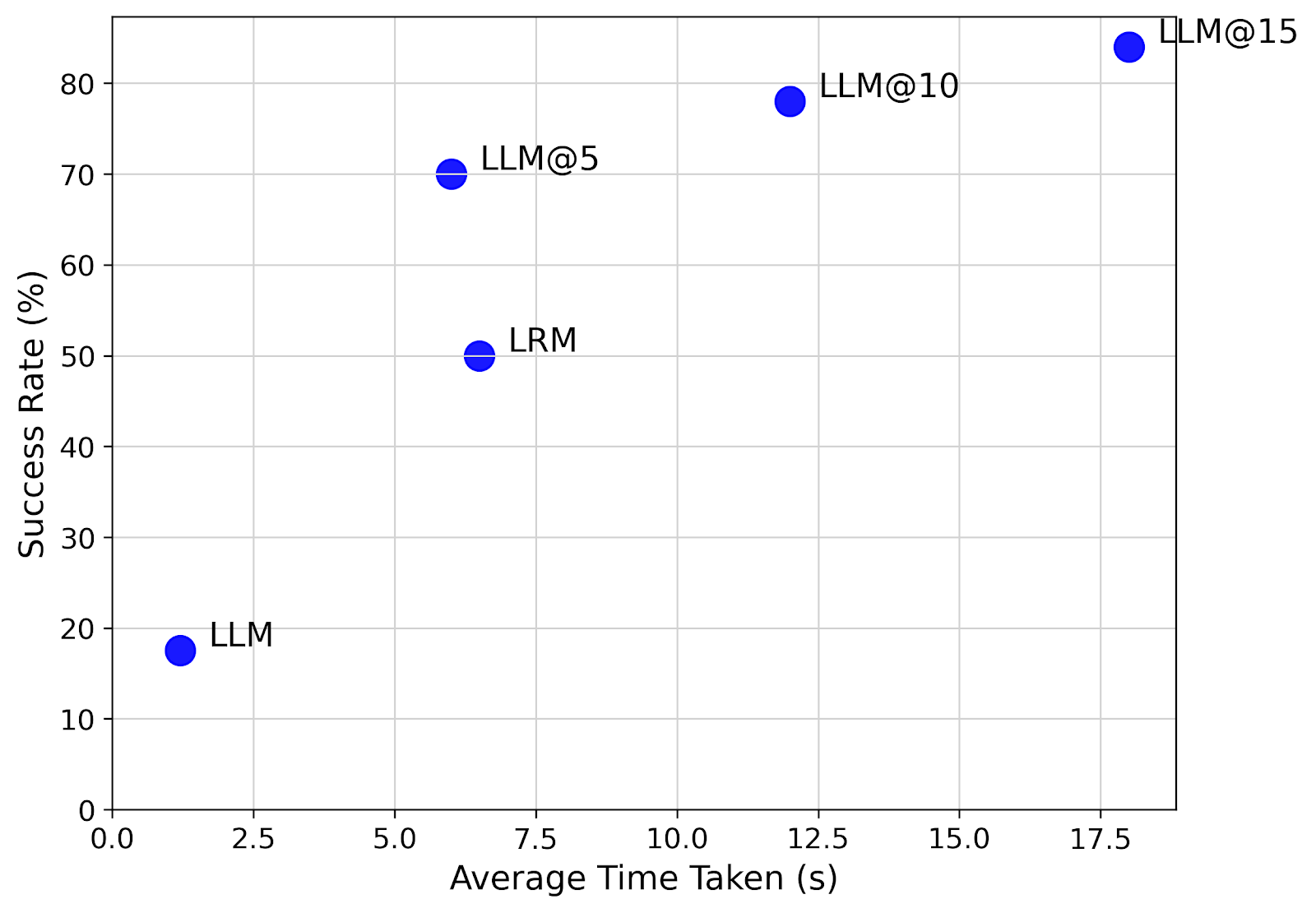}
    \caption{RQ1 - Model Combination 2, Size 5}
    \label{fig:rq1_mc2_5}
\end{figure}

\begin{figure}[h!]
    \centering
    \includegraphics[width=0.95\linewidth]{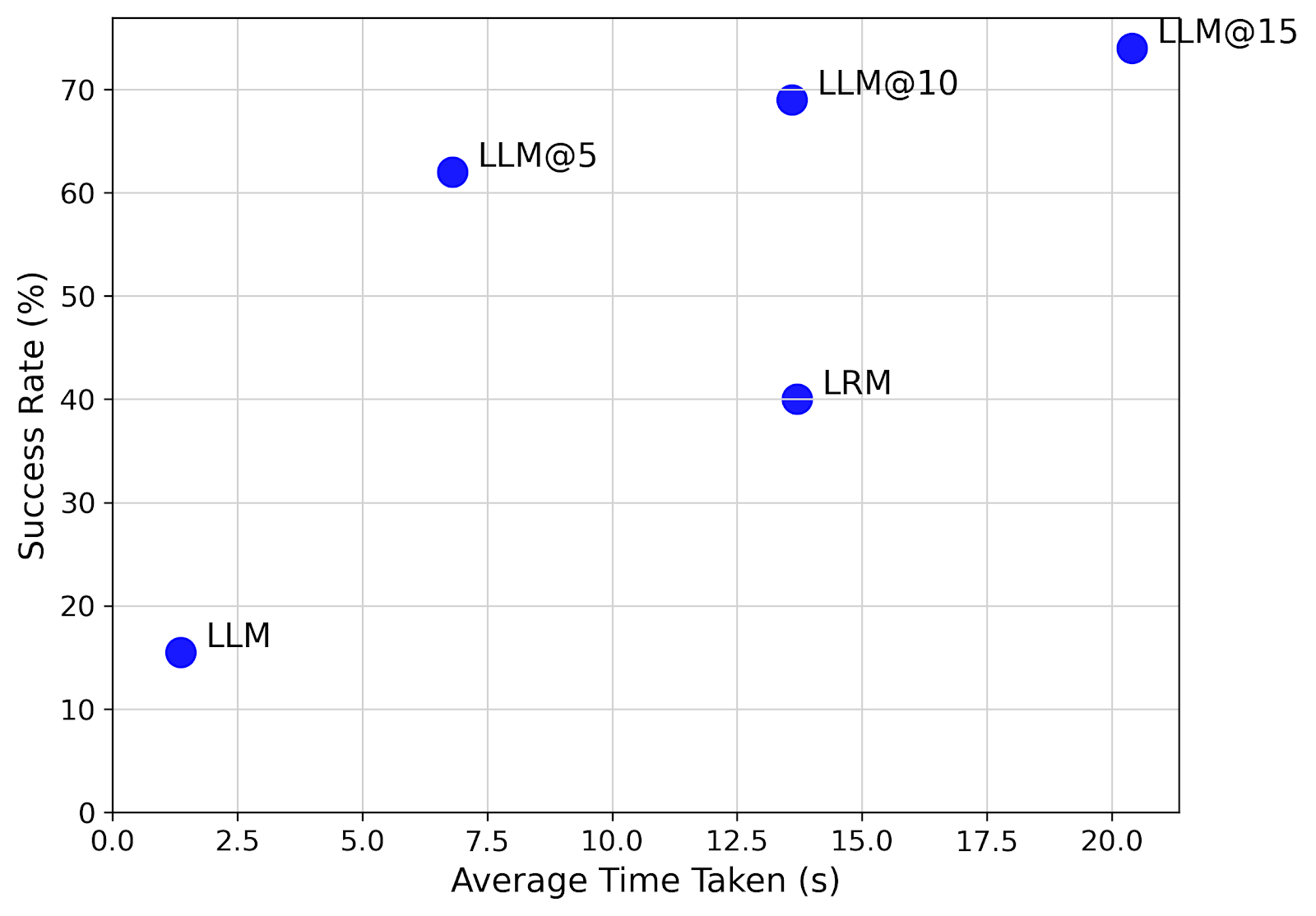}
    \caption{RQ1 - Model Combination 2, Size 10}
    \label{fig:rq1_mc2_10}
\end{figure}

\begin{figure}[h!]
    \centering
    \includegraphics[width=0.95\linewidth]{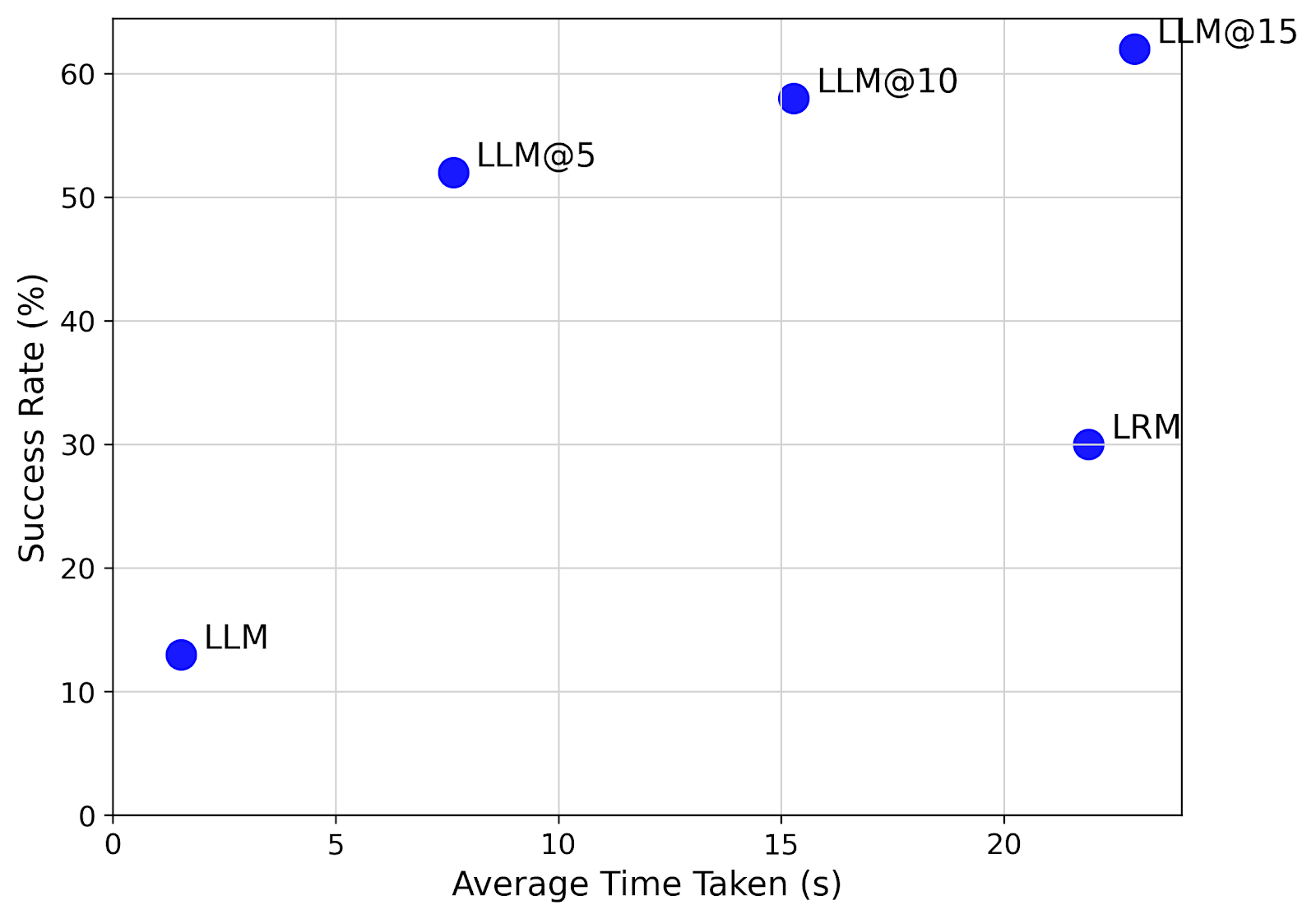}
    \caption{RQ1 - Model Combination 2, Size 15}
    \label{fig:rq1_mc2_15}
\end{figure}

\begin{figure}[h!]
    \centering
    \includegraphics[width=0.95\linewidth]{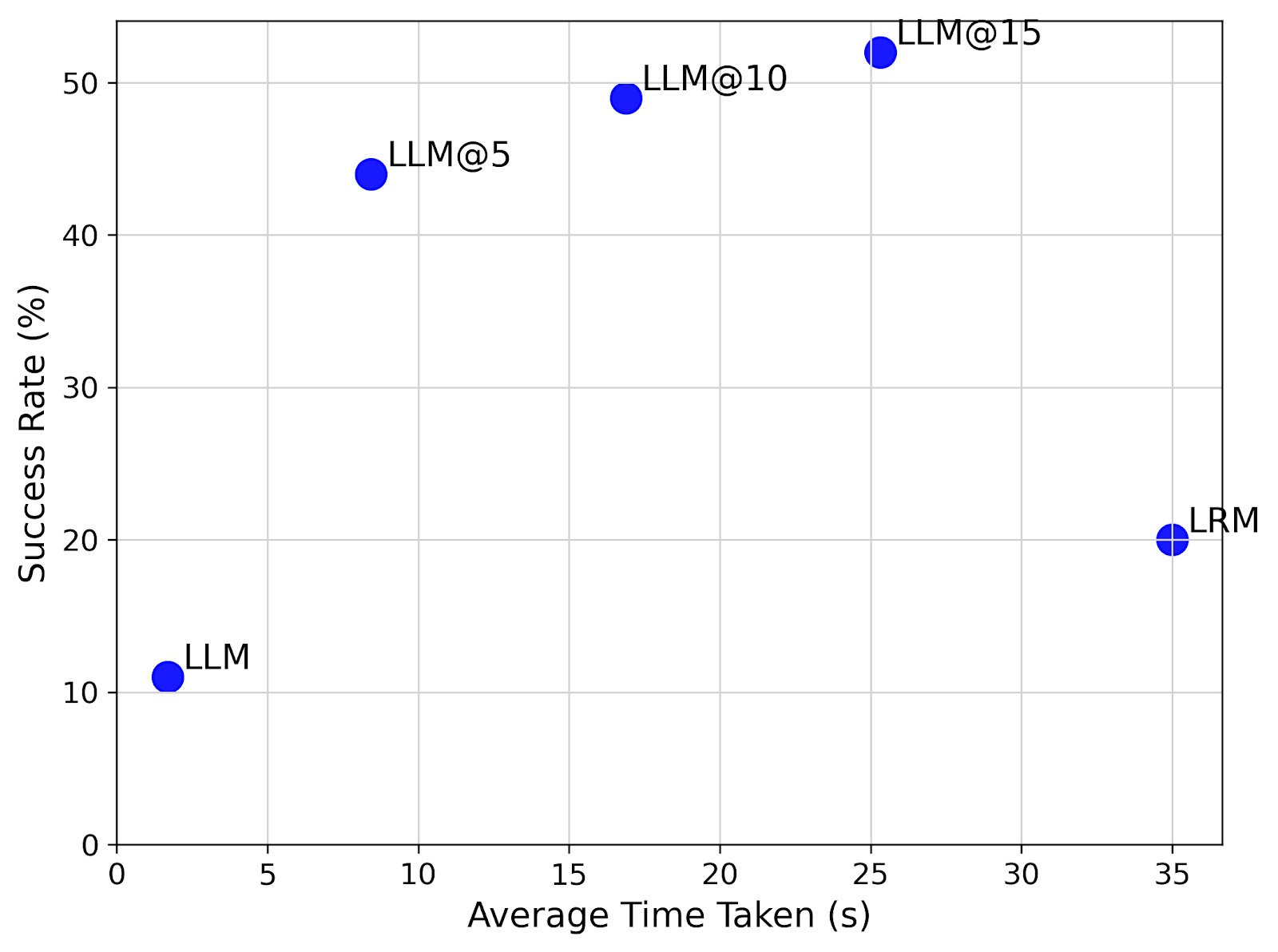}
    \caption{RQ1 - Model Combination 2, Size 20}
    \label{fig:rq1_mc2_20}
\end{figure}

\begin{figure}[h!]
    \centering
    \includegraphics[width=0.95\linewidth]{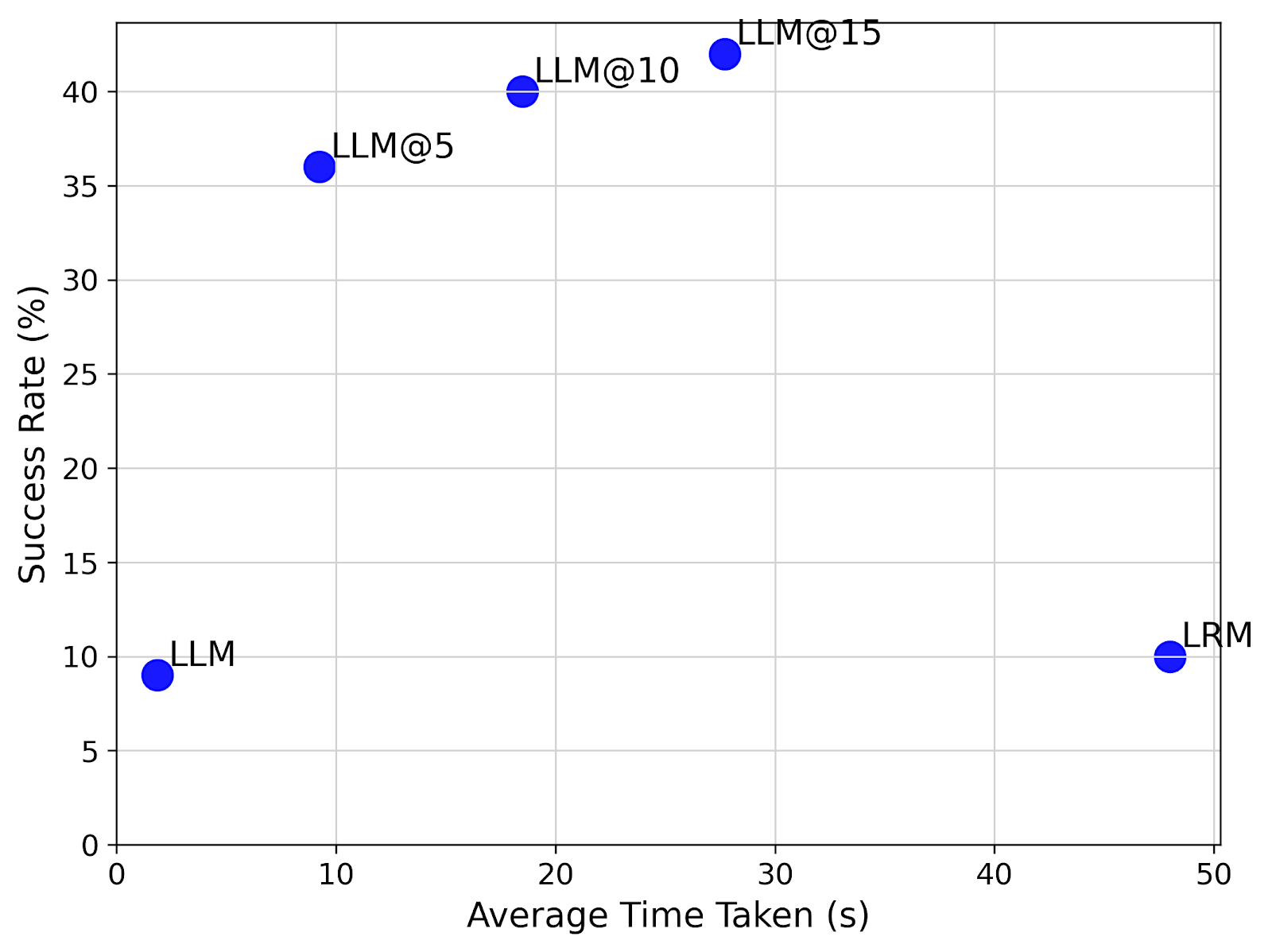}
    \caption{RQ1 - Model Combination 2, Size 25}
    \label{fig:rq1_mc2_25}
\end{figure}

\paragraph{Description}
This series of plots (Figures \ref{fig:rq1_mc2_5} through \ref{fig:rq1_mc2_25}) illustrates the performance trade-offs between using Granite 3.3B in an iterative feedback loop (LLM) versus its standalone reasoning mode (LRM). While the LRM is faster for smaller problems (sizes 5-15), its success rate is consistently lower than the iterative LLM configurations (LLM@5 and higher). For larger, more complex problems (sizes 20 and 25), the iterative LLM not only achieves a significantly higher success rate but also does so in less time than the standalone LRM. This demonstrates a clear advantage for the iterative, feedback-driven approach, even when comparing against the same base model in a specialized reasoning mode.

\paragraph{Summary Finding}
Iterative feedback enables a base LLM to achieve superior accuracy compared to the same model operating as a standalone LRM, and this performance advantage becomes more pronounced in both success rate and time efficiency as graph size increases.


\clearpage

\subsubsection{Model Combination 3: Granite 3.3B (w/o thinking) vs. Qwen 2.5 Pro}
\begin{figure}[h!]
    \centering
    \includegraphics[width=0.95\linewidth]{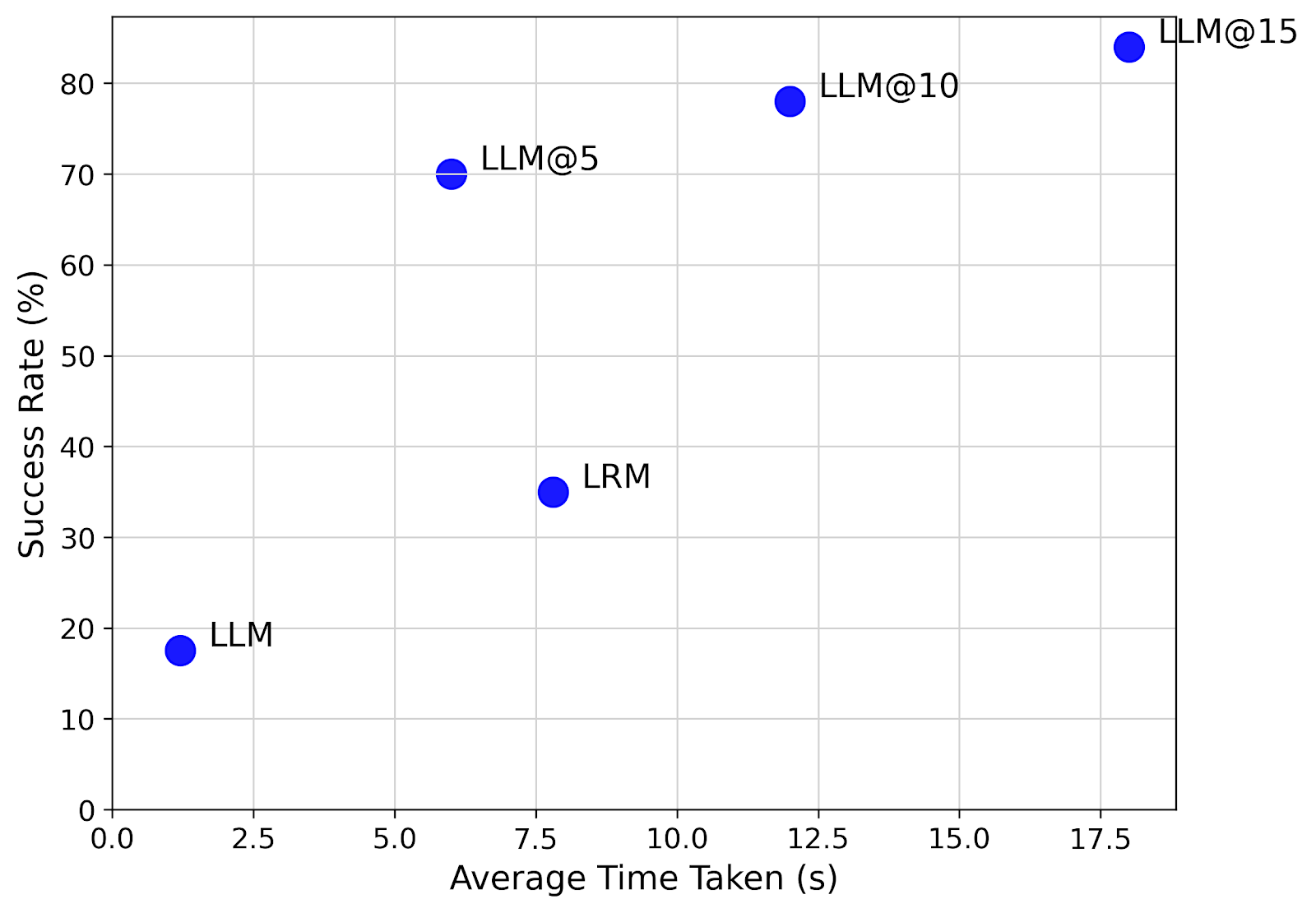}
    \caption{RQ1 - Model Combination 3, Size 5}
    \label{fig:rq1_mc3_5}
\end{figure}

\begin{figure}[h!]
    \centering
    \includegraphics[width=0.95\linewidth]{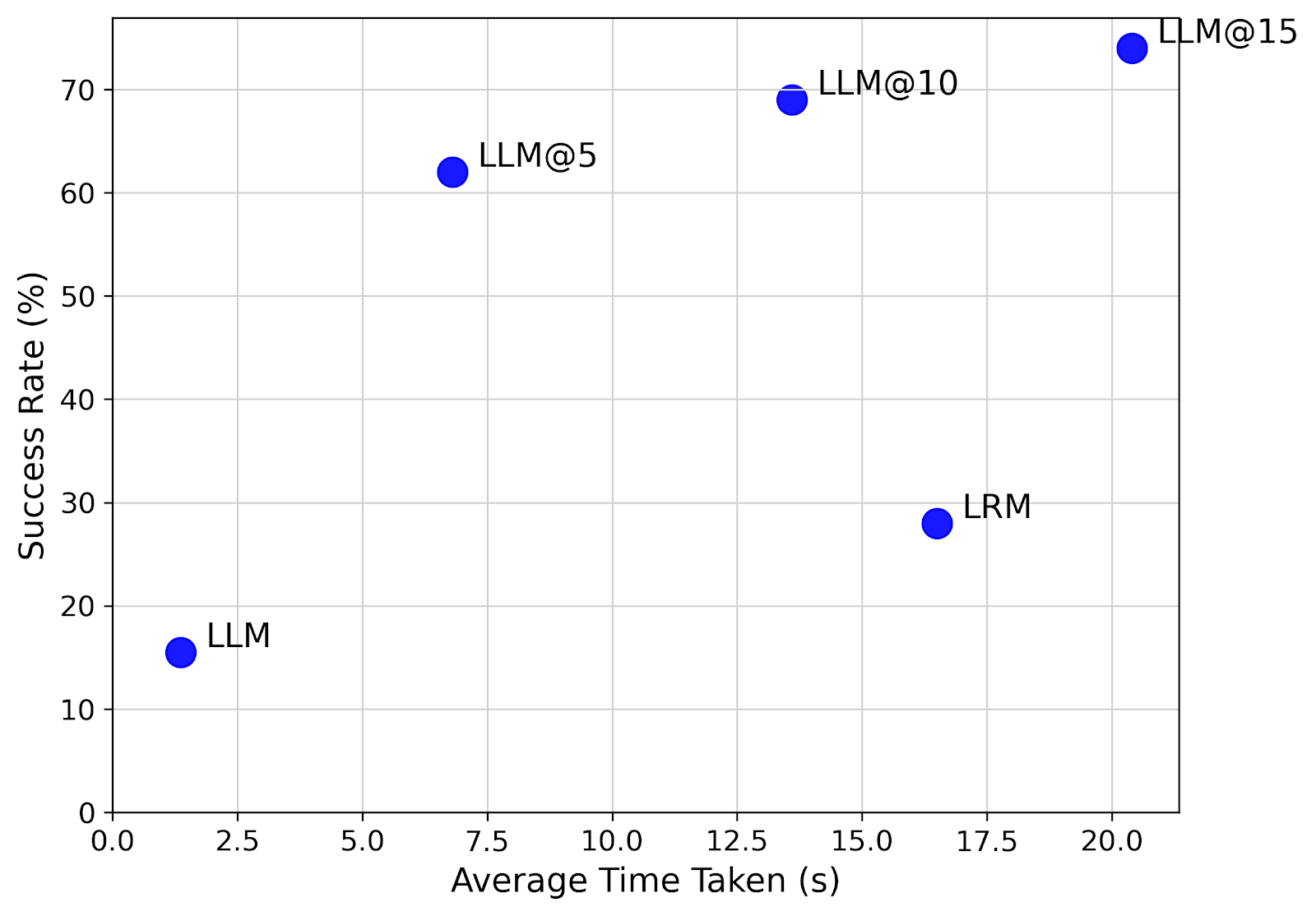}
    \caption{RQ1 - Model Combination 3, Size 10}
    \label{fig:rq1_mc3_10}
\end{figure}

\begin{figure}[h!]
    \centering
    \includegraphics[width=0.95\linewidth]{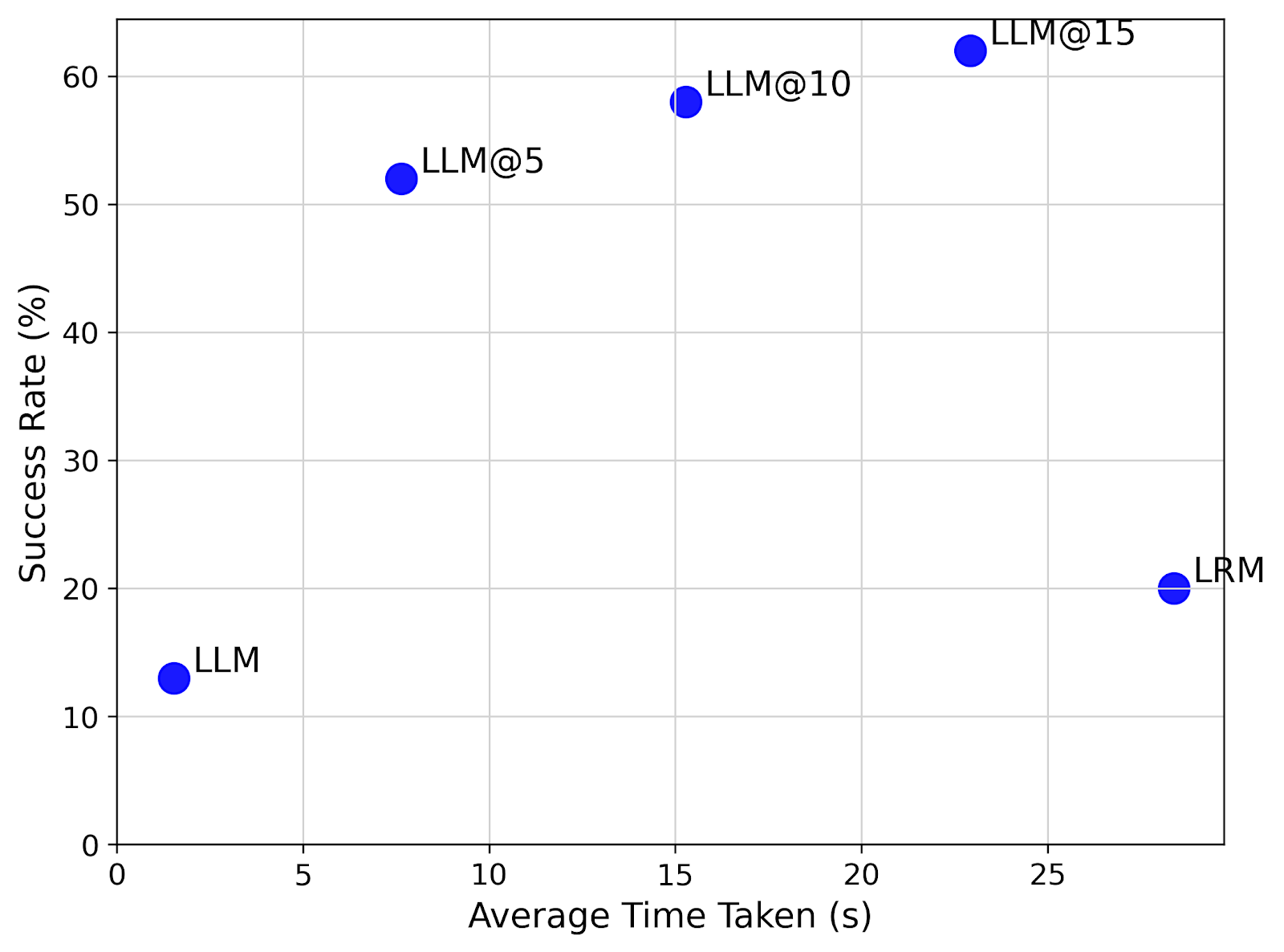}
    \caption{RQ1 - Model Combination 3, Size 15}
    \label{fig:rq1_mc3_15}
\end{figure}

\begin{figure}[h!]
    \centering
    \includegraphics[width=0.95\linewidth]{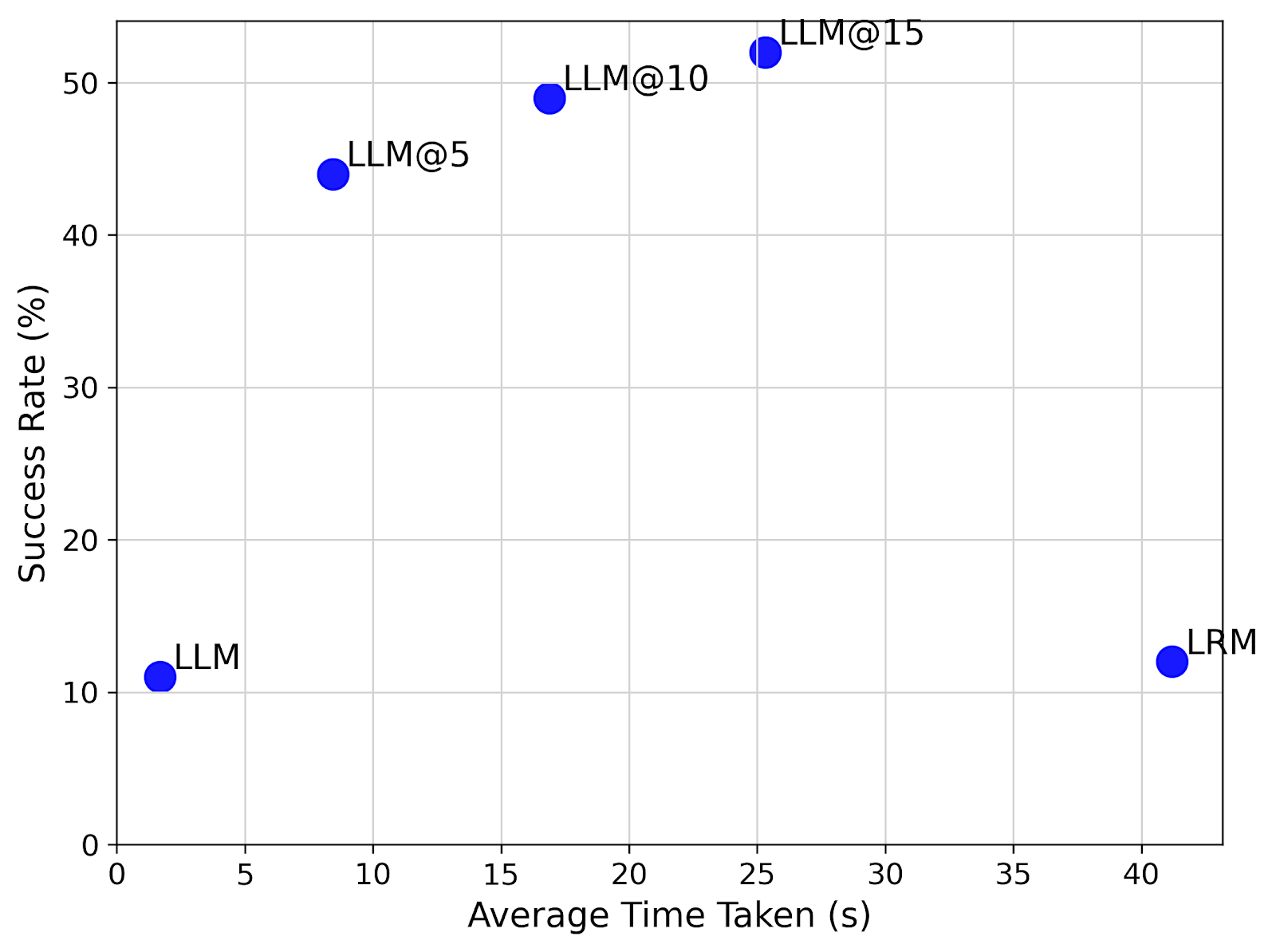}
    \caption{RQ1 - Model Combination 3, Size 20}
    \label{fig:rq1_mc3_20}
\end{figure}

\begin{figure}[h!]
    \centering
    \includegraphics[width=0.95\linewidth]{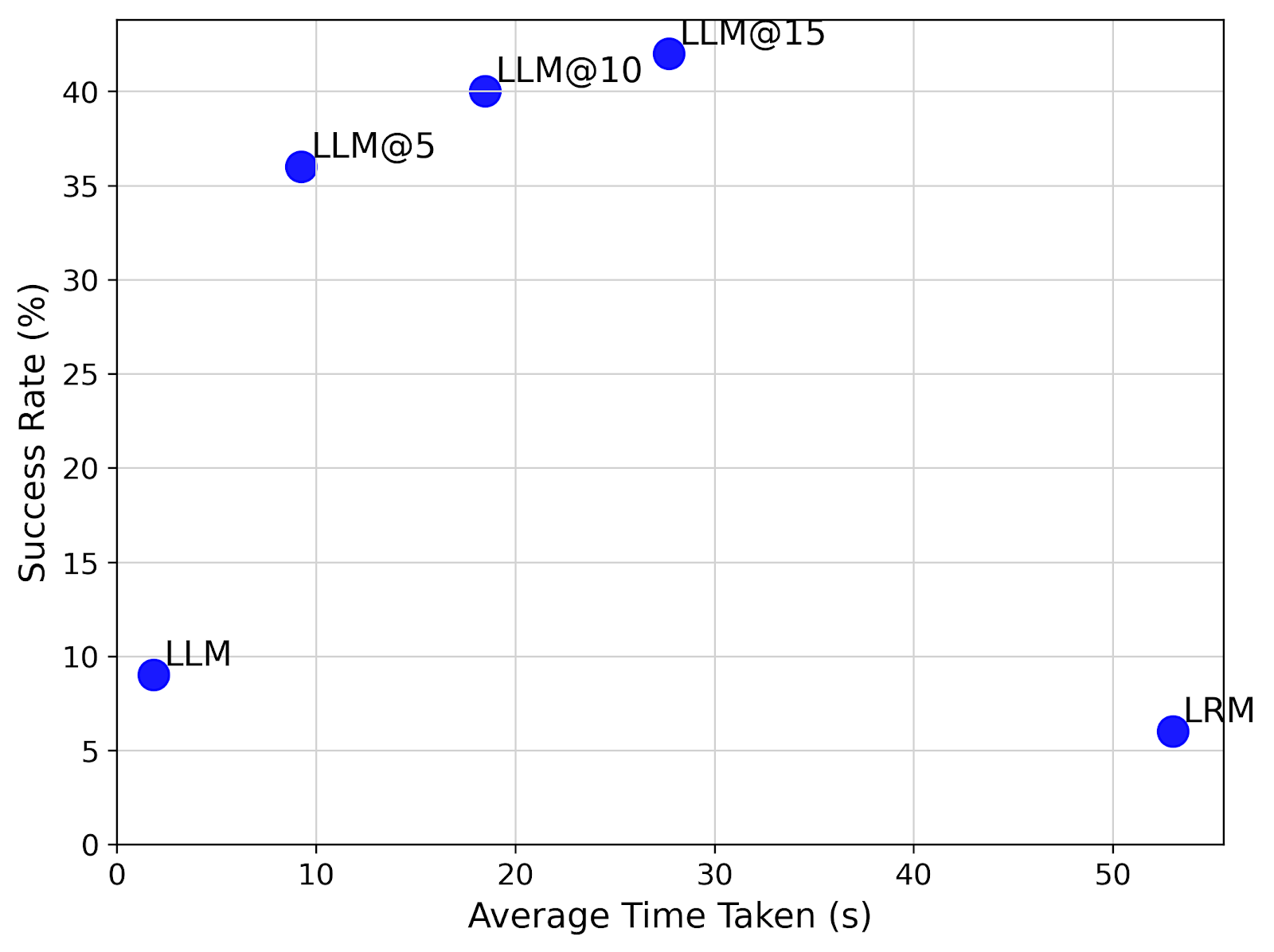}
    \caption{RQ1 - Model Combination 3, Size 25}
    \label{fig:rq1_mc3_25}
\end{figure}

\paragraph{Description}
These figures (Figures \ref{fig:rq1_mc3_5} through \ref{fig:rq1_mc3_25}) show the performance of the iterative Granite 3.3B LLM against the Qwen 2.5 Pro model acting as a standalone LRM. The trend is consistent across all graph sizes: the iterative LLM (LLM@5 and above) consistently achieves a much higher success rate than the Qwen 2.5 Pro LRM. While the LRM is faster, its low success rate makes it an unreliable standalone solver, especially as problem size increases. The feedback loop clearly empowers the Granite model to find correct solutions far more frequently, demonstrating a robust performance advantage over a capable LRM from a different model family.

\paragraph{Summary Finding}
An iterative LLM with feedback demonstrates a commanding performance lead in solution accuracy over a different, powerful standalone LRM, confirming that the benefit of the iterative architecture is not limited to a single model family.


\clearpage

\subsubsection{Model Combination 4: Llama 3.1 vs. DeepSeek R1 8B}
\begin{figure}[h!]
    \centering
    \includegraphics[width=0.95\linewidth]{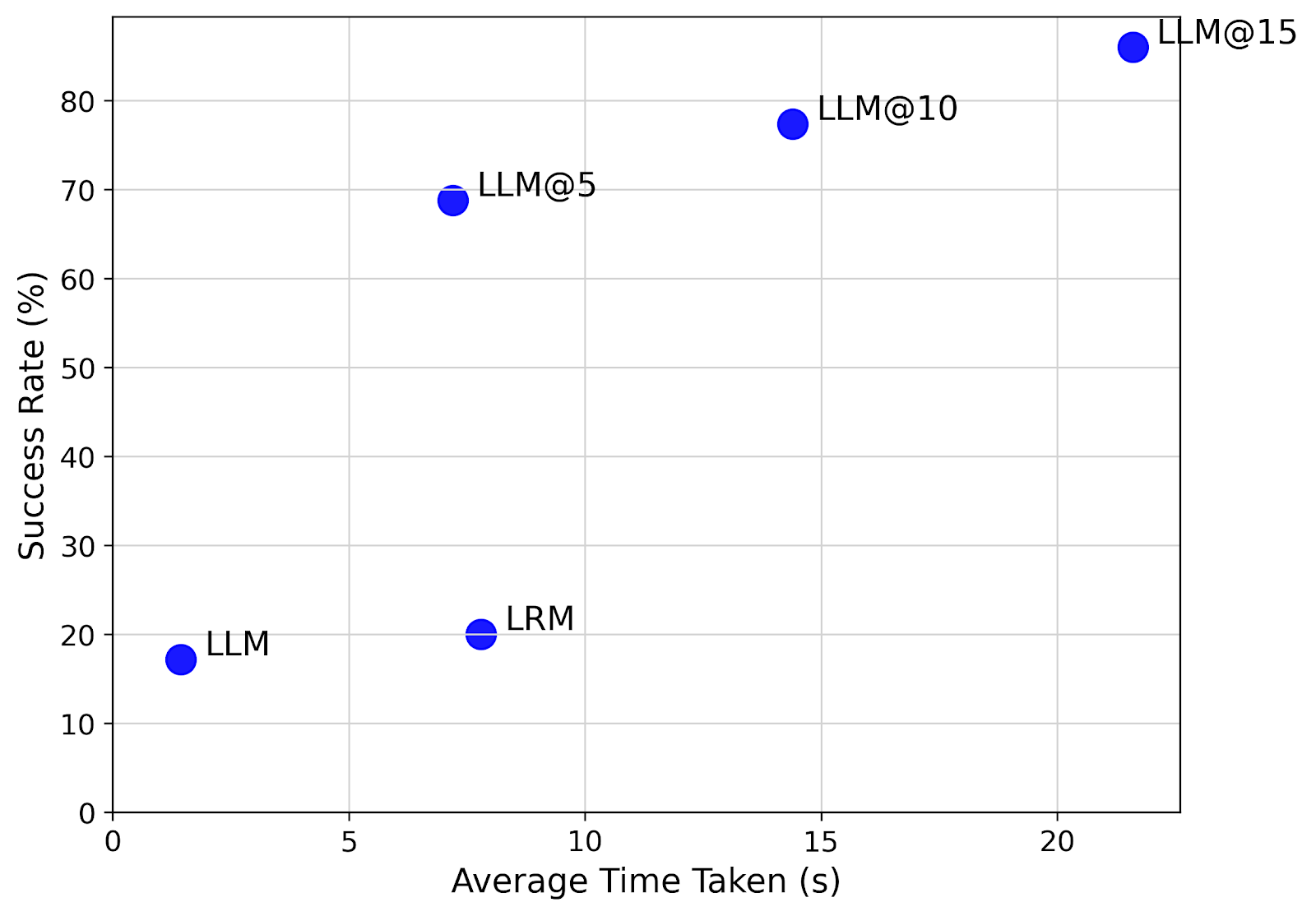}
    \caption{RQ1 - Model Combination 4, Size 5}
    \label{fig:rq1_mc4_5}
\end{figure}

\begin{figure}[h!]
    \centering
    \includegraphics[width=0.95\linewidth]{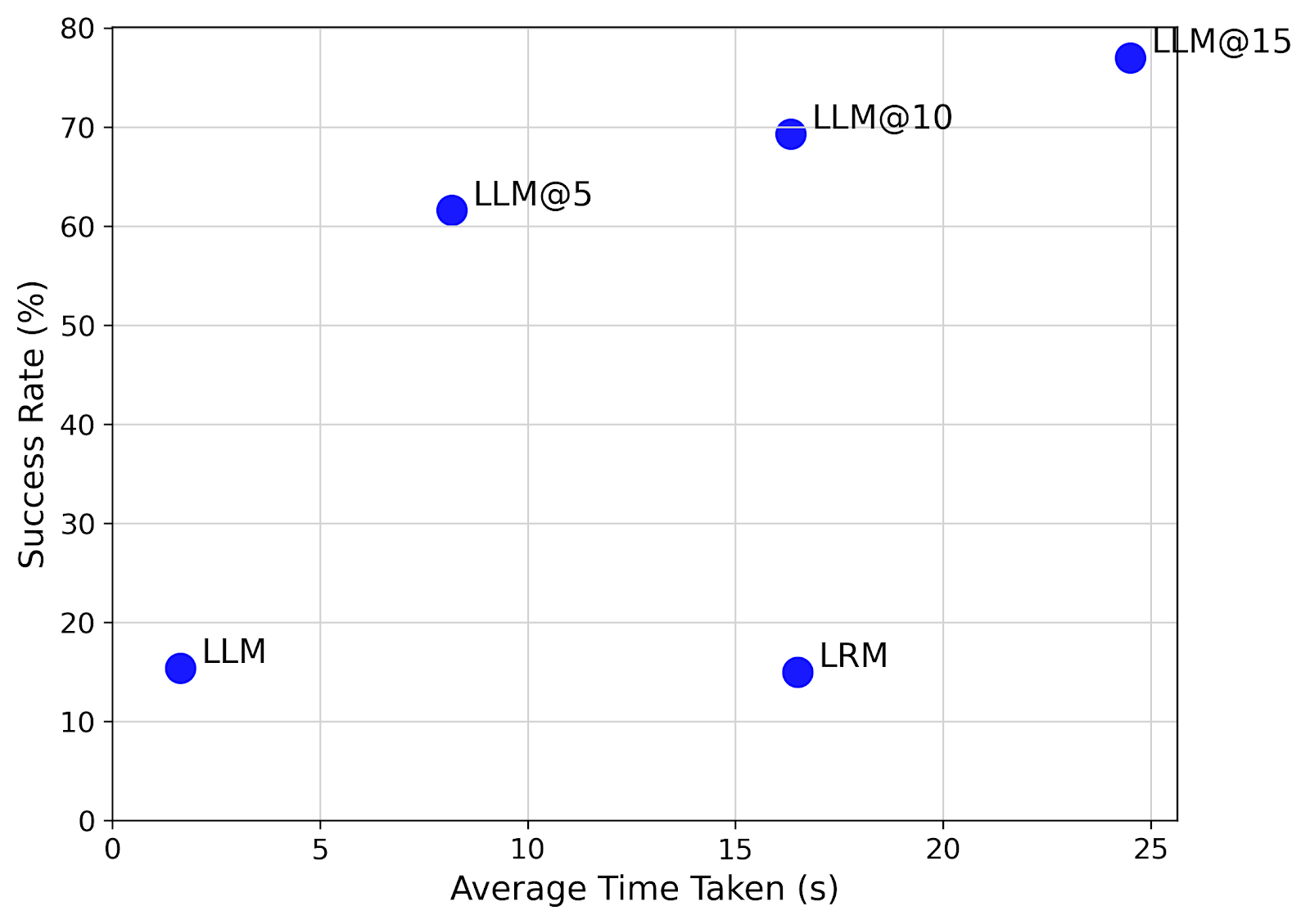}
    \caption{RQ1 - Model Combination 4, Size 10}
    \label{fig:rq1_mc4_10}
\end{figure}

\begin{figure}[h!]
    \centering
    \includegraphics[width=0.95\linewidth]{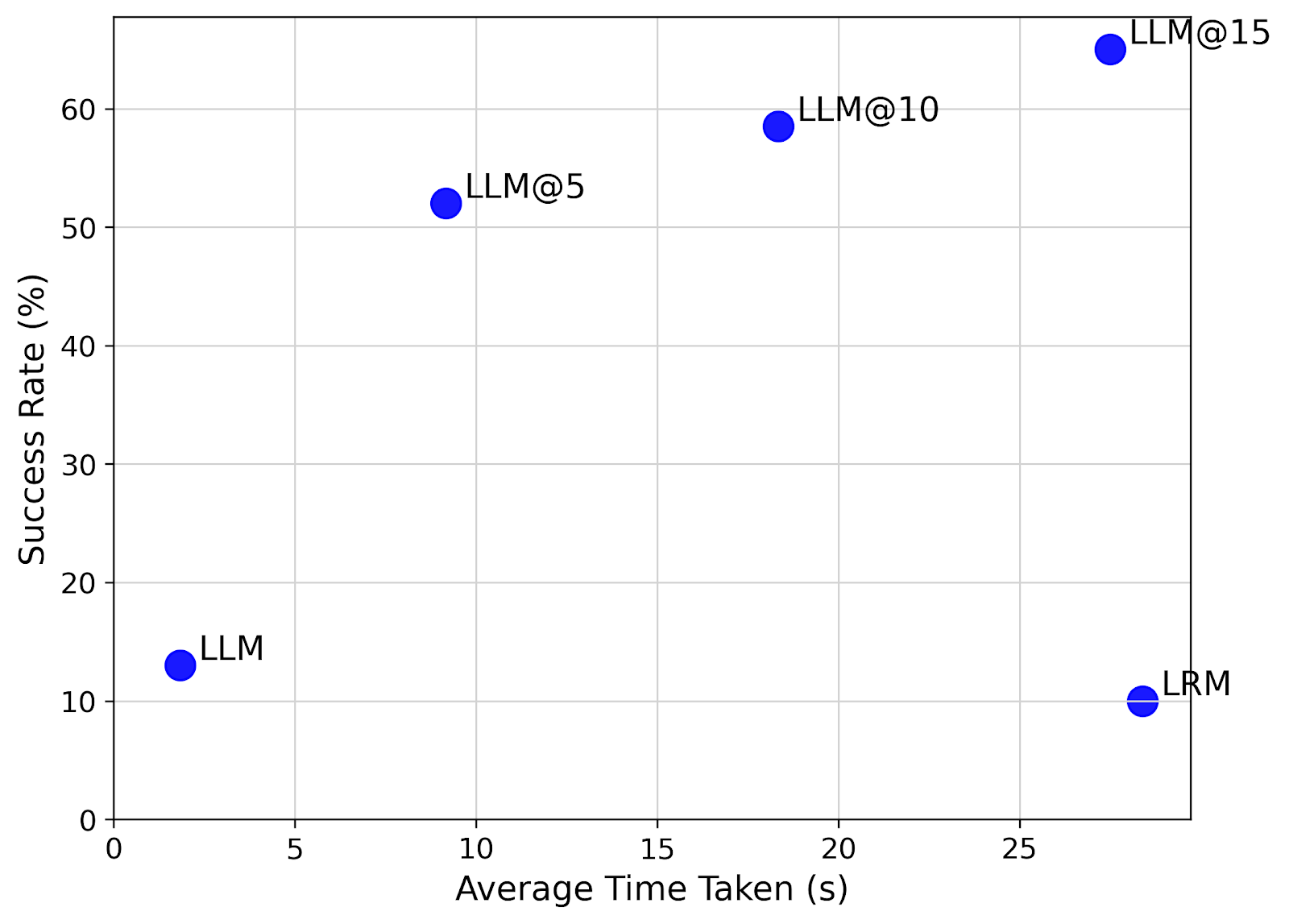}
    \caption{RQ1 - Model Combination 4, Size 15}
    \label{fig:rq1_mc4_15}
\end{figure}

\begin{figure}[h!]
    \centering
    \includegraphics[width=0.95\linewidth]{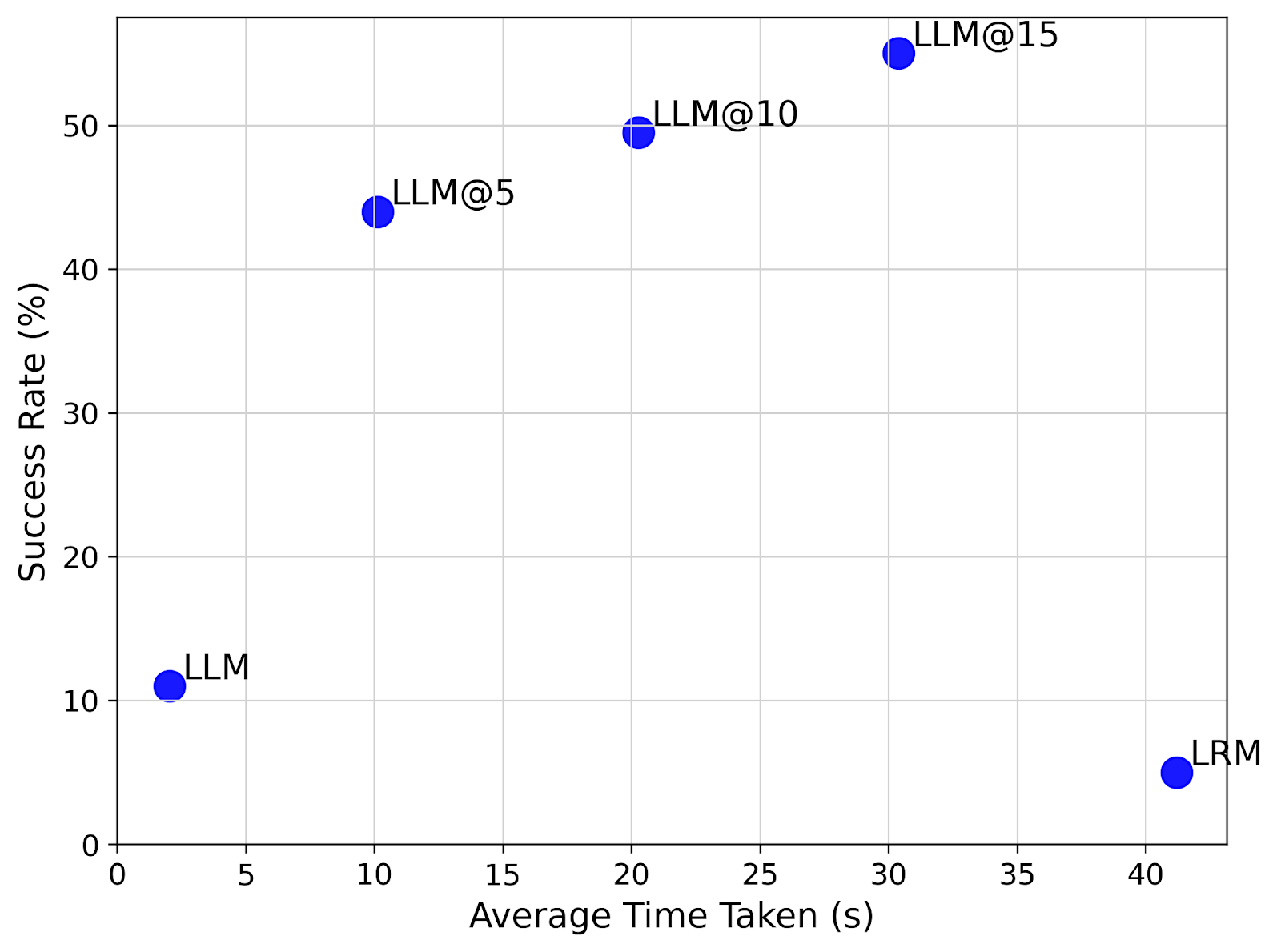}
    \caption{RQ1 - Model Combination 4, Size 20}
    \label{fig:rq1_mc4_20}
\end{figure}

\begin{figure}[h!]
    \centering
    \includegraphics[width=0.95\linewidth]{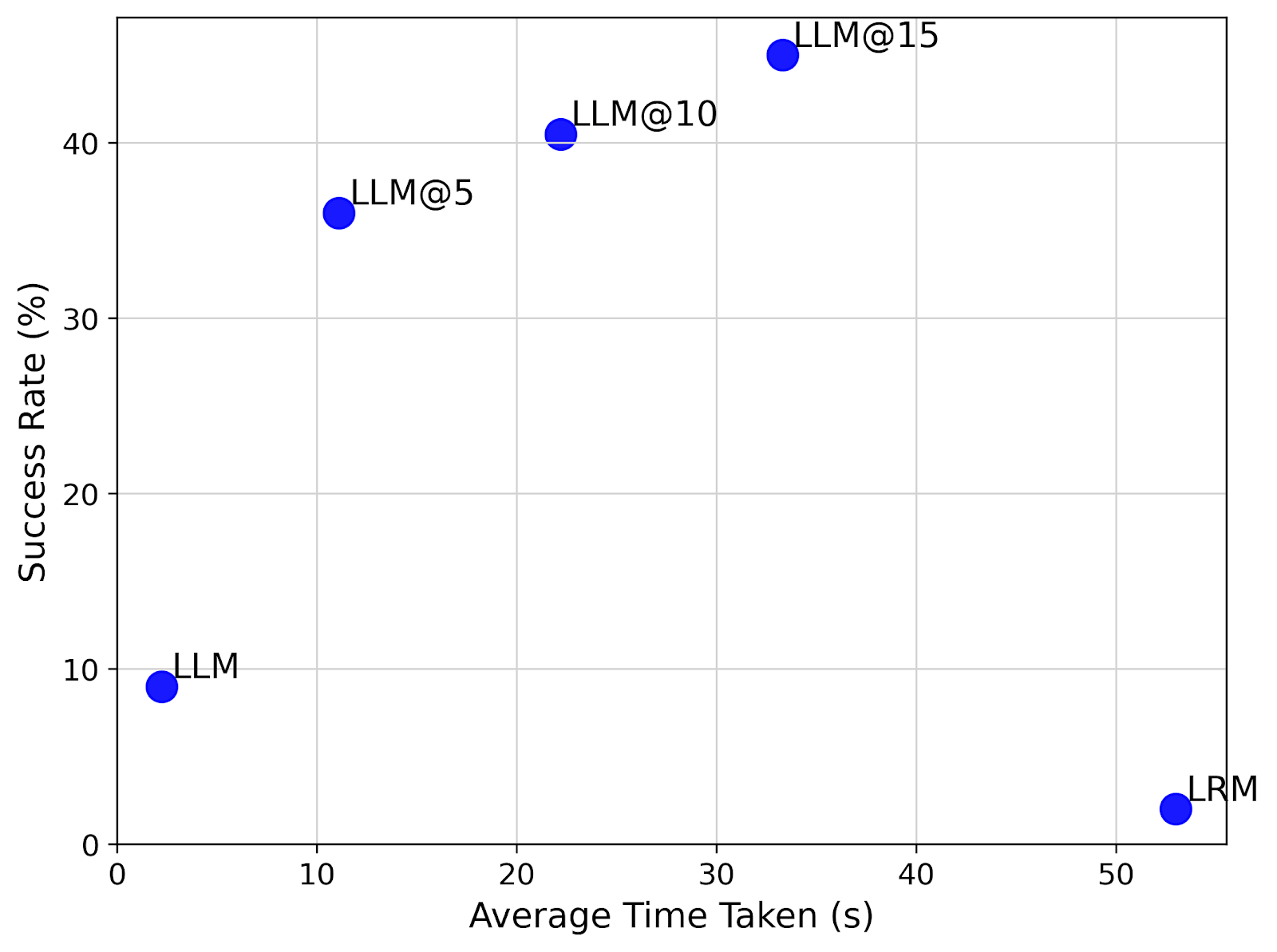}
    \caption{RQ1 - Model Combination 4, Size 25}
    \label{fig:rq1_mc4_25}
\end{figure}

\paragraph{Description}
This final set of figures (Figures \ref{fig:rq1_mc4_5} through \ref{fig:rq1_mc4_25}) contrasts the performance of Llama 3.1, a powerful LLM, in a feedback loop against the DeepSeek R1 8B LRM. The results reinforce the core finding: the iterative feedback mechanism is critical for high performance. Even though Llama 3.1 requires slightly more time, it achieves a high success rate that is vastly superior to the standalone LRM, especially as graph size increases. The DeepSeek R1 8B LRM's performance degrades significantly on larger problems, while the iterative Llama 3.1 maintains strong performance, proving the general applicability of the feedback-driven method with different state-of-the-art LLMs.

\paragraph{Summary Finding}
The performance benefits of the iterative feedback architecture are model-agnostic, enabling another powerful LLM like Llama 3.1 to significantly outperform a standalone LRM on complex reasoning tasks.


\clearpage

\subsection{RQ2: How do the type of feedback and the format of episodic memory affect a feedback-driven LLM?}
The following figures evaluate the impact of different feedback and memory strategies on LLM performance. The configurations are Multi-Line Feedback (MLF) and Single-Line Feedback (SLF), each paired with Minimal Episodic Memory (MEM) or Extended Episodic Memory (EEM).

\subsubsection{LLM: Granite 3.3B (without thinking)}
\begin{figure}[h!]
    \centering
    \includegraphics[width=0.95\linewidth]{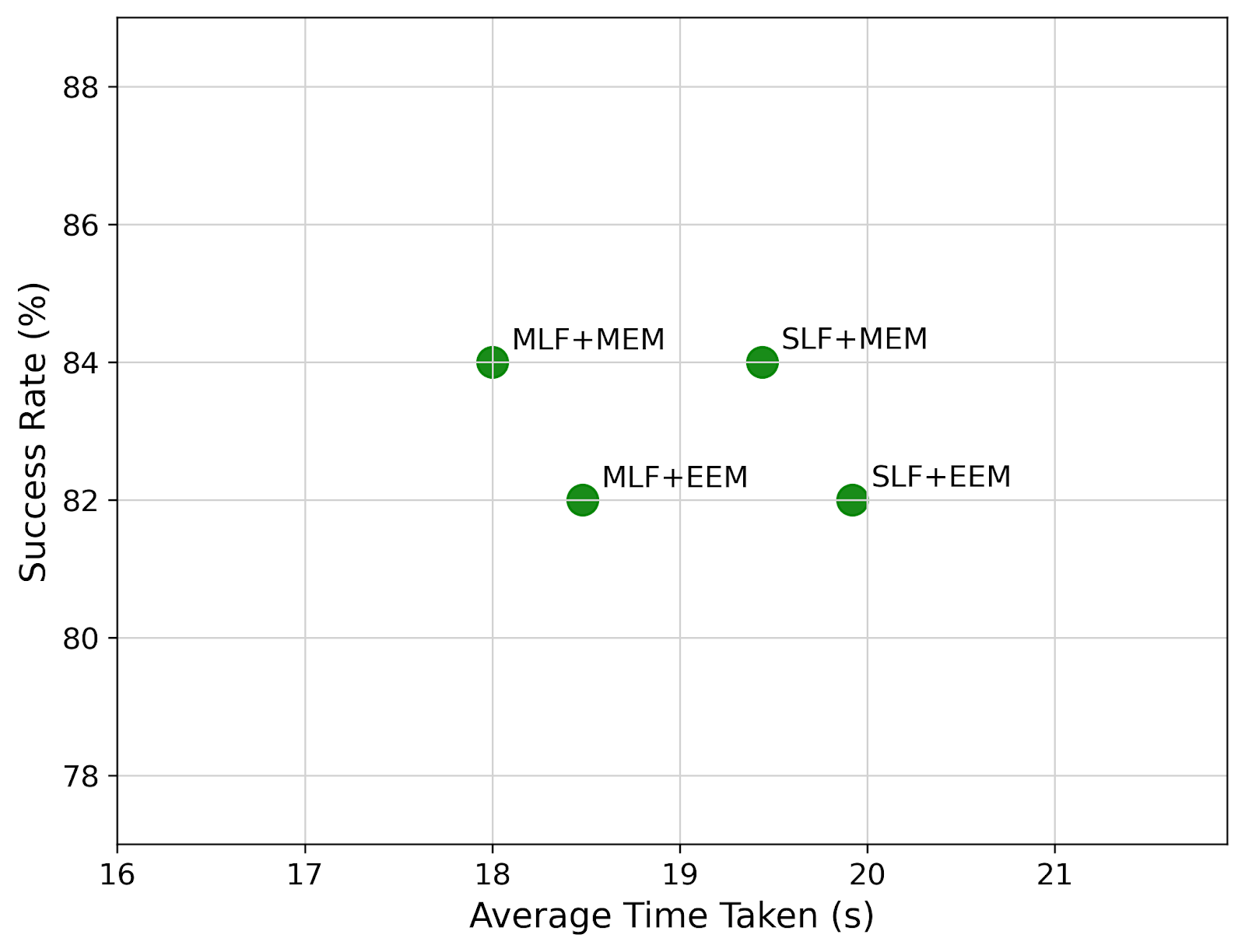}
    \caption{RQ2 - LLM: Granite 3.3B, Size 5}
    \label{fig:rq2_mc1_5}
\end{figure}

\begin{figure}[h!]
    \centering
    \includegraphics[width=0.95\linewidth]{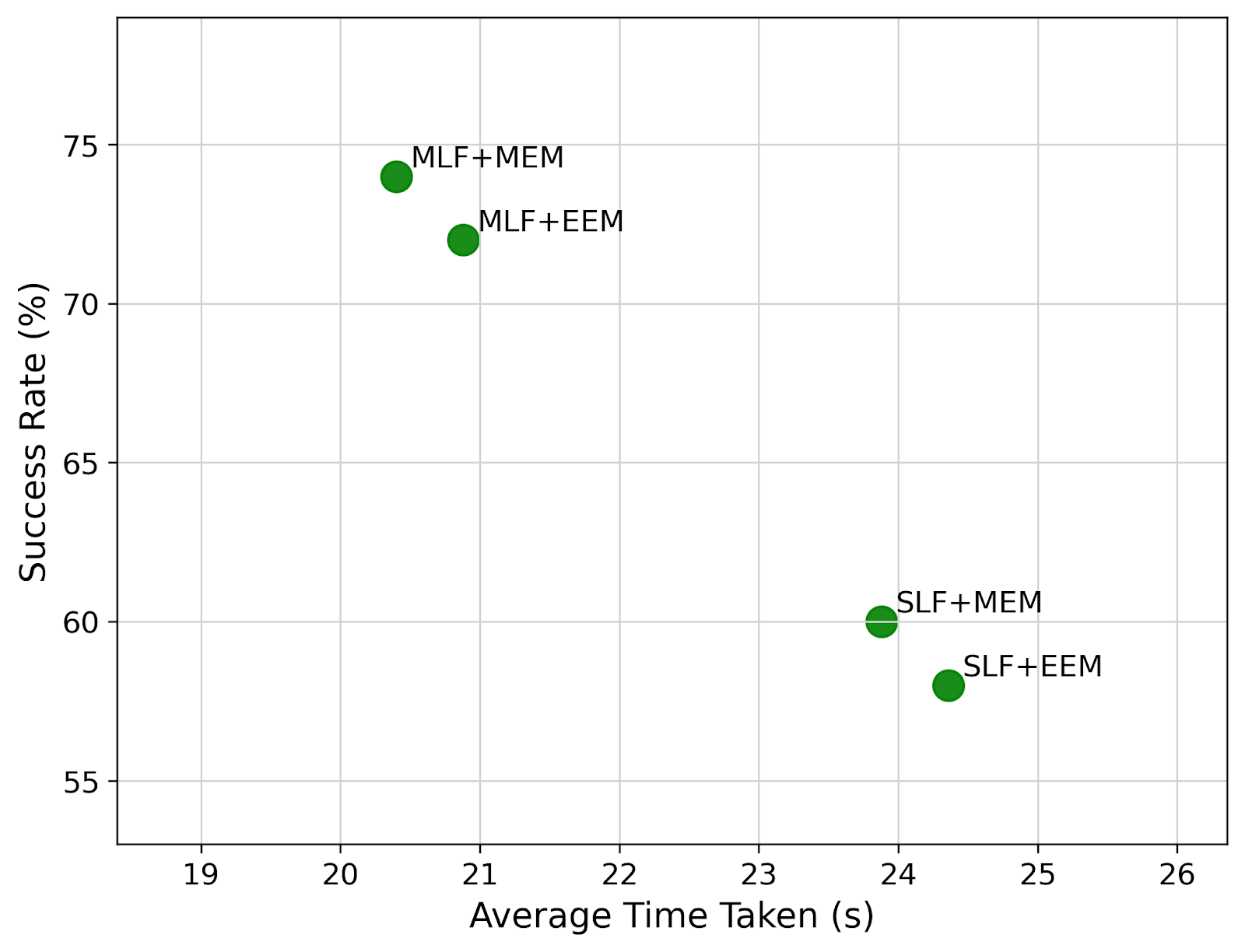}
    \caption{RQ2 - LLM: Granite 3.3B, Size 10}
    \label{fig:rq2_mc1_10}
\end{figure}

\begin{figure}[h!]
    \centering
    \includegraphics[width=0.95\linewidth]{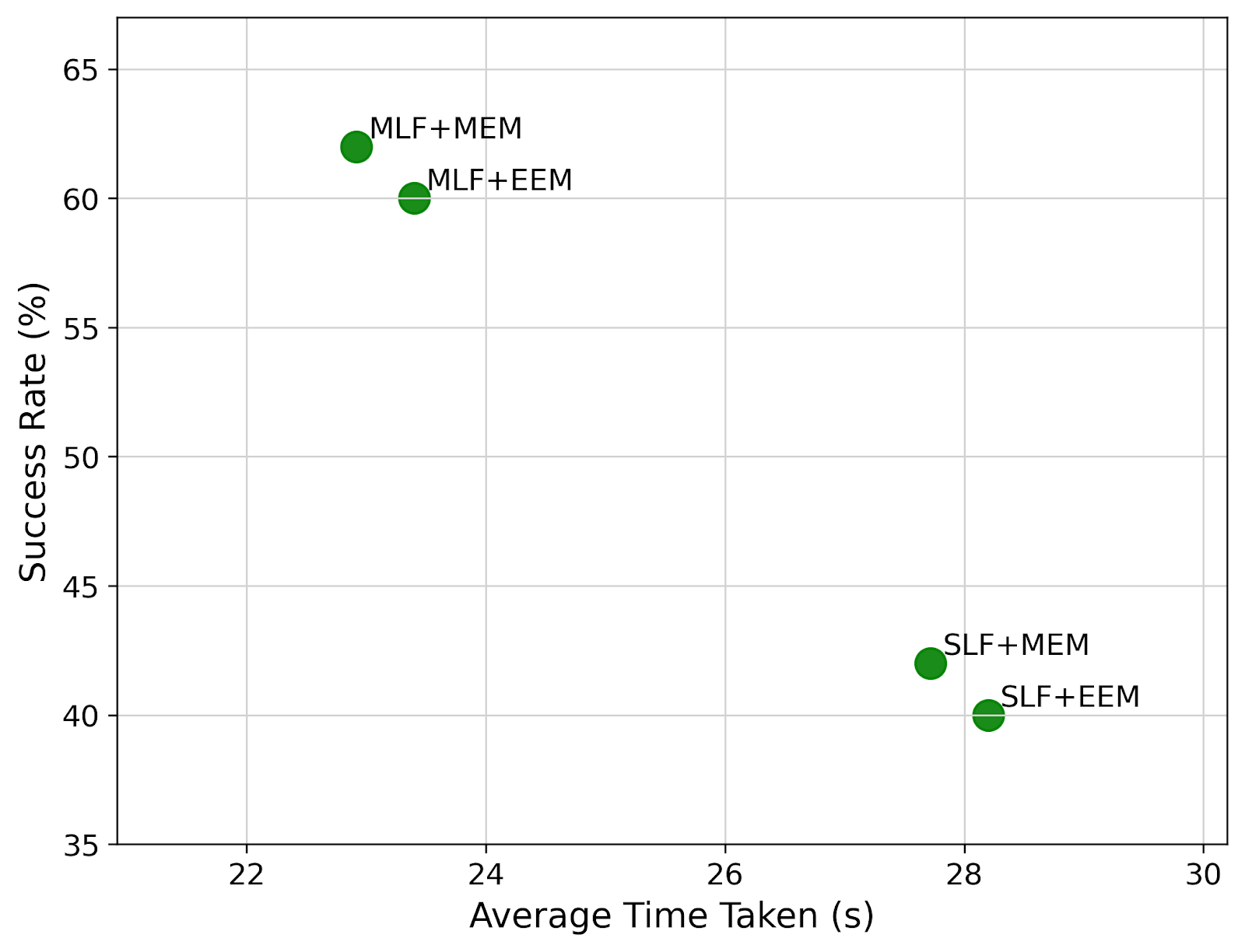}
    \caption{RQ2 - LLM: Granite 3.3B, Size 15}
    \label{fig:rq2_mc1_15}
\end{figure}

\begin{figure}[h!]
    \centering
    \includegraphics[width=0.95\linewidth]{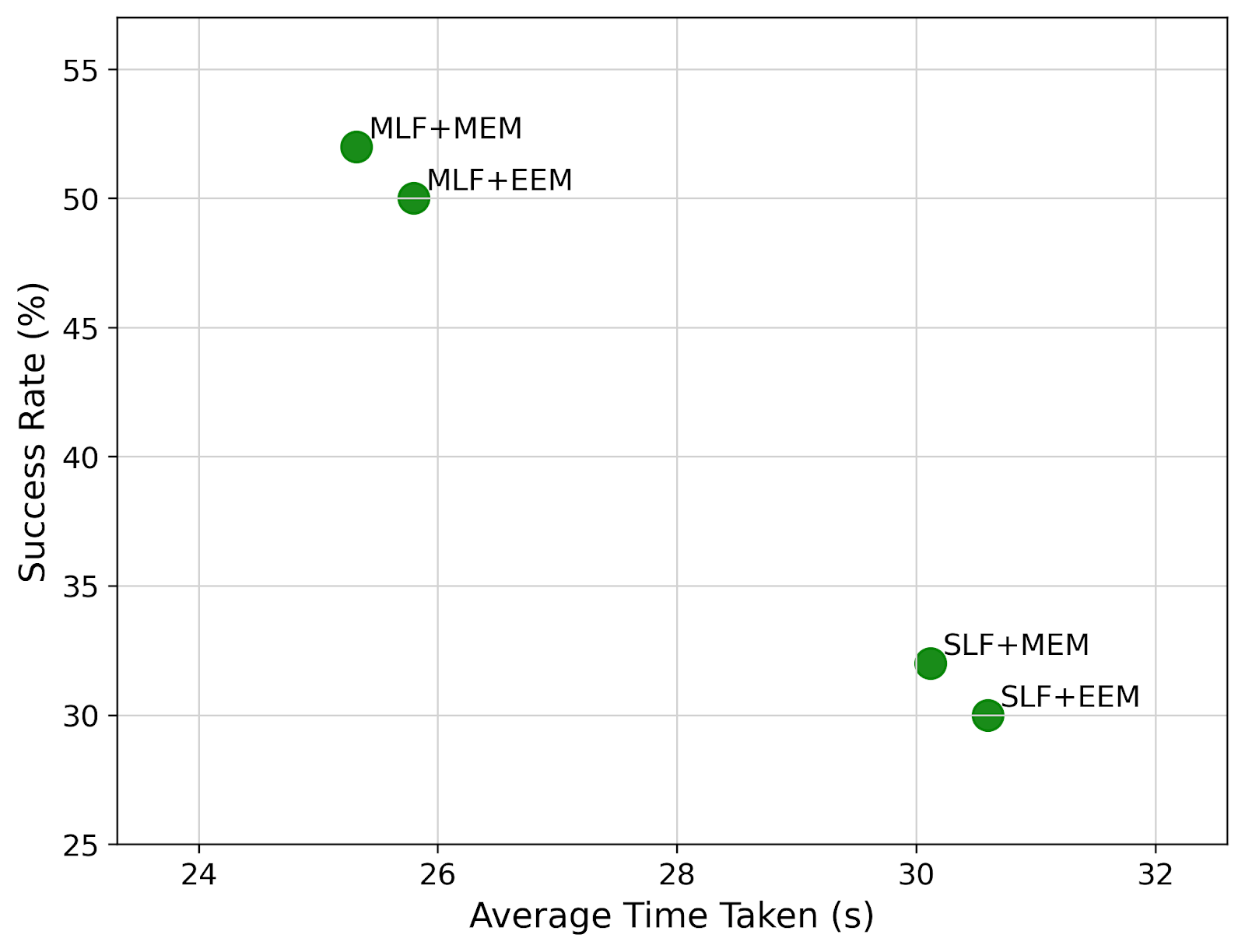}
    \caption{RQ2 - LLM: Granite 3.3B, Size 20}
    \label{fig:rq2_mc1_20}
\end{figure}

\paragraph{Description}
This set of figures (Figures \ref{fig:rq2_mc1_5} through \ref{fig:rq2_mc1_20}) evaluates the impact of different feedback and memory strategies on the performance of the Granite 3.3B LLM. A clear and consistent trend emerges: configurations using Multi-Line Feedback (MLF) consistently outperform those using Single-Line Feedback (SLF) in success rate, often in less time. Within both groups, Minimal Episodic Memory (MEM) generally provides a slight edge over Extended Episodic Memory (EEM). The most effective configuration across all problem sizes is MLF+MEM.

\paragraph{Summary Finding}
For the Granite 3.3B model, detailed Multi-Line Feedback combined with a concise Minimal Episodic Memory (MLF+MEM) provides the optimal balance of high success rate and low computational time.


\clearpage

\subsubsection{LLM: Llama 3.1}
\begin{figure}[h!]
    \centering
    \includegraphics[width=0.95\linewidth]{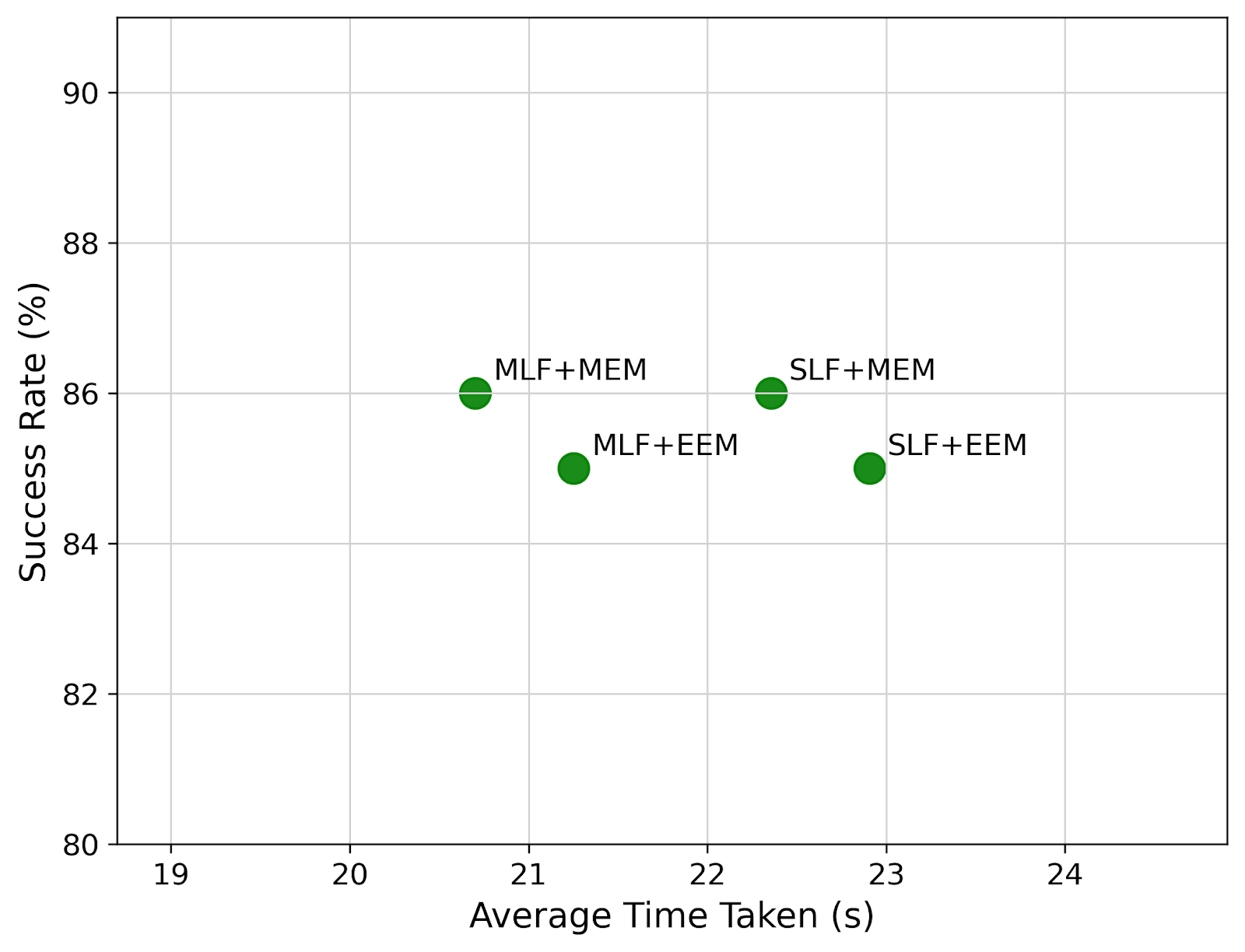}
    \caption{RQ2 - LLM: Llama 3.1, Size 5}
    \label{fig:rq2_mc2_5}
\end{figure}

\begin{figure}[h!]
    \centering
    \includegraphics[width=0.95\linewidth]{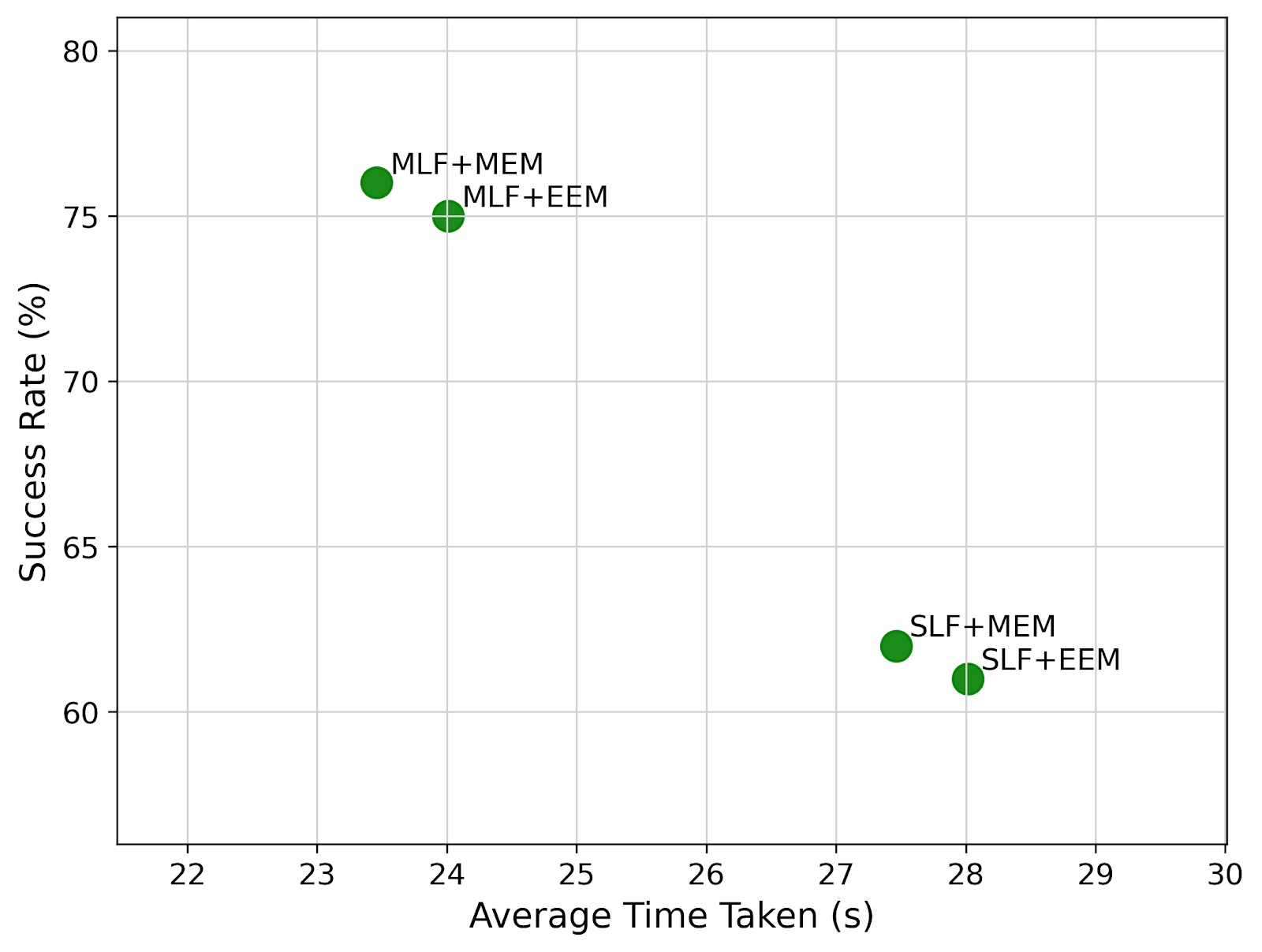}
    \caption{RQ2 - LLM: Llama 3.1, Size 10}
    \label{fig:rq2_mc2_10}
\end{figure}

\begin{figure}[h!]
    \centering
    \includegraphics[width=0.95\linewidth]{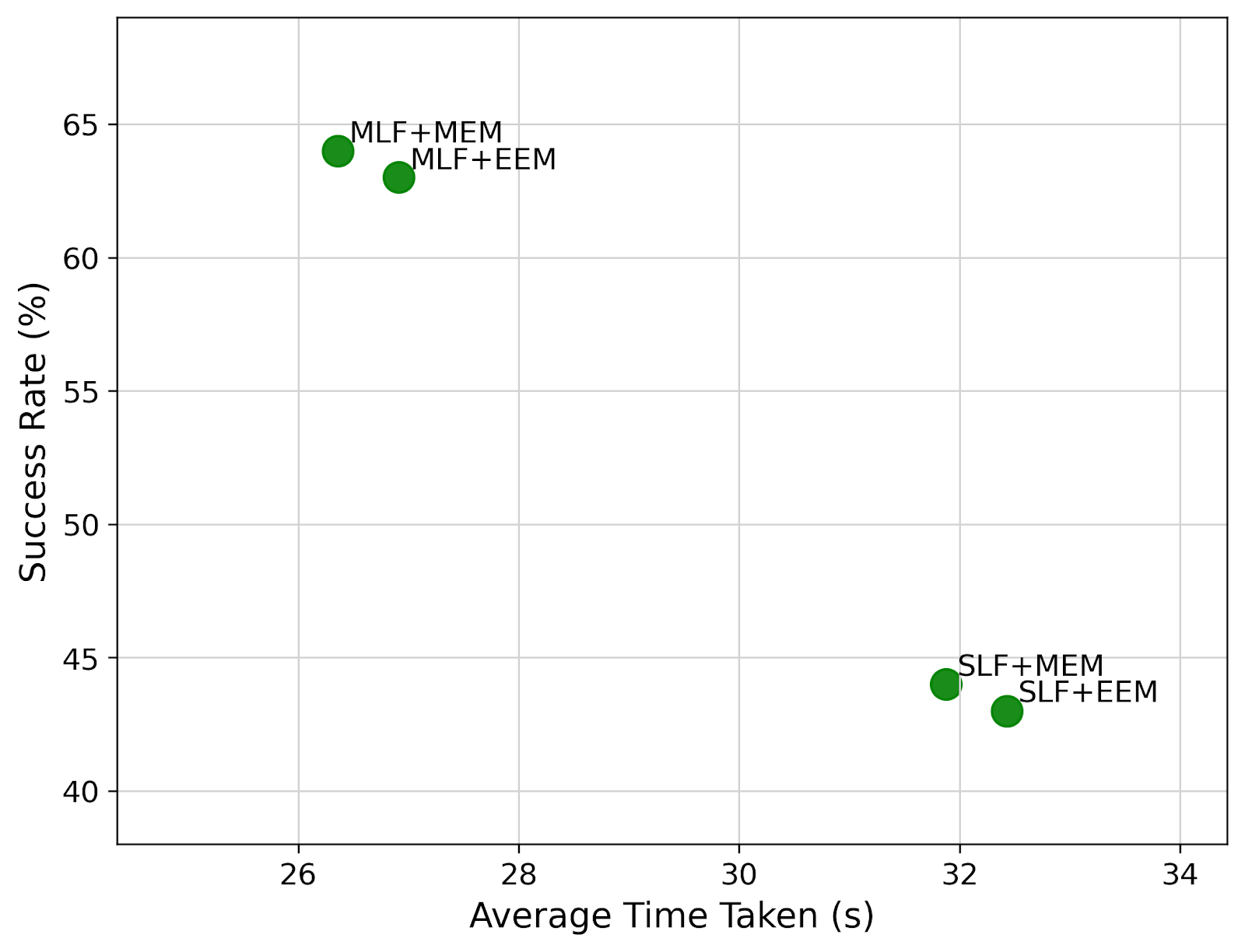}
    \caption{RQ2 - LLM: Llama 3.1, Size 15}
    \label{fig:rq2_mc2_15}
\end{figure}

\begin{figure}[h!]
    \centering
    \includegraphics[width=0.95\linewidth]{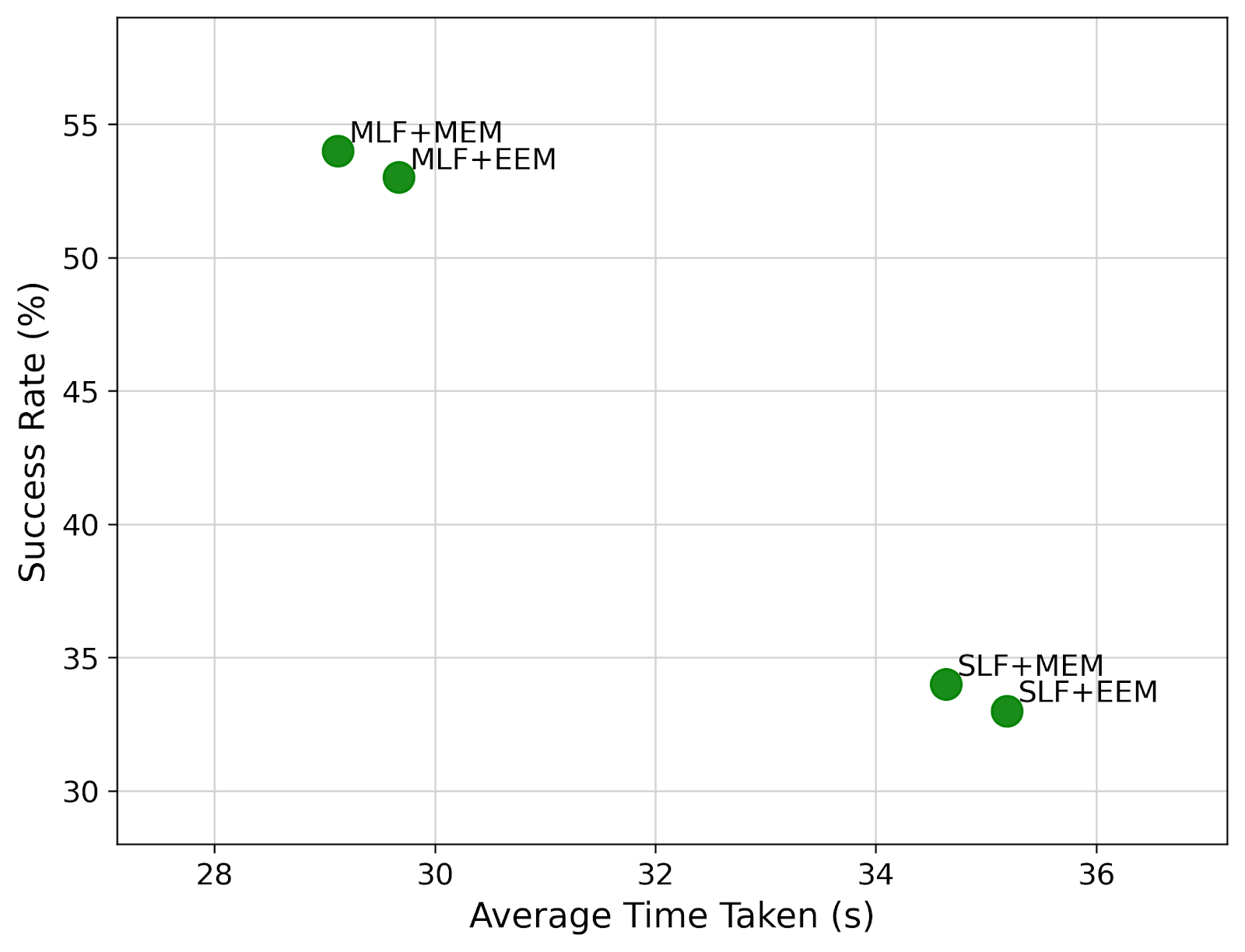}
    \caption{RQ2 - LLM: Llama 3.1, Size 20}
    \label{fig:rq2_mc2_20}
\end{figure}

\begin{figure}[h!]
    \centering
    \includegraphics[width=0.95\linewidth]{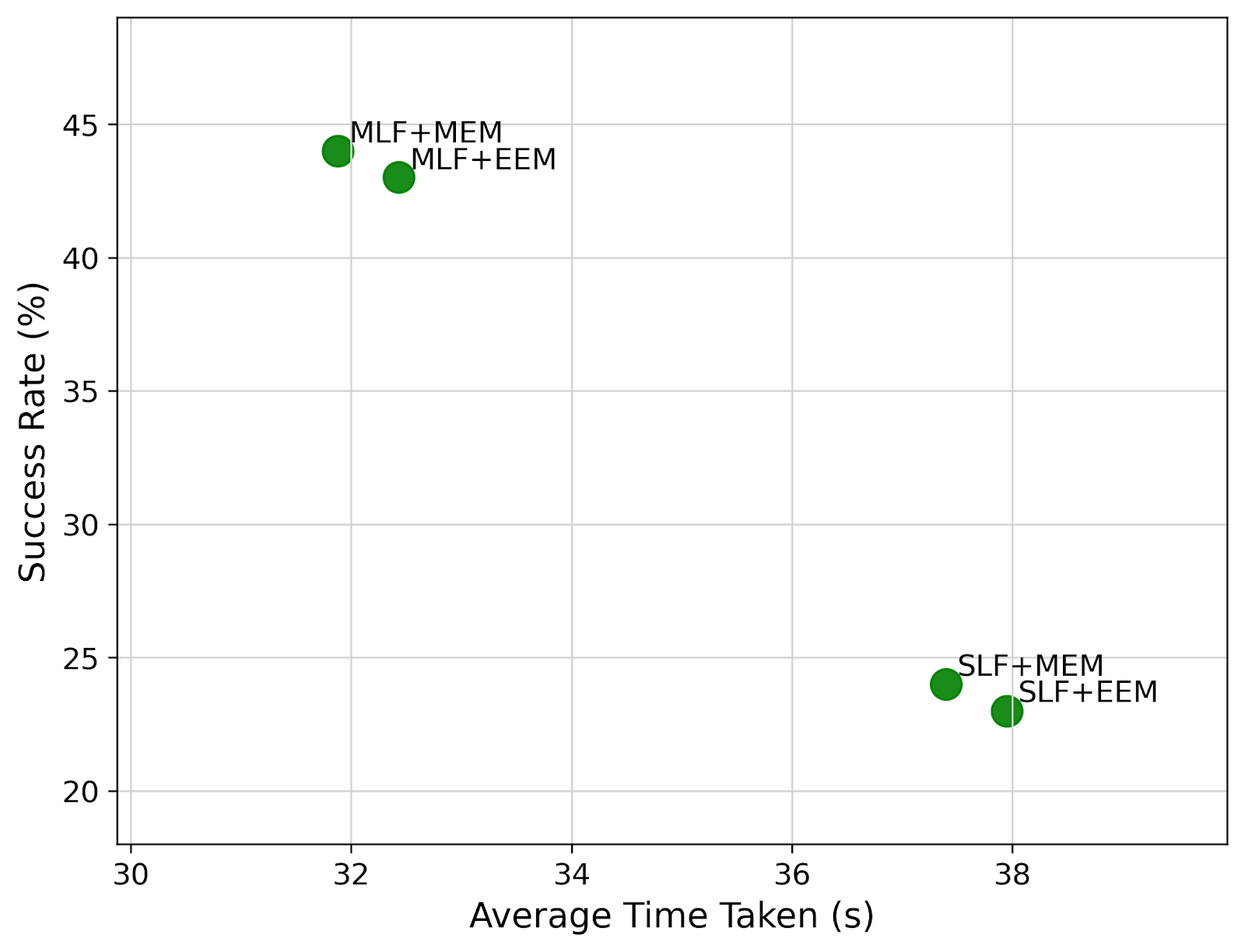}
    \caption{RQ2 - LLM: Llama 3.1, Size 25}
    \label{fig:rq2_mc2_25}
\end{figure}

\paragraph{Description}
This series of plots (Figures \ref{fig:rq2_mc2_5} through \ref{fig:rq2_mc2_25}) shows the performance of the four metacognitive configurations for the Llama 3.1 LLM. The results demonstrate that the relative effectiveness of the different feedback and memory strategies is consistent across different LLMs. As with the Granite model, configurations with Multi-Line Feedback (MLF) achieve higher success rates than their Single-Line Feedback (SLF) counterparts. The MLF+MEM combination again emerges as the most efficient.

\paragraph{Summary Finding}
The principle that detailed, multi-line feedback with minimal episodic memory (MLF+MEM) provides the best performance holds true for the Llama 3.1 model, indicating that the observed trends for feedback and memory types are robust across different underlying LLMs.


\clearpage

\subsection{RQ3: Can the information gathered by SOFAI-LM, when used iteratively with an LLM, enhance the performance of an LRM?}
The following figures analyze the effectiveness of three different LRM prompting strategies within the SOFAI-LM framework: Problem-Only (PO), Best Attempt (BA), and Full History (FH).

\subsubsection{Pipeline: Granite 3.3B \(\rightarrow\) DeepSeek R1 8B}
\begin{figure}[h!]
    \centering
    \includegraphics[width=0.95\linewidth]{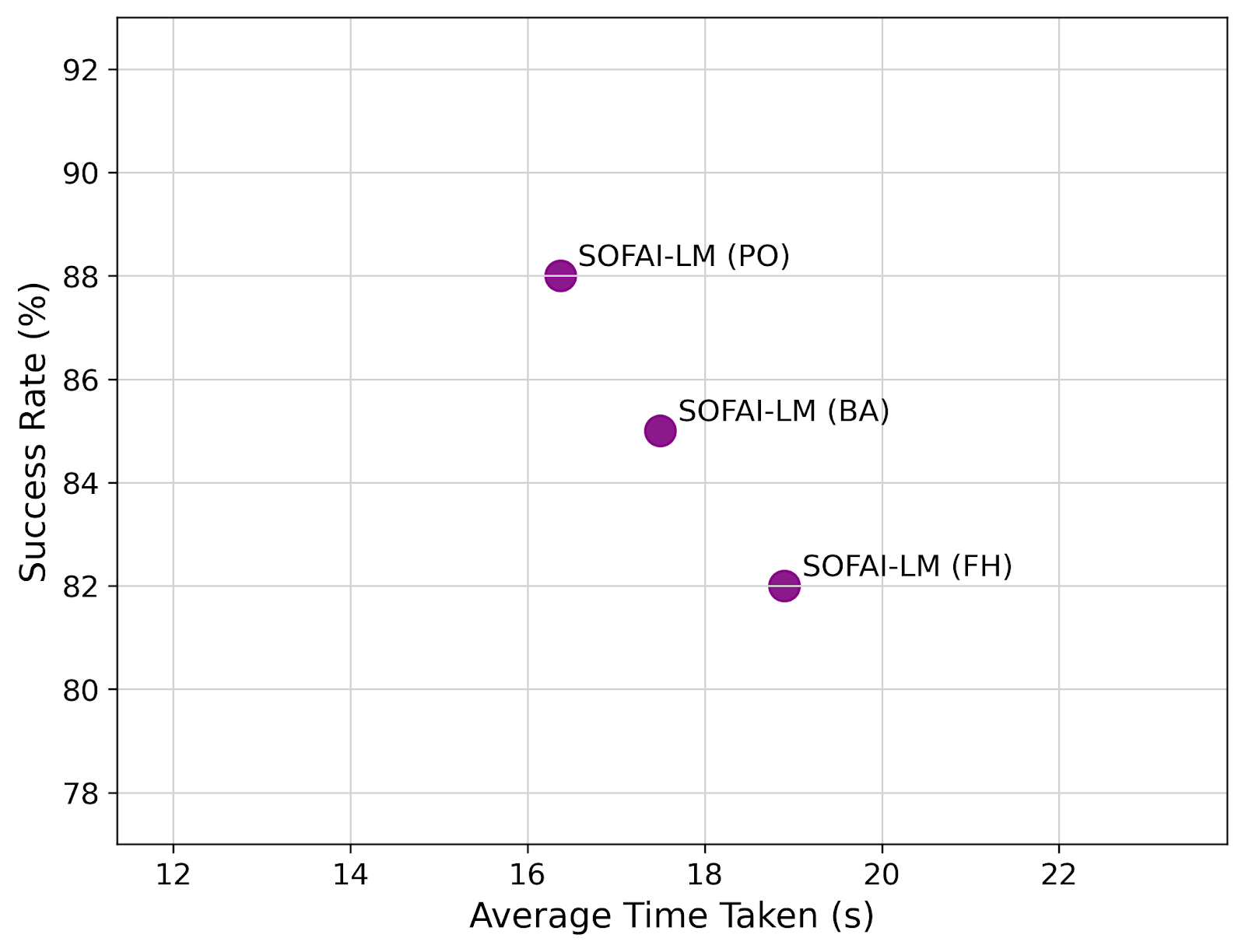}
    \caption{RQ3 - Pipeline: Granite 3.3B \(\rightarrow\) DeepSeek R1 8B, Size 5}
    \label{fig:rq3_mc1_5}
\end{figure}

\begin{figure}[h!]
    \centering
    \includegraphics[width=0.95\linewidth]{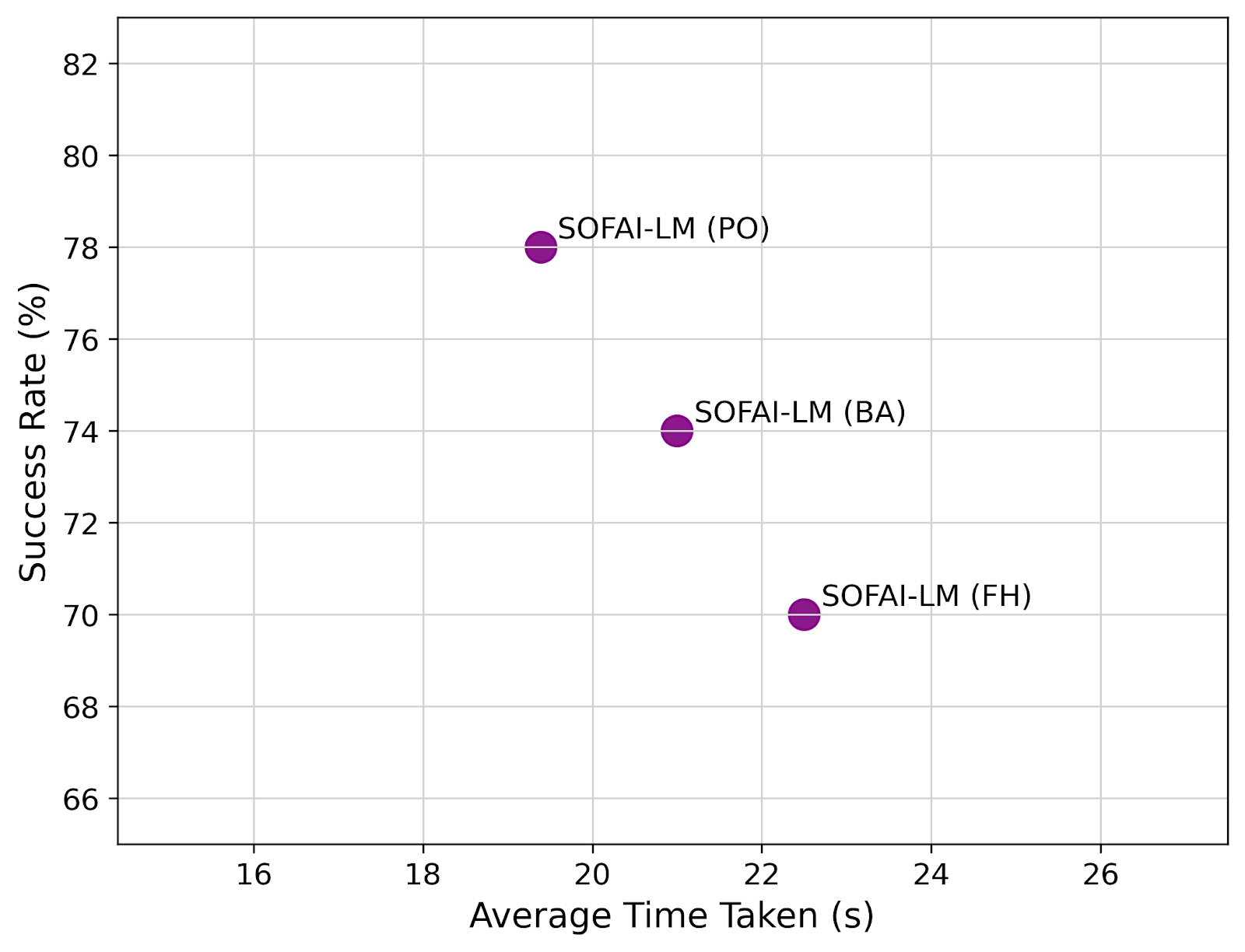}
    \caption{RQ3 - Pipeline: Granite 3.3B \(\rightarrow\) DeepSeek R1 8B, Size 10}
    \label{fig:rq3_mc1_10}
\end{figure}

\begin{figure}[h!]
    \centering
    \includegraphics[width=0.95\linewidth]{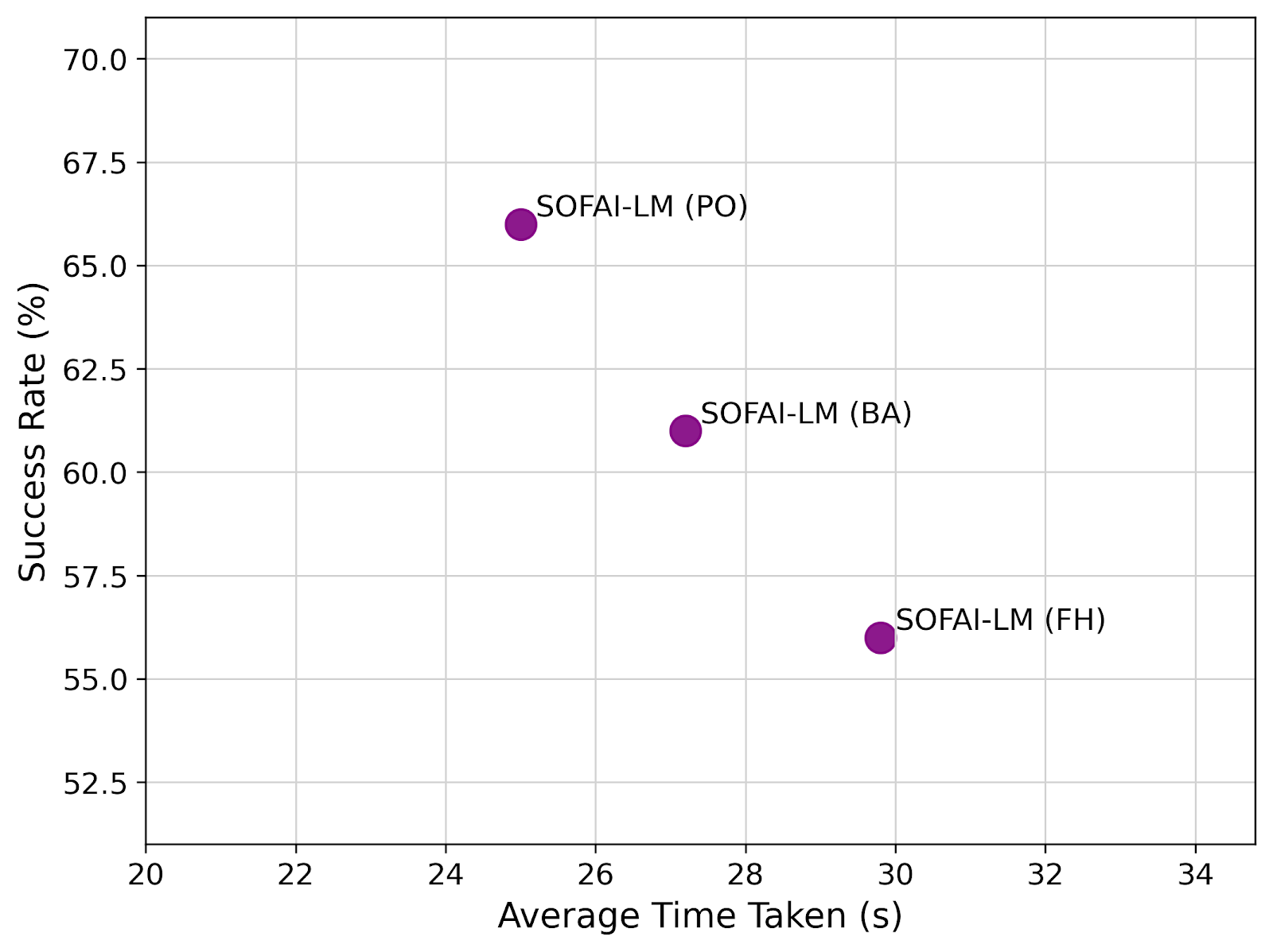}
    \caption{RQ3 - Pipeline: Granite 3.3B \(\rightarrow\) DeepSeek R1 8B, Size 15}
    \label{fig:rq3_mc1_15}
\end{figure}

\begin{figure}[h!]
    \centering
    \includegraphics[width=0.95\linewidth]{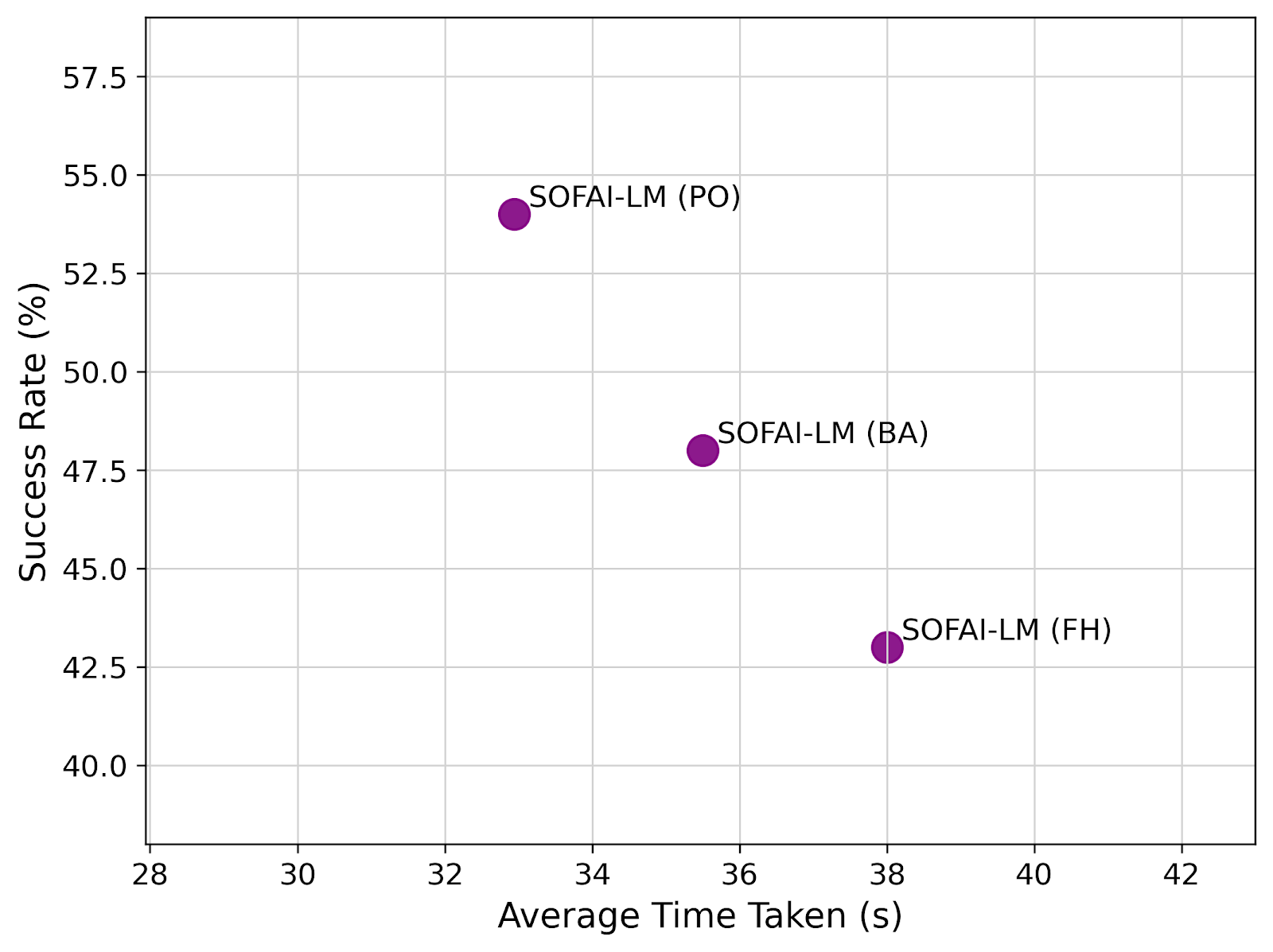}
    \caption{RQ3 - Pipeline: Granite 3.3B \(\rightarrow\) DeepSeek R1 8B, Size 20}
    \label{fig:rq3_mc1_20}
\end{figure}

\paragraph{Description}
This series of plots (Figures \ref{fig:rq3_mc1_5} through \ref{fig:rq3_mc1_20}) analyzes the LRM prompting strategies where Granite 3.3B is the LLM and DeepSeek R1 8B is the LRM. The `Problem-Only (PO)` strategy consistently delivers the highest success rate in the shortest amount of time. Providing the LRM with the `Best Attempt (BA)` or the `Full History (FH)` leads to a degradation in performance, suggesting that for a globally constrained problem like graph coloring, the incorrect partial solutions act as noise.

\paragraph{Summary Finding}
For graph coloring tasks, providing the LRM with only the original problem instance (PO) is the most effective strategy, as additional context from the LLM's attempts degrades both success rate and efficiency.


\clearpage

\subsubsection{Pipeline: Granite 3.3B \(\rightarrow\) Granite 3.3B}
\begin{figure}[h!]
    \centering
    \includegraphics[width=0.95\linewidth]{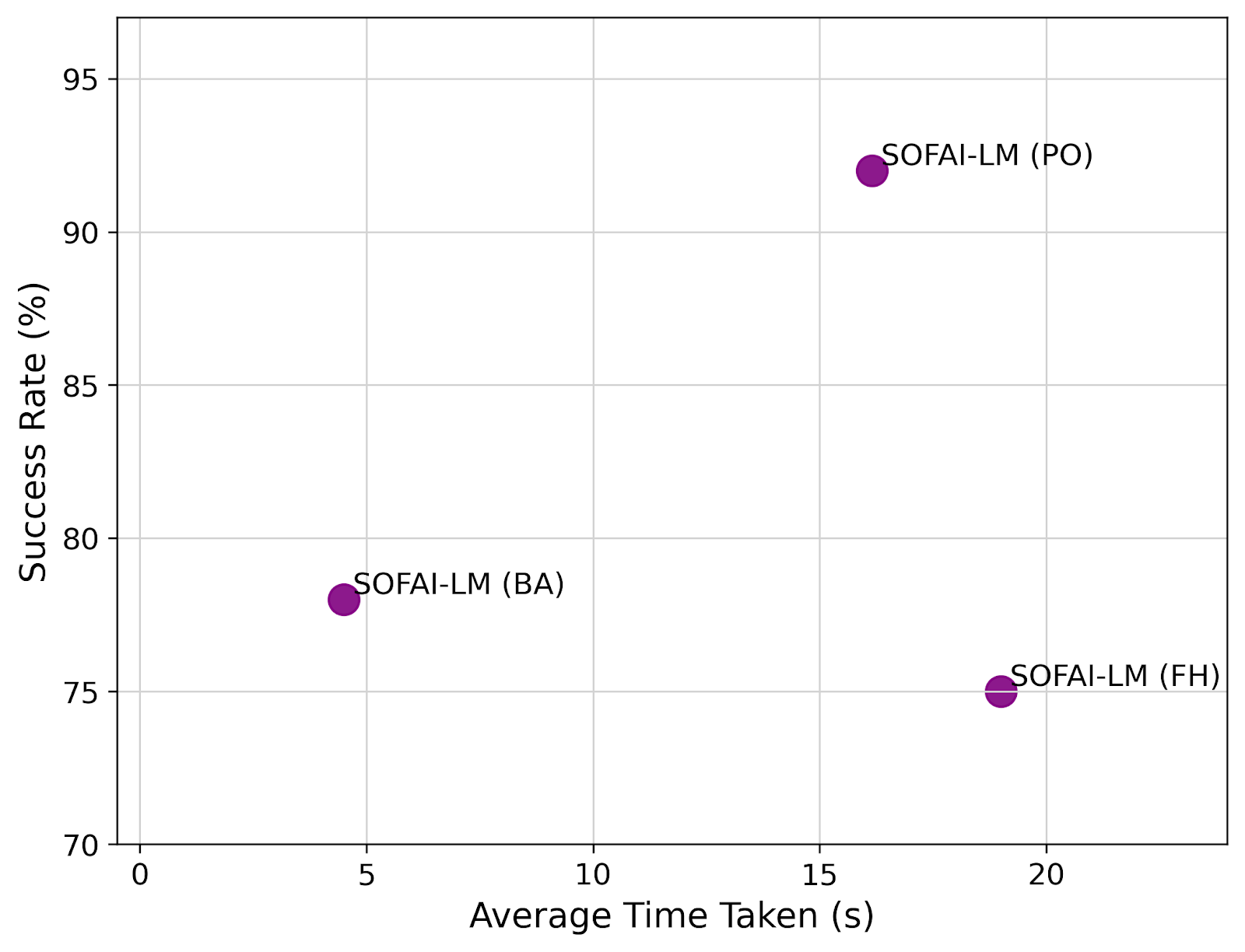}
    \caption{RQ3 - Pipeline: Granite 3.3B \(\rightarrow\) Granite 3.3B, Size 5}
    \label{fig:rq3_mc2_5}
\end{figure}

\begin{figure}[h!]
    \centering
    \includegraphics[width=0.95\linewidth]{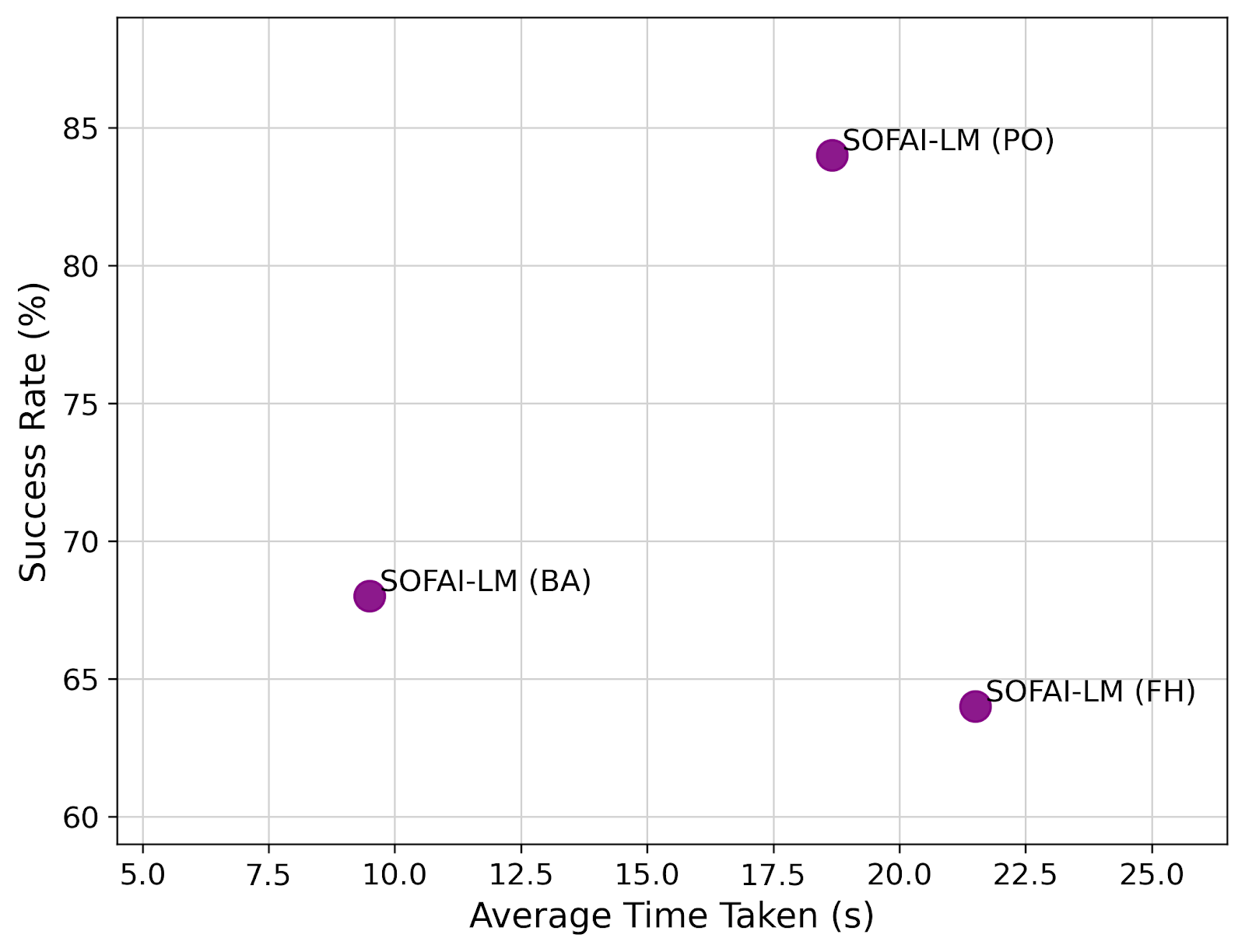}
    \caption{RQ3 - Pipeline: Granite 3.3B \(\rightarrow\) Granite 3.3B, Size 10}
    \label{fig:rq3_mc2_10}
\end{figure}

\begin{figure}[h!]
    \centering
    \includegraphics[width=0.95\linewidth]{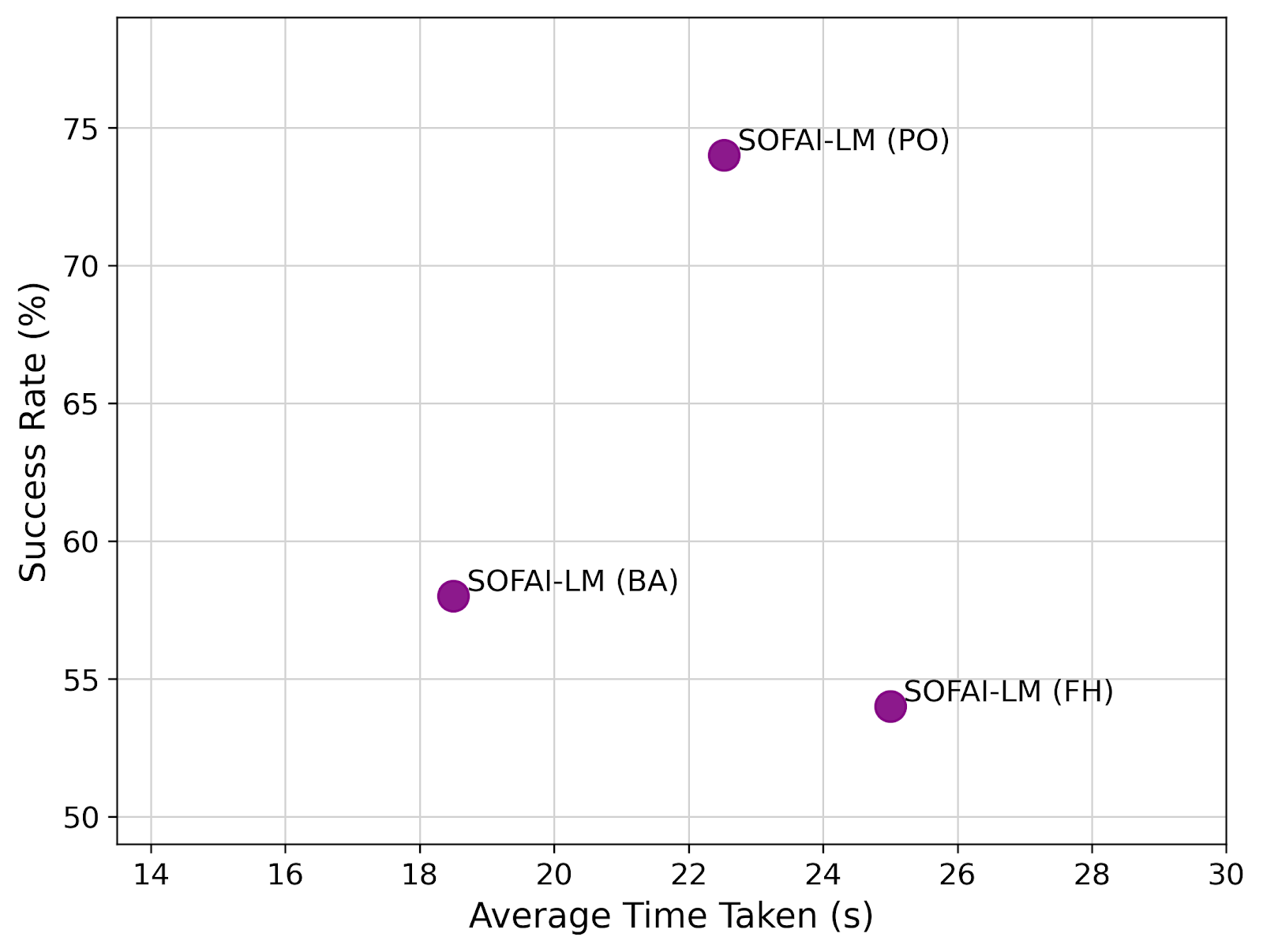}
    \caption{RQ3 - Pipeline: Granite 3.3B \(\rightarrow\) Granite 3.3B, Size 15}
    \label{fig:rq3_mc2_15}
\end{figure}

\begin{figure}[h!]
    \centering
    \includegraphics[width=0.95\linewidth]{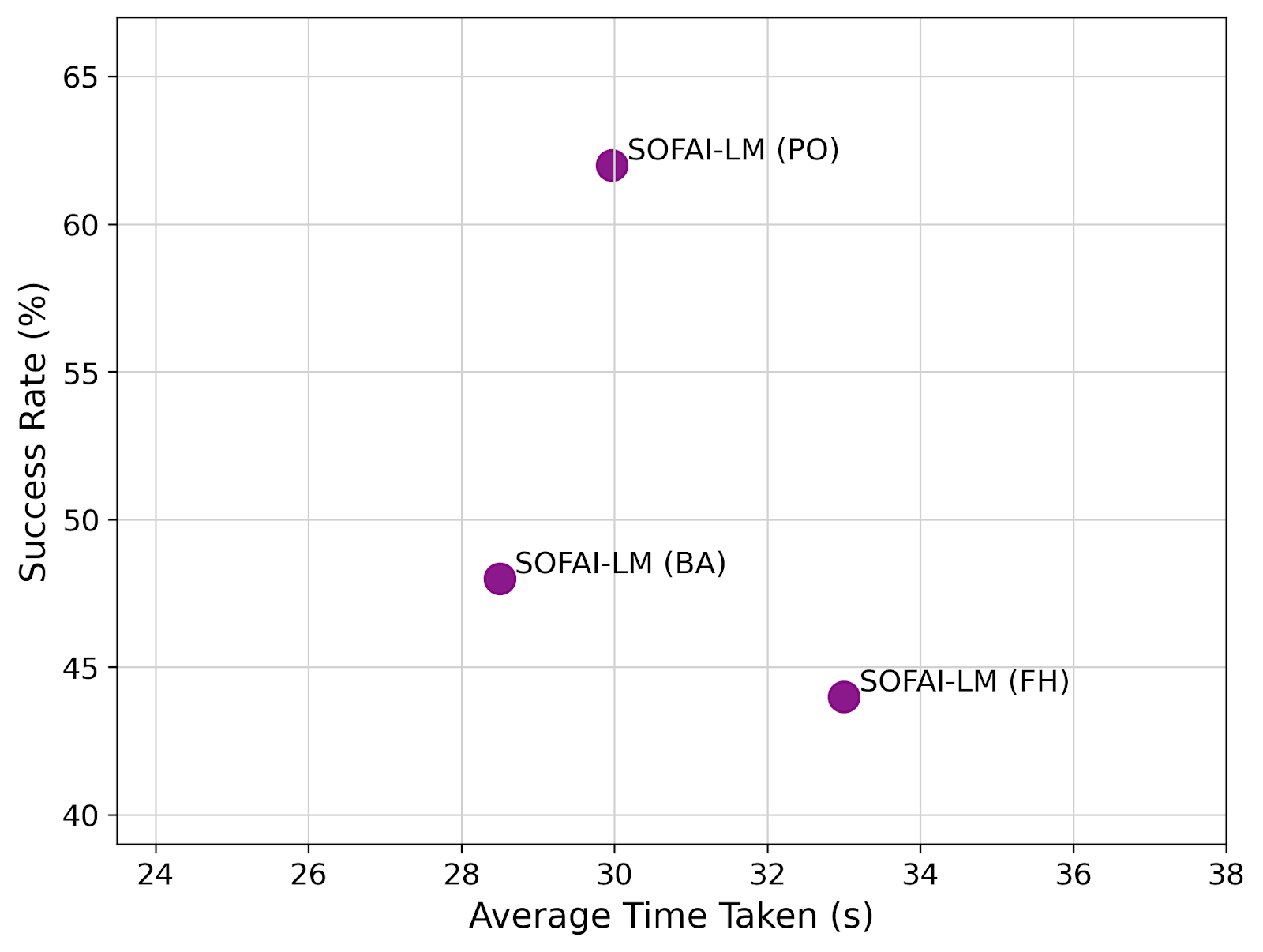}
    \caption{RQ3 - Pipeline: Granite 3.3B \(\rightarrow\) Granite 3.3B, Size 20}
    \label{fig:rq3_mc2_20}
\end{figure}

\begin{figure}[h!]
    \centering
    \includegraphics[width=0.95\linewidth]{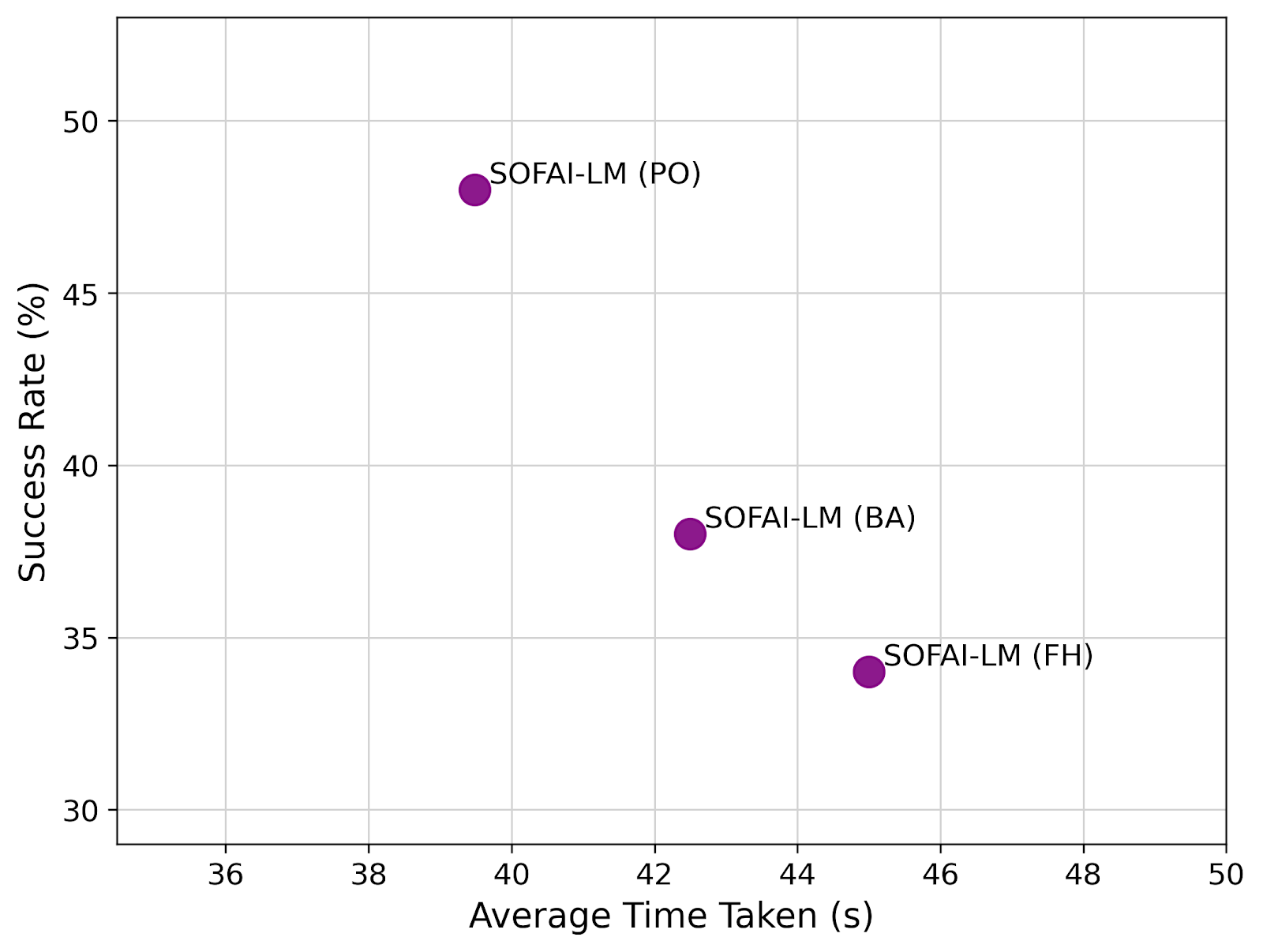}
    \caption{RQ3 - Pipeline: Granite 3.3B \(\rightarrow\) Granite 3.3B, Size 25}
    \label{fig:rq3_mc2_25}
\end{figure}

\paragraph{Description}
This set of plots (Figures \ref{fig:rq3_mc2_5} through \ref{fig:rq3_mc2_25}) examines the three prompting strategies when using Granite 3.3B for both the LLM and LRM roles. The results confirm the trend observed previously: the `Problem-Only (PO)` strategy is superior, achieving the highest success rate. This demonstrates that even when the LLM and LRM are the same underlying model, providing historical context of failed attempts hinders the model's ability to find a globally consistent solution.

\paragraph{Summary Finding}
When using a homogeneous model setup (Granite to Granite), a "clean slate" `Problem-Only` prompt for the LRM is still the most effective strategy, reinforcing that context from failed attempts is detrimental for this class of problem.


\clearpage

\subsubsection{Pipeline: Granite 3.3B \(\rightarrow\) Qwen 2.5 Pro}
\begin{figure}[h!]
    \centering
    \includegraphics[width=0.95\linewidth]{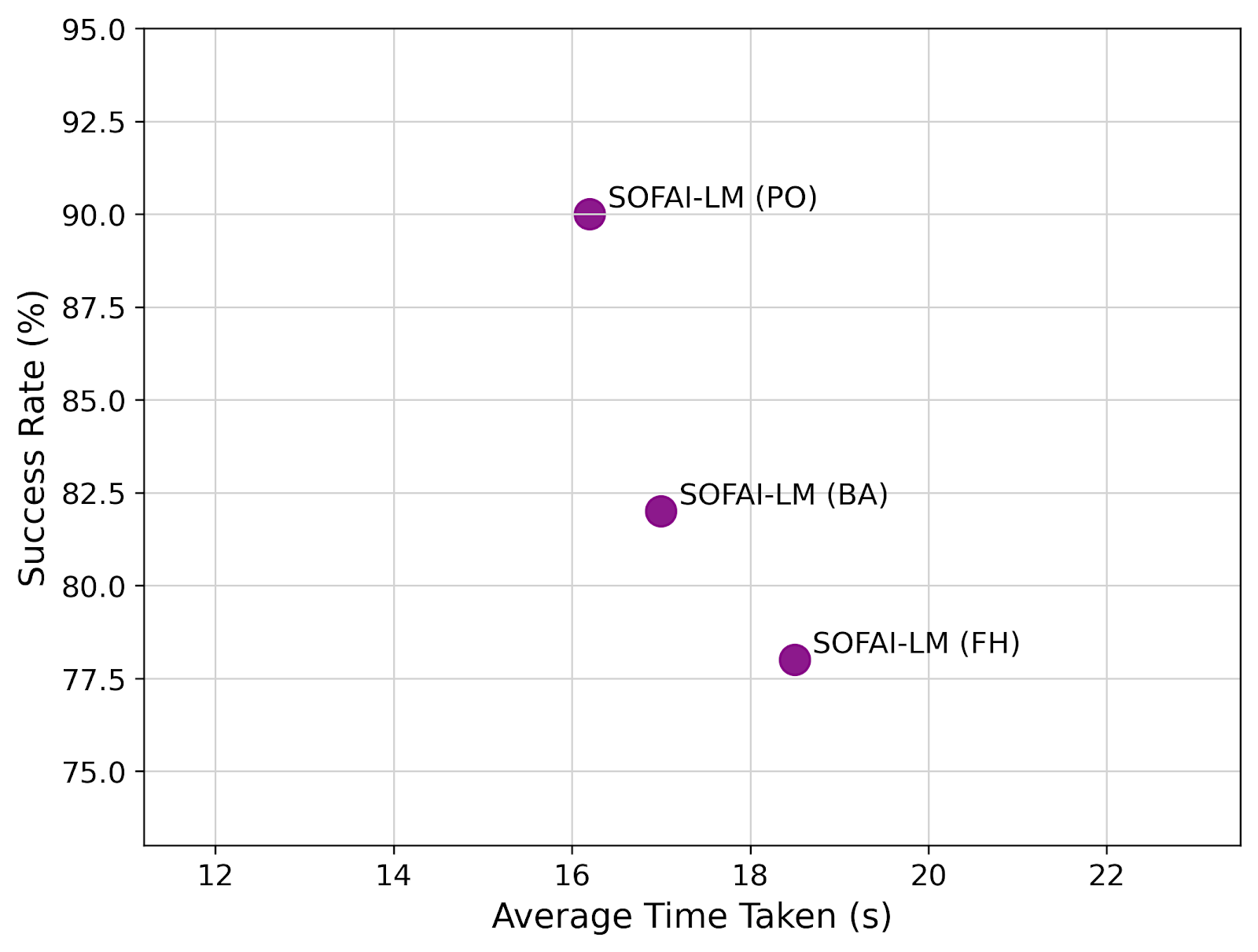}
    \caption{RQ3 - Pipeline: Granite 3.3B \(\rightarrow\) Qwen 2.5 Pro, Size 5}
    \label{fig:rq3_mc3_5}
\end{figure}

\begin{figure}[h!]
    \centering
    \includegraphics[width=0.95\linewidth]{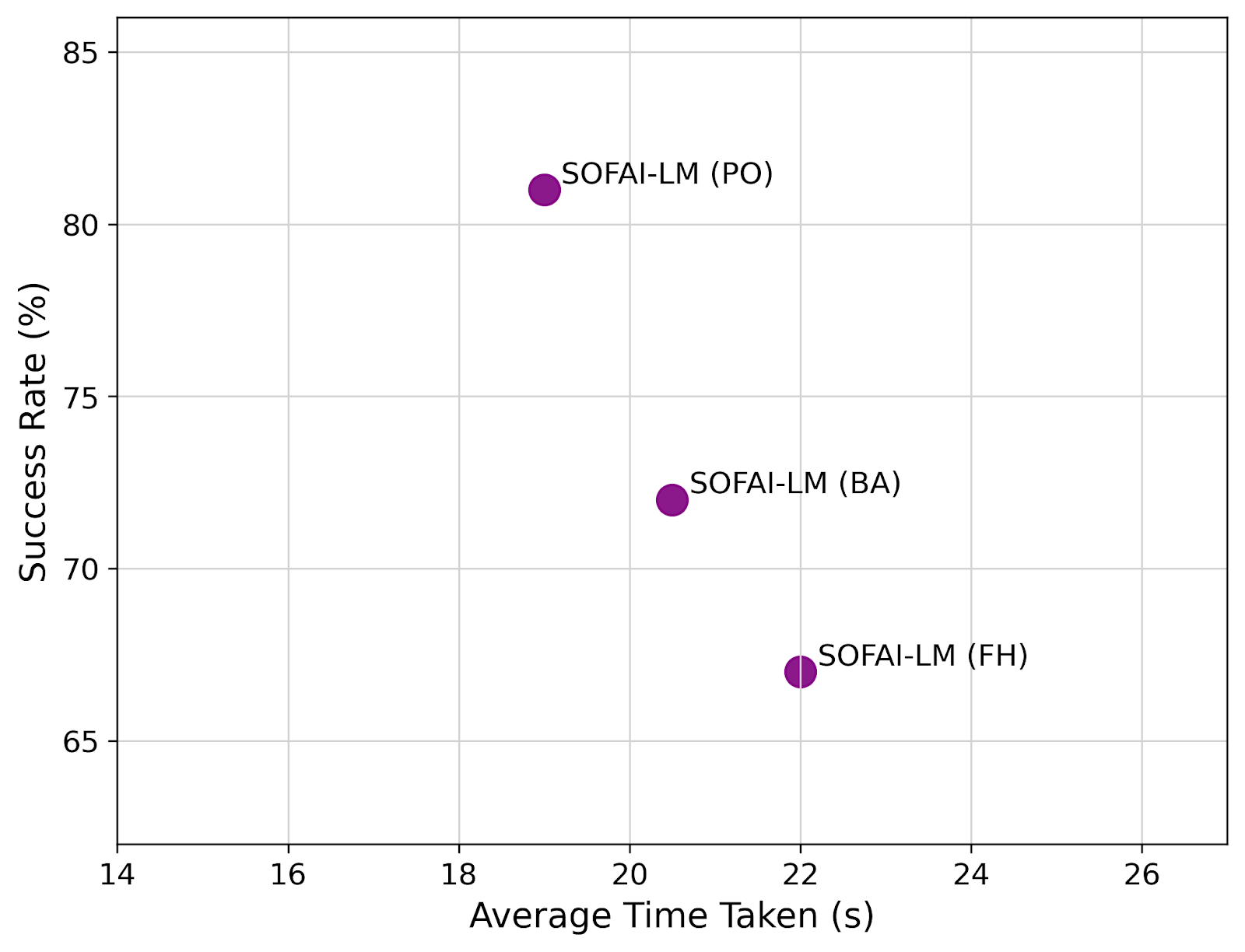}
    \caption{RQ3 - Pipeline: Granite 3.3B \(\rightarrow\) Qwen 2.5 Pro, Size 10}
    \label{fig:rq3_mc3_10}
\end{figure}

\begin{figure}[h!]
    \centering
    \includegraphics[width=0.95\linewidth]{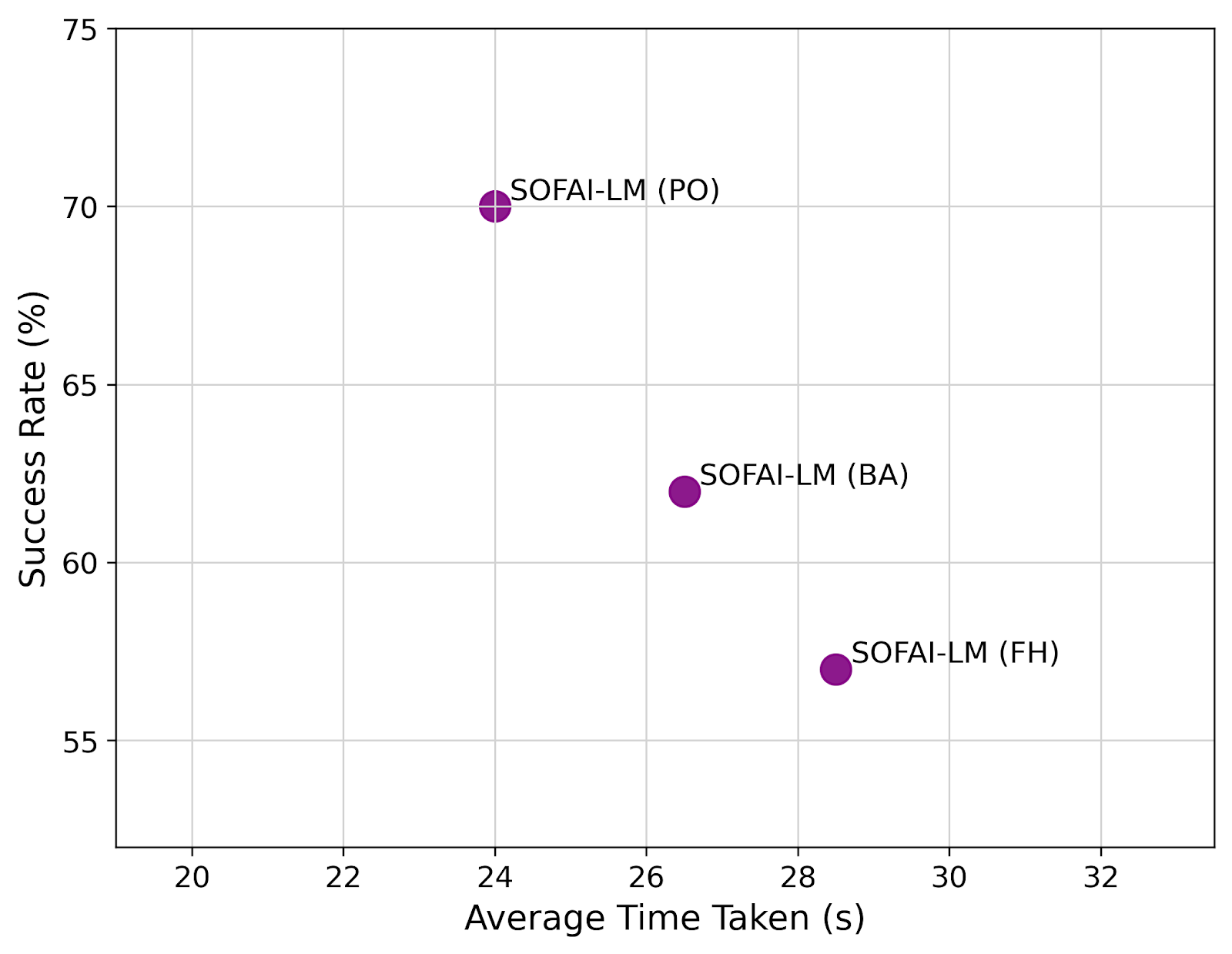}
    \caption{RQ3 - Pipeline: Granite 3.3B \(\rightarrow\) Qwen 2.5 Pro, Size 15}
    \label{fig:rq3_mc3_15}
\end{figure}

\begin{figure}[h!]
    \centering
    \includegraphics[width=0.95\linewidth]{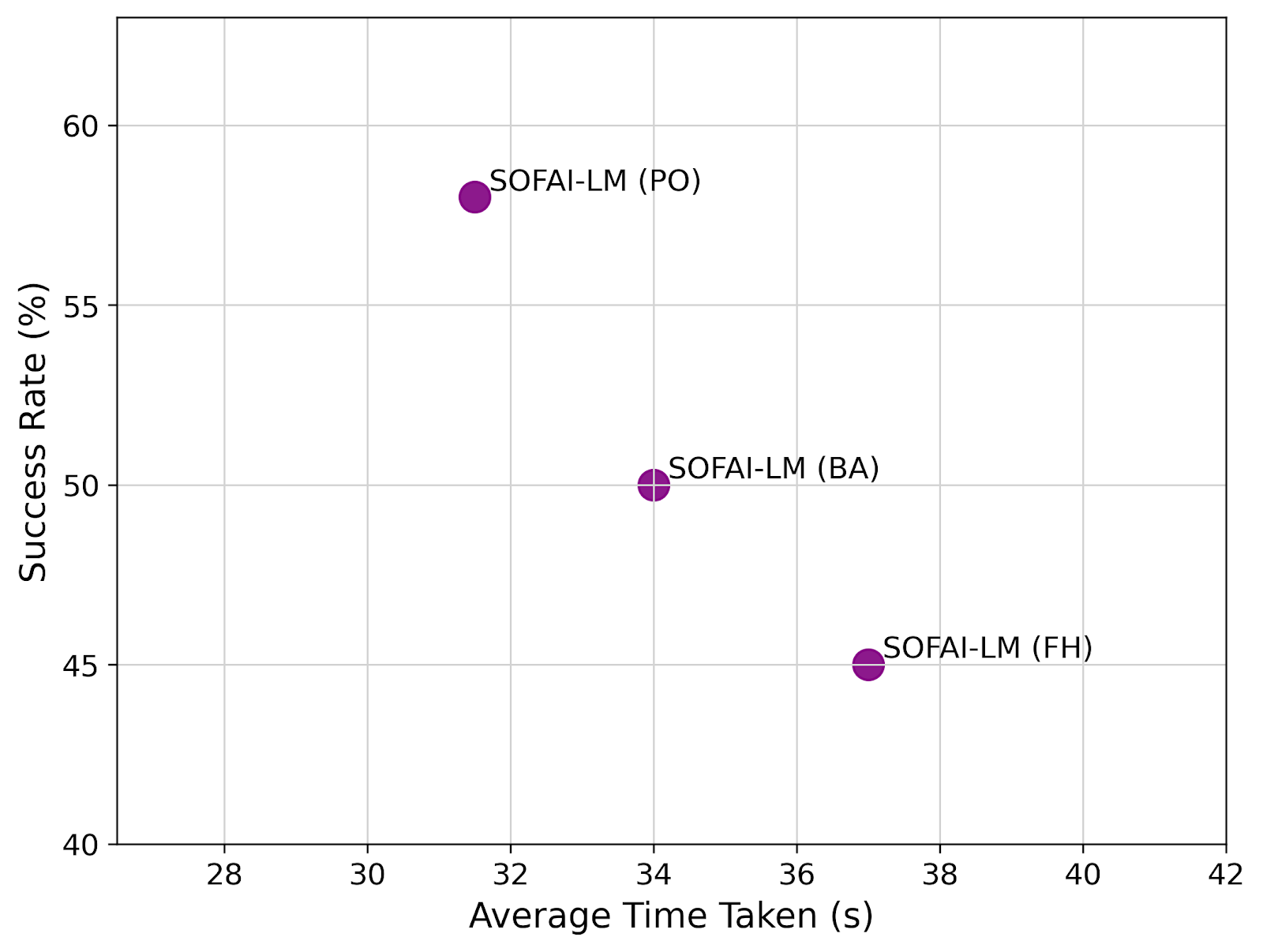}
    \caption{RQ3 - Pipeline: Granite 3.3B \(\rightarrow\) Qwen 2.5 Pro, Size 20}
    \label{fig:rq3_mc3_20}
\end{figure}

\begin{figure}[h!]
    \centering
    \includegraphics[width=0.95\linewidth]{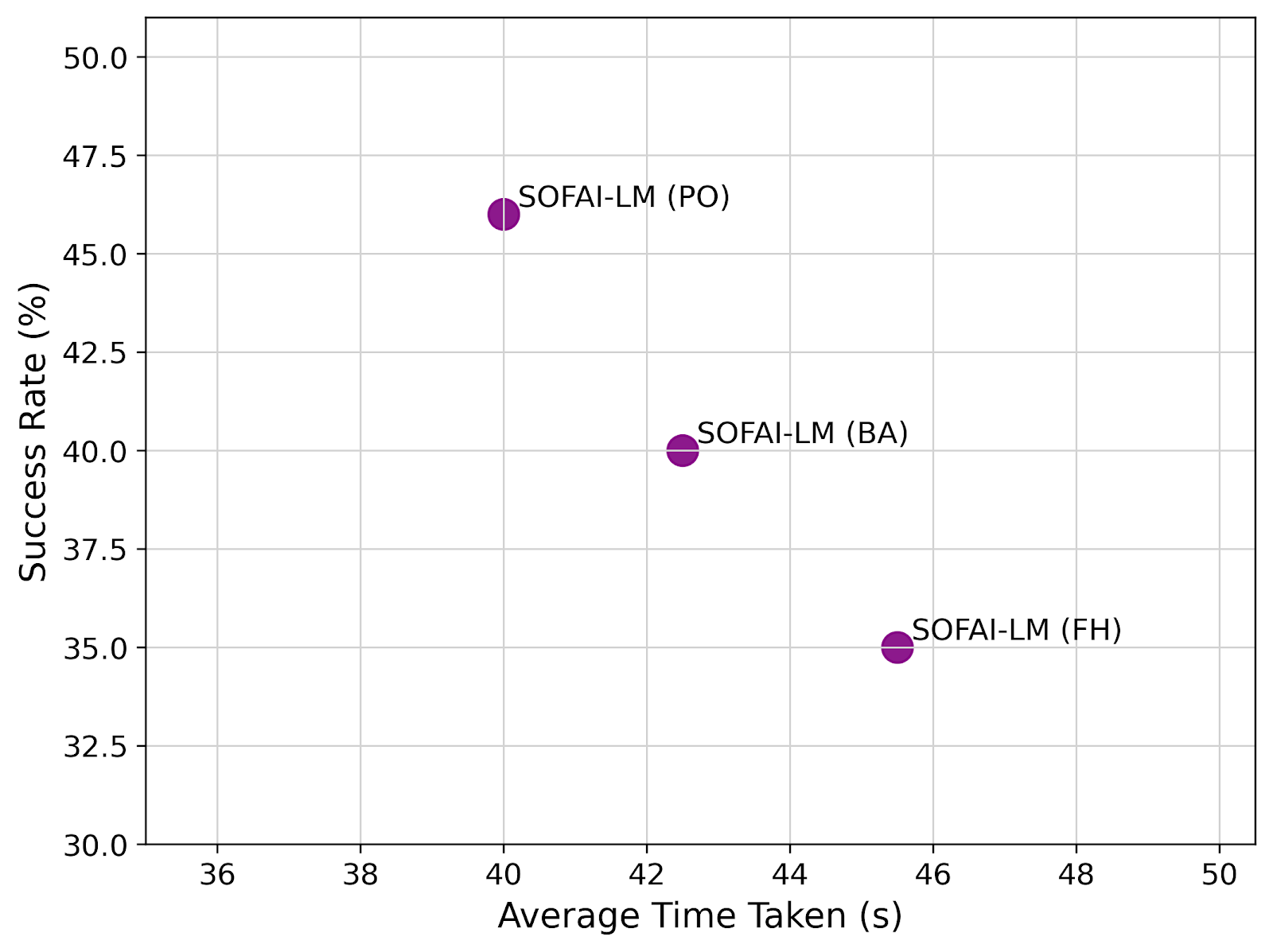}
    \caption{RQ3 - Pipeline: Granite 3.3B \(\rightarrow\) Qwen 2.5 Pro, Size 25}
    \label{fig:rq3_mc3_25}
\end{figure}

\paragraph{Description}
This series of plots (Figures \ref{fig:rq3_mc3_5} through \ref{fig:rq3_mc3_25}) shows the performance of the three prompting strategies when pairing the Granite 3.3B LLM with the Qwen 2.5 Pro LRM. The results are consistent with the previous findings. The `Problem-Only (PO)` strategy provides the best balance of success rate and time efficiency. Passing the `Best Attempt (BA)` or `Full History (FH)` to the LRM results in a clear decline in performance.

\paragraph{Summary Finding}
The superiority of the `Problem-Only` prompting strategy for graph coloring is robust across different S2 models, as the Granite \(\rightarrow\) Qwen pipeline shows the same performance degradation with added context as other model combinations.


\clearpage

\subsubsection{Pipeline: Llama 3.1 \(\rightarrow\) DeepSeek R1 8B}
\begin{figure}[h!]
    \centering
    \includegraphics[width=0.95\linewidth]{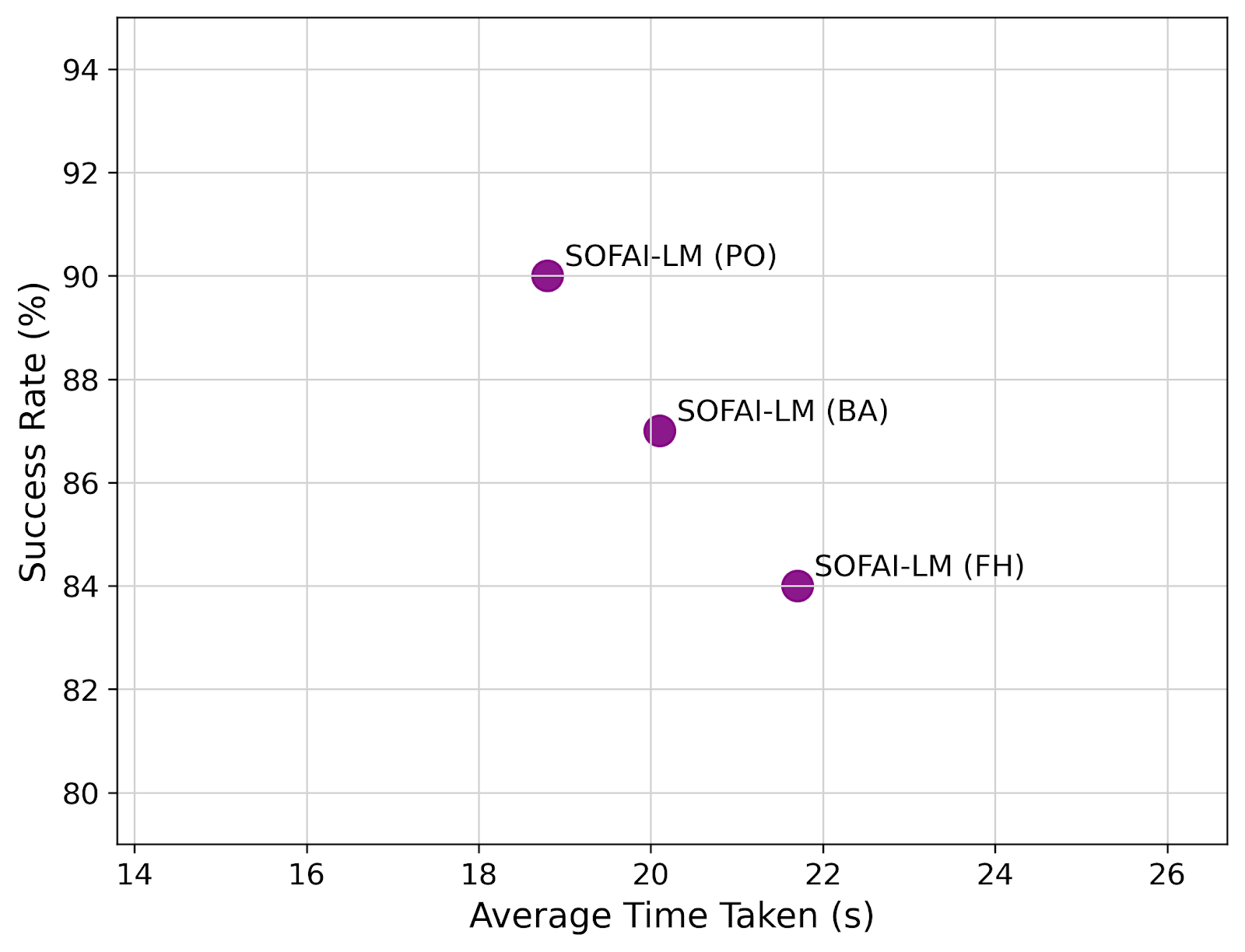}
    \caption{RQ3 - Pipeline: Llama 3.1 \(\rightarrow\) DeepSeek R1 8B, Size 5}
    \label{fig:rq3_mc4_5}
\end{figure}

\begin{figure}[h!]
    \centering
    \includegraphics[width=0.95\linewidth]{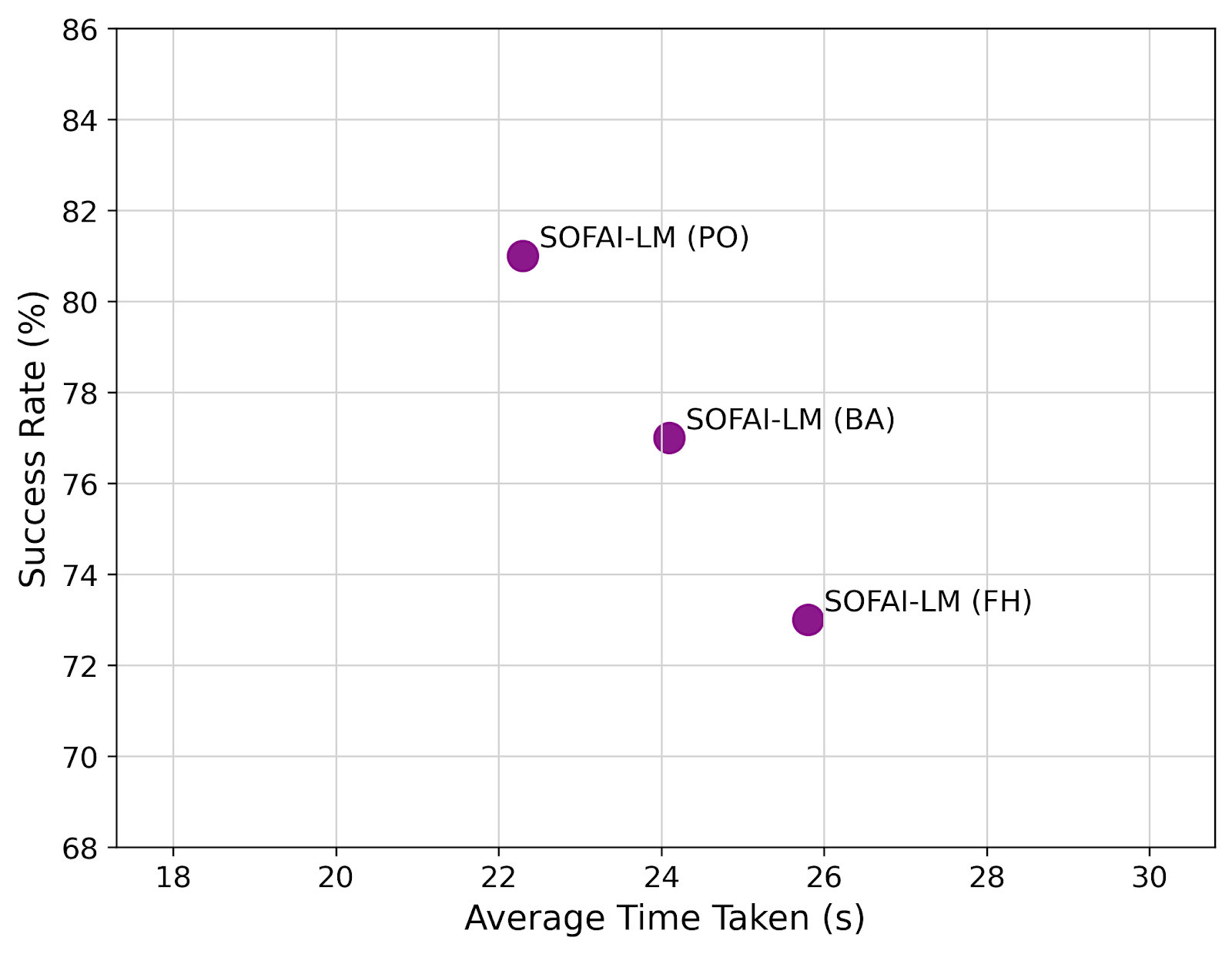}
    \caption{RQ3 - Pipeline: Llama 3.1 \(\rightarrow\) DeepSeek R1 8B, Size 10}
    \label{fig:rq3_mc4_10}
\end{figure}

\begin{figure}[h!]
    \centering
    \includegraphics[width=0.95\linewidth]{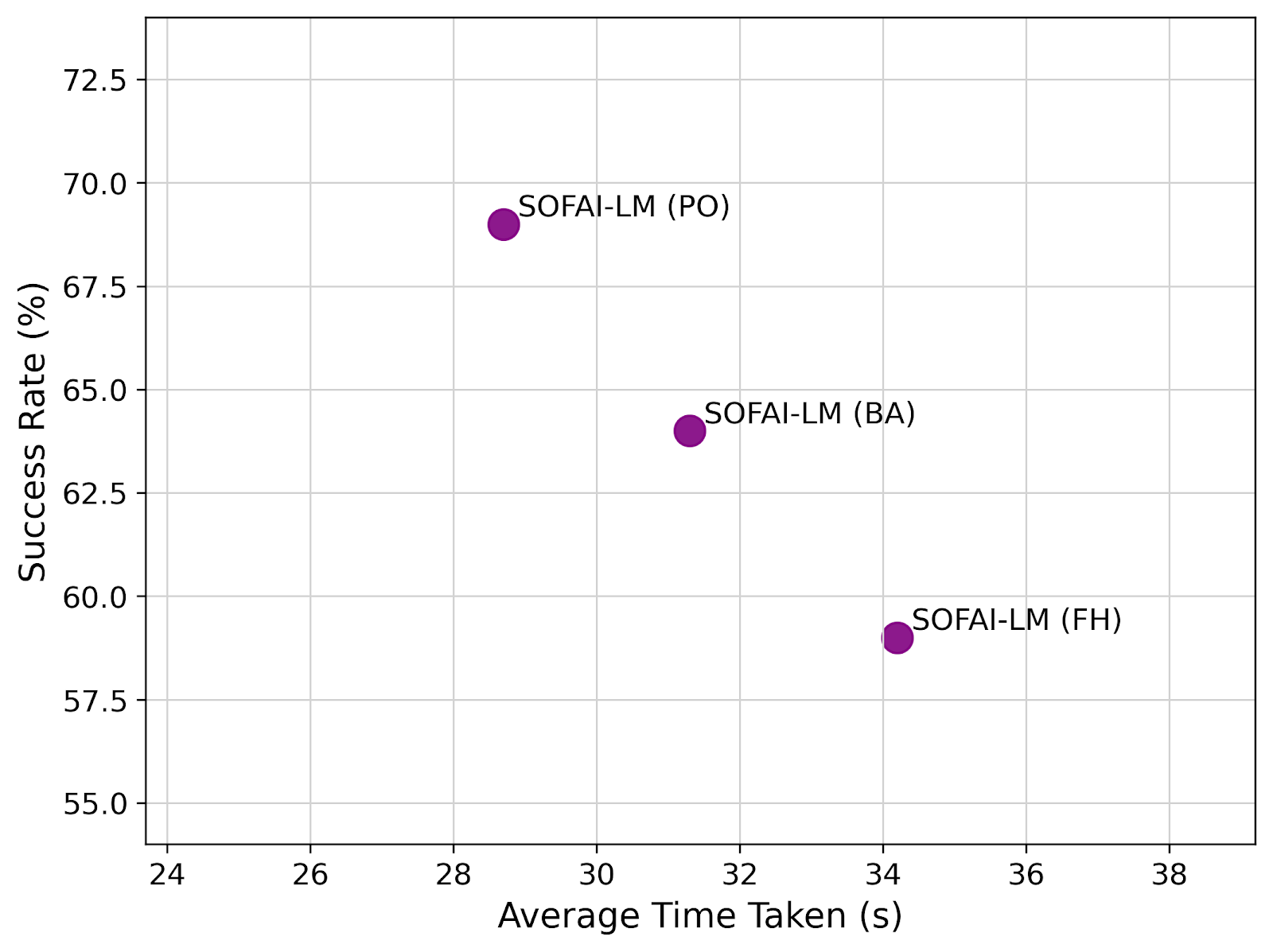}
    \caption{RQ3 - Pipeline: Llama 3.1 \(\rightarrow\) DeepSeek R1 8B, Size 15}
    \label{fig:rq3_mc4_15}
\end{figure}

\begin{figure}[h!]
    \centering
    \includegraphics[width=0.95\linewidth]{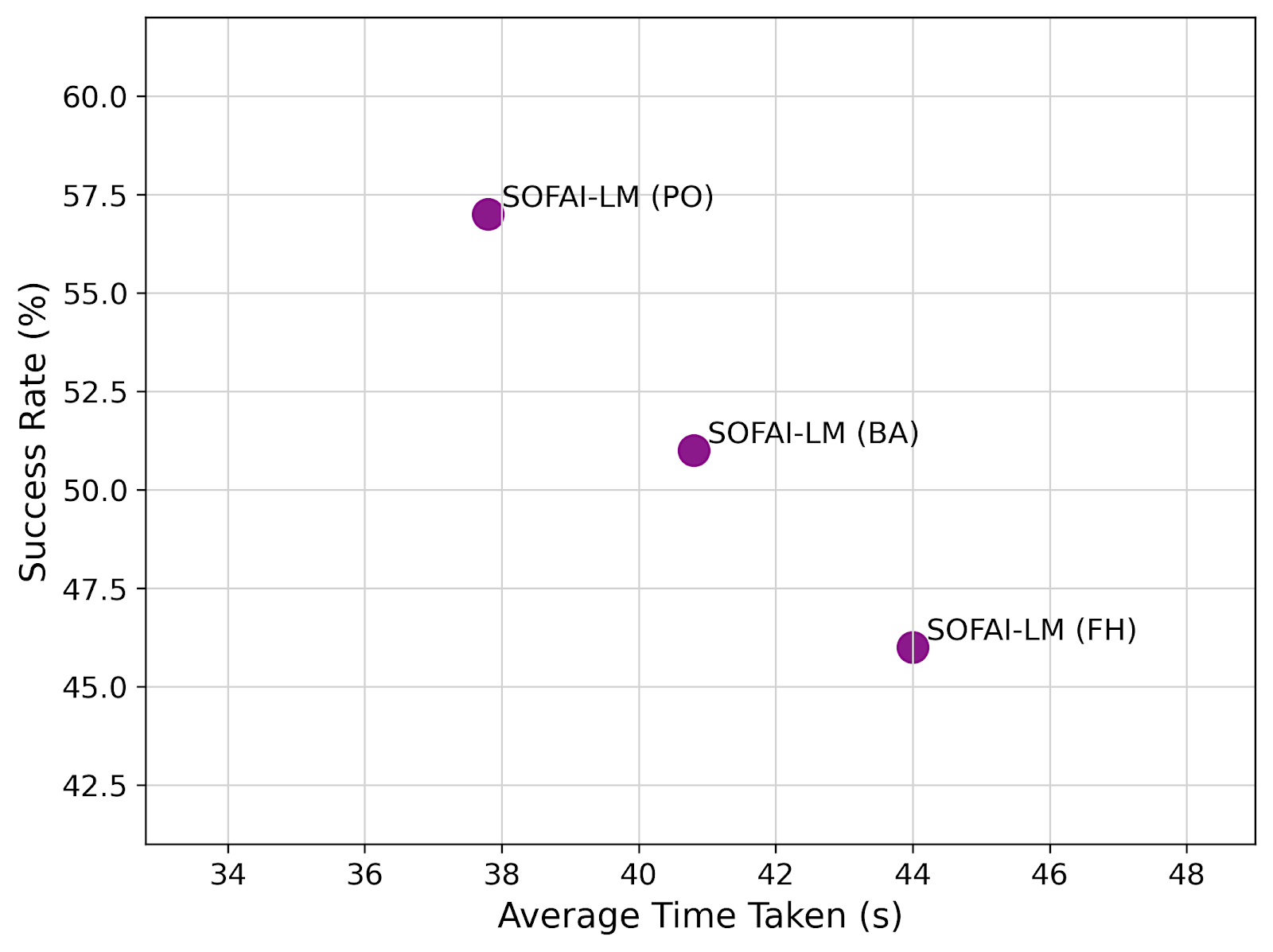}
    \caption{RQ3 - Pipeline: Llama 3.1 \(\rightarrow\) DeepSeek R1 8B, Size 20}
    \label{fig:rq3_mc4_20}
\end{figure}

\begin{figure}[h!]
    \centering
    \includegraphics[width=0.95\linewidth]{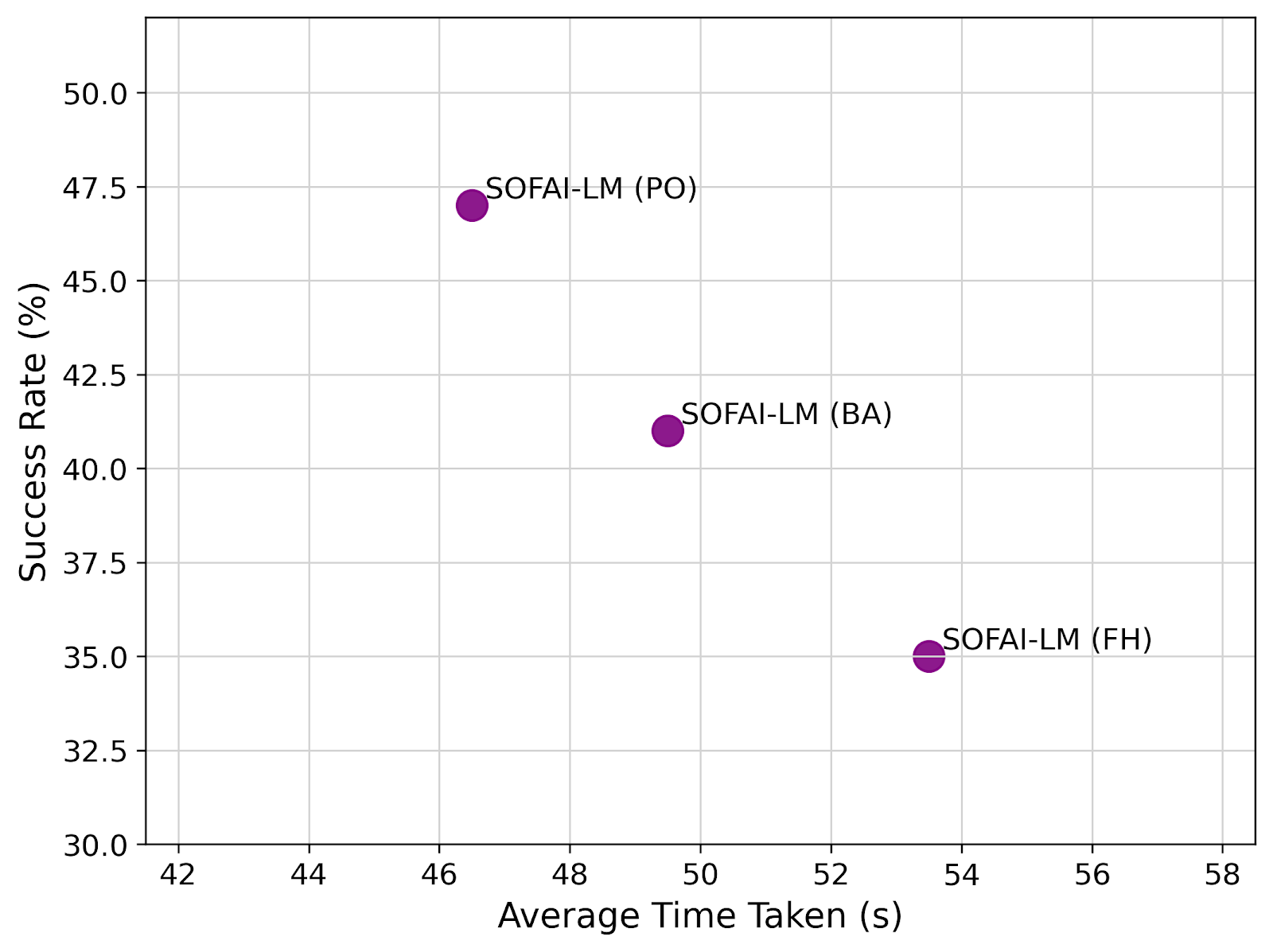}
    \caption{RQ3 - Pipeline: Llama 3.1 \(\rightarrow\) DeepSeek R1 8B, Size 25}
    \label{fig:rq3_mc4_25}
\end{figure}

\paragraph{Description}
This final set of plots for RQ3 (Figures \ref{fig:rq3_mc4_5} through \ref{fig:rq3_mc4_25}) shows the performance of the prompting strategies when using Llama 3.1 as the initial LLM. The results are unequivocally consistent with all previous findings for the graph coloring domain. The `Problem-Only (PO)` strategy provides the best results, achieving the highest success rates for the most reasonable time cost.

\paragraph{Summary Finding}
The finding that providing historical context harms LRM performance on graph coloring is robust and holds true even when a different powerful LLM, Llama 3.1, is used, confirming the generality of this key insight for the SOFAI-LM architecture in this domain.


\clearpage

\subsection{RQ4: Does SOFAI-LM perform better than its LRM counterpart?}
The following figures provide a direct comparison between the complete, optimized SOFAI-LM architecture and using the LRM as a standalone solver.

\subsubsection{Combination 1: SOFAI-LM (Granite \(\rightarrow\) DeepSeek) vs. LRM (DeepSeek)}
\begin{figure}[h!]
    \centering
    \includegraphics[width=0.95\linewidth]{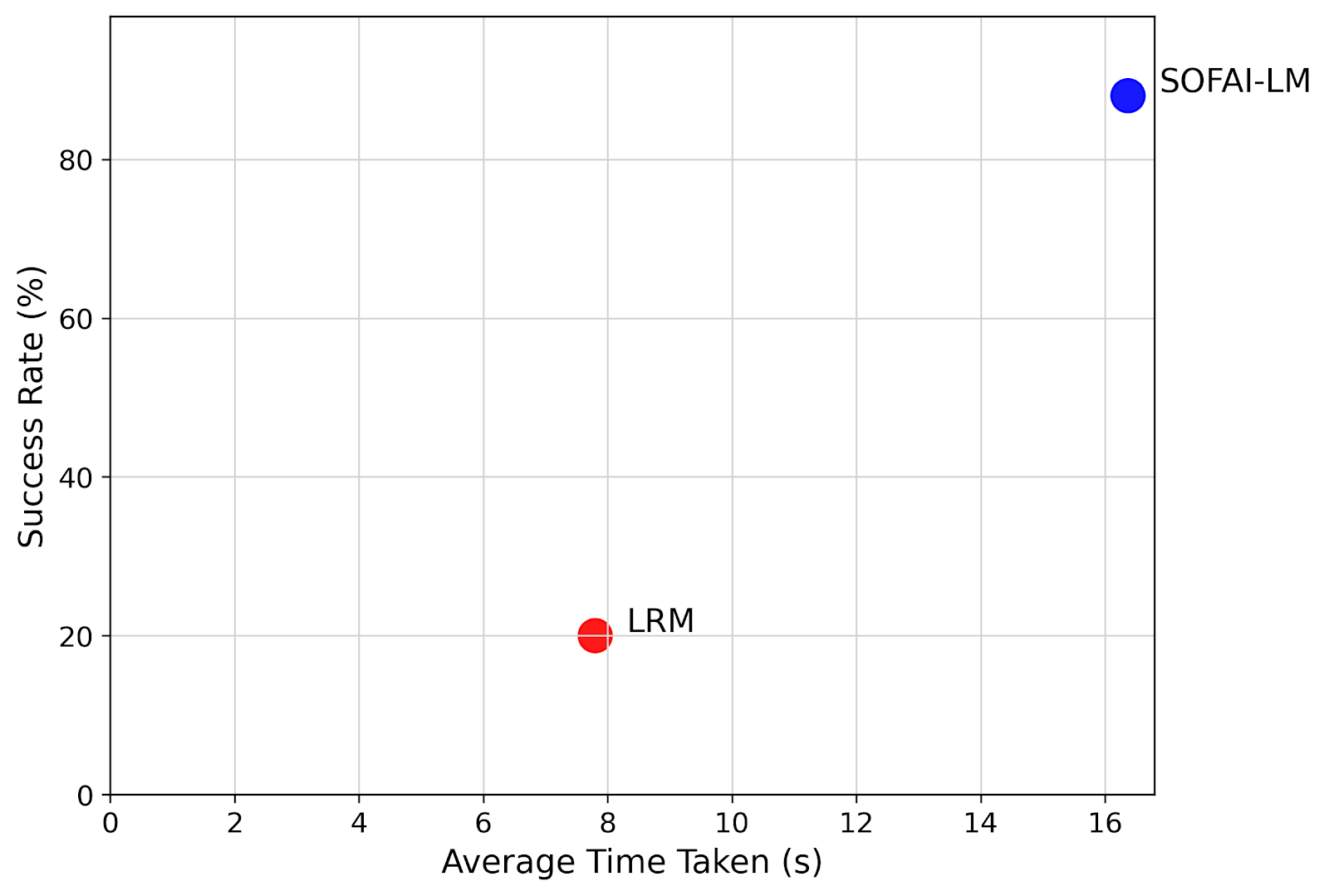}
    \caption{RQ4 - Combination 1: SOFAI-LM (Granite \(\rightarrow\) DeepSeek) vs. LRM (DeepSeek), Size 5}
    \label{fig:rq4_mc1_5}
\end{figure}

\begin{figure}[h!]
    \centering
    \includegraphics[width=0.95\linewidth]{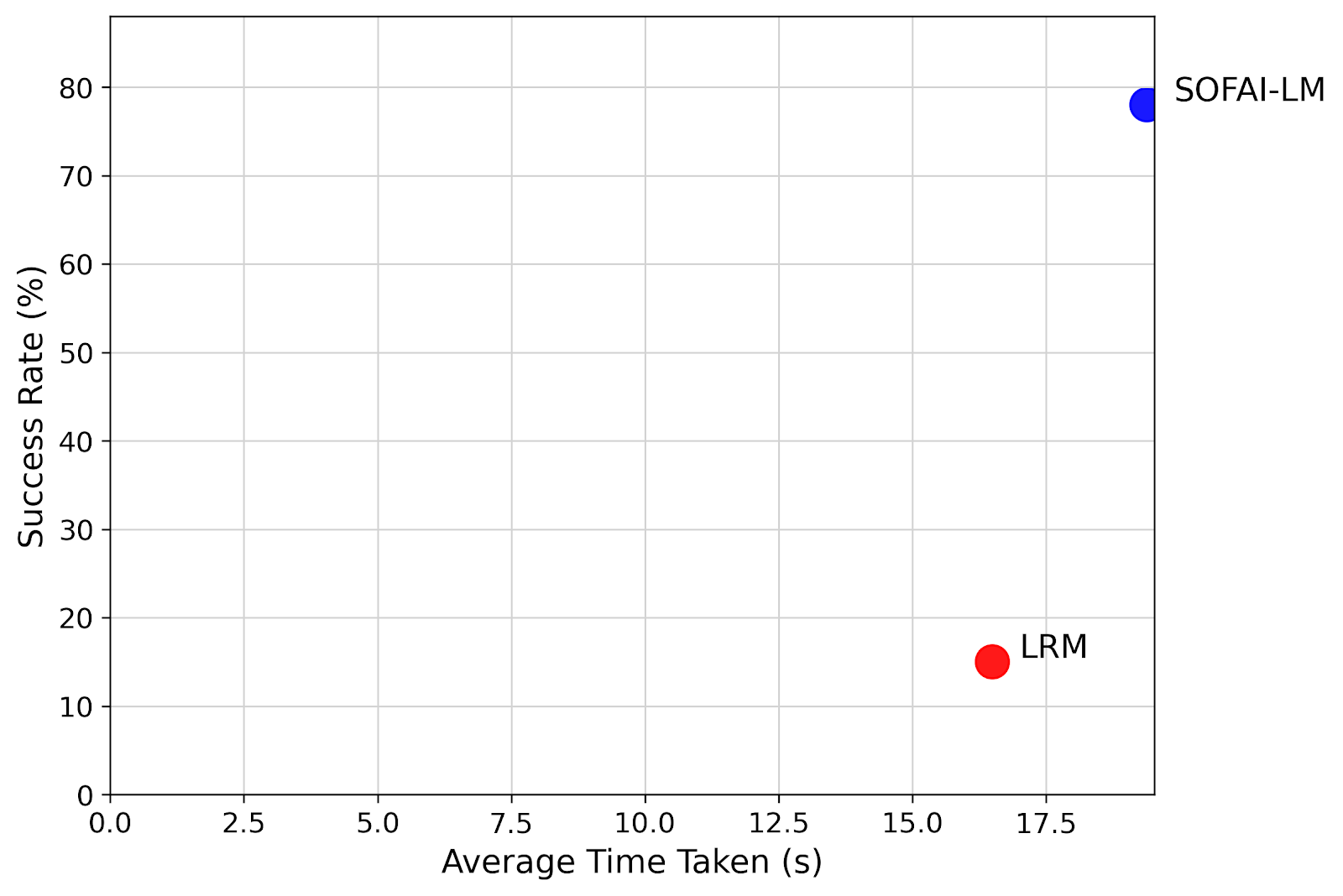}
    \caption{RQ4 - Combination 1: SOFAI-LM (Granite \(\rightarrow\) DeepSeek) vs. LRM (DeepSeek), Size 10}
    \label{fig:rq4_mc1_10}
\end{figure}

\begin{figure}[h!]
    \centering
    \includegraphics[width=0.95\linewidth]{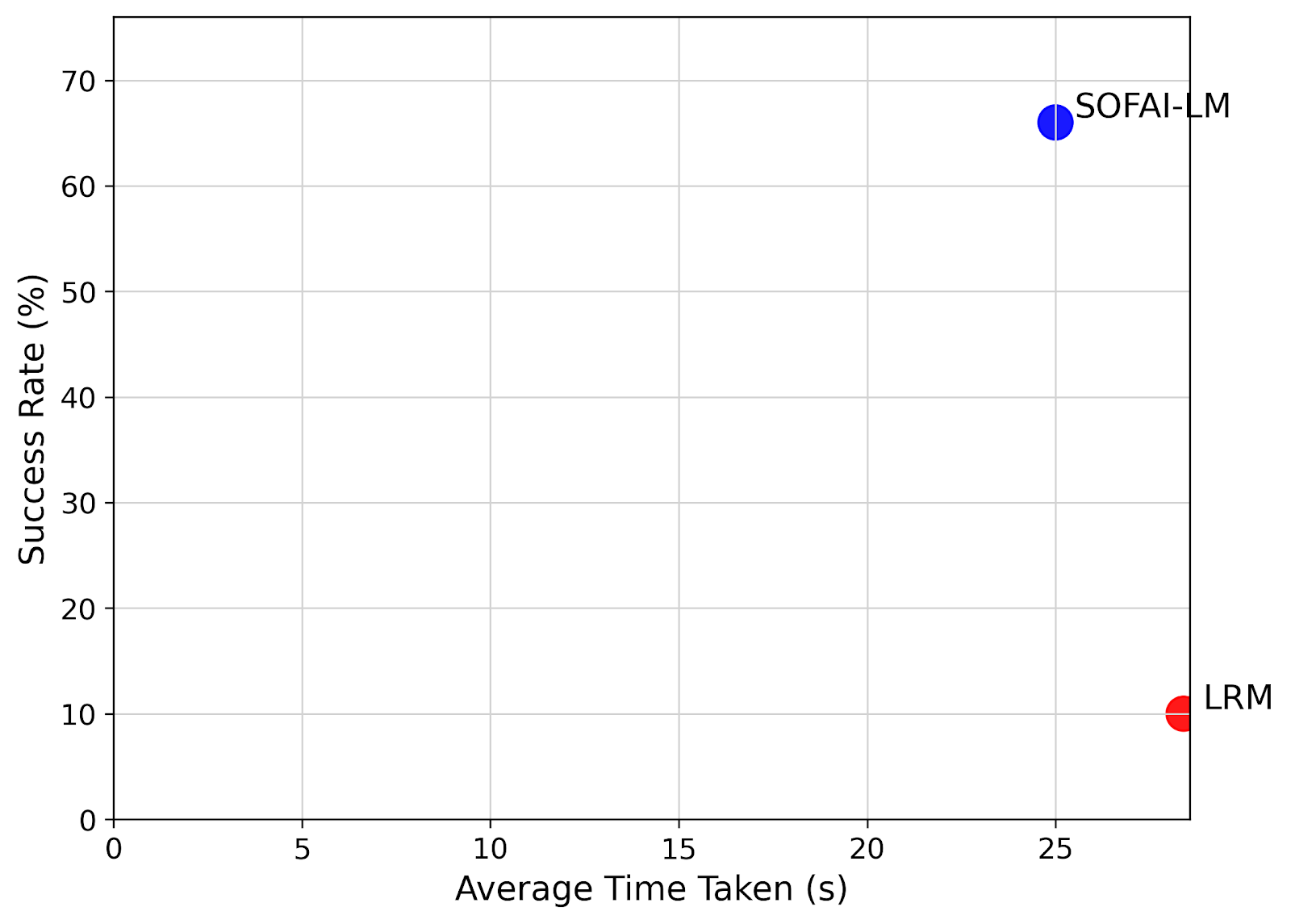}
    \caption{RQ4 - Combination 1: SOFAI-LM (Granite \(\rightarrow\) DeepSeek) vs. LRM (DeepSeek), Size 15}
    \label{fig:rq4_mc1_15}
\end{figure}

\begin{figure}[h!]
    \centering
    \includegraphics[width=0.95\linewidth]{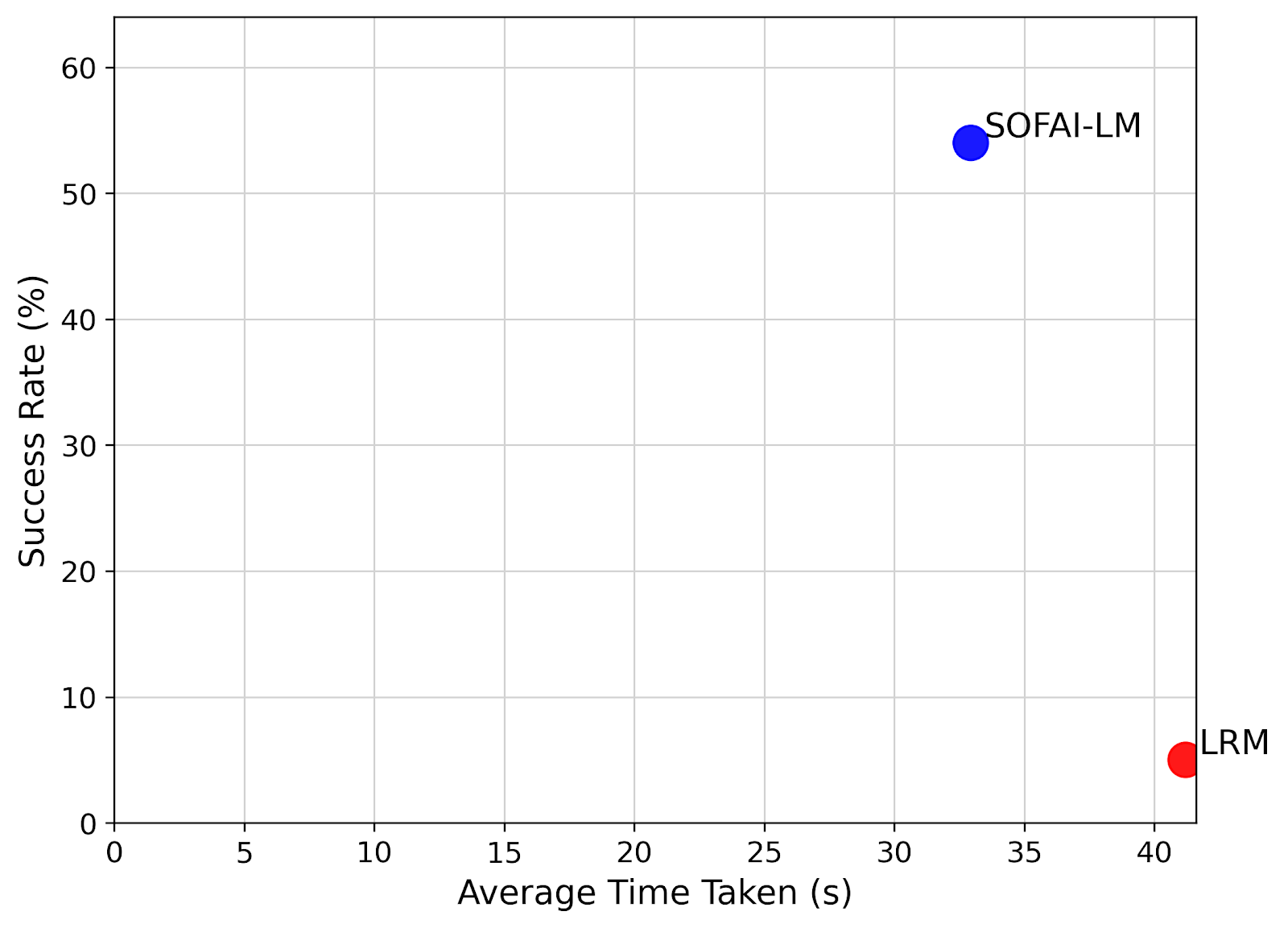}
    \caption{RQ4 - Combination 1: SOFAI-LM (Granite \(\rightarrow\) DeepSeek) vs. LRM (DeepSeek), Size 20}
    \label{fig:rq4_mc1_20}
\end{figure}

\paragraph{Description}
This series of plots (Figures \ref{fig:rq4_mc1_5} through \ref{fig:rq4_mc1_20}) provides a direct comparison between the integrated SOFAI-LM architecture and a standalone LRM (DeepSeek R1 8B). The results decisively show the superiority of the SOFAI-LM approach. Across every graph size, SOFAI-LM achieves a dramatically higher success rate. For larger problems (size 20 and 25), SOFAI-LM is not only substantially more accurate but also faster than the standalone LRM.

\paragraph{Summary Finding}
The SOFAI-LM architecture significantly outperforms a standalone LRM, delivering vastly superior success rates and better time efficiency, especially as graph size increases.

\clearpage

\subsubsection{Combination 2: SOFAI-LM (Granite \(\rightarrow\) Granite) vs. LRM (Granite)}
\begin{figure}[h!]
    \centering
    \includegraphics[width=0.95\linewidth]{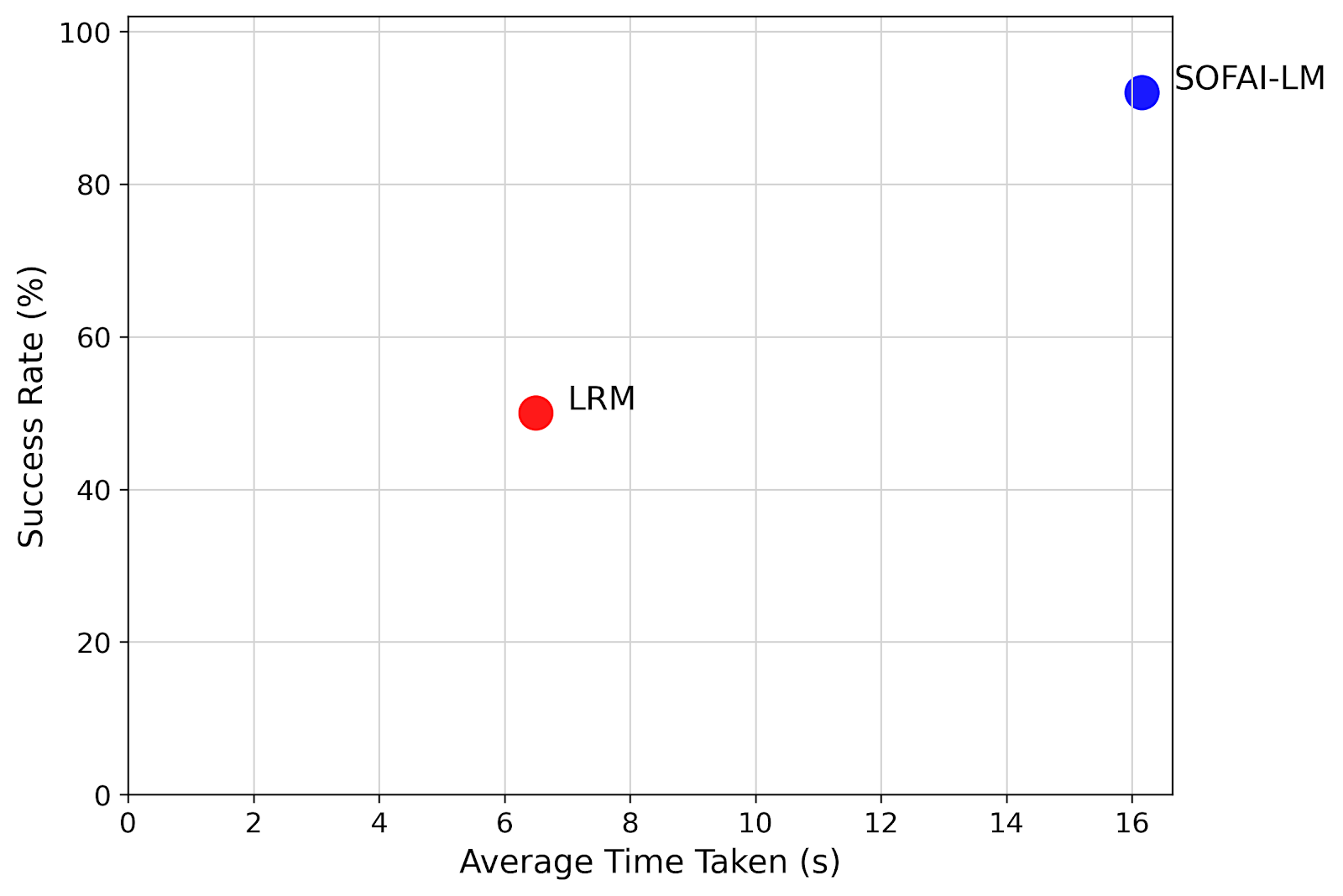}
    \caption{RQ4 - Combination 2: SOFAI-LM (Granite \(\rightarrow\) Granite) vs. LRM (Granite), Size 5}
    \label{fig:rq4_mc2_5}
\end{figure}

\begin{figure}[h!]
    \centering
    \includegraphics[width=0.95\linewidth]{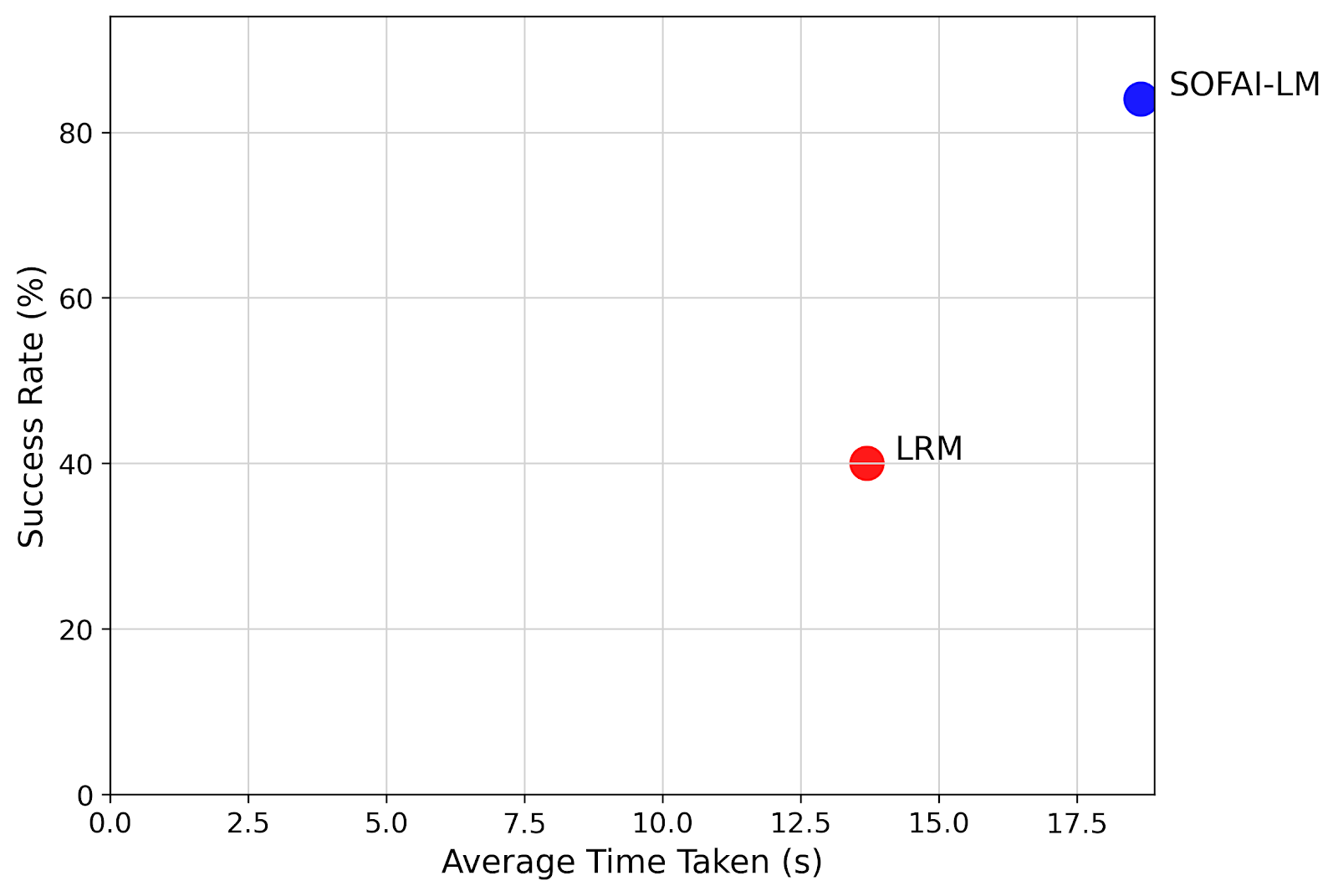}
    \caption{RQ4 - Combination 2: SOFAI-LM (Granite \(\rightarrow\) Granite) vs. LRM (Granite), Size 10}
    \label{fig:rq4_mc2_10}
\end{figure}

\begin{figure}[h!]
    \centering
    \includegraphics[width=0.95\linewidth]{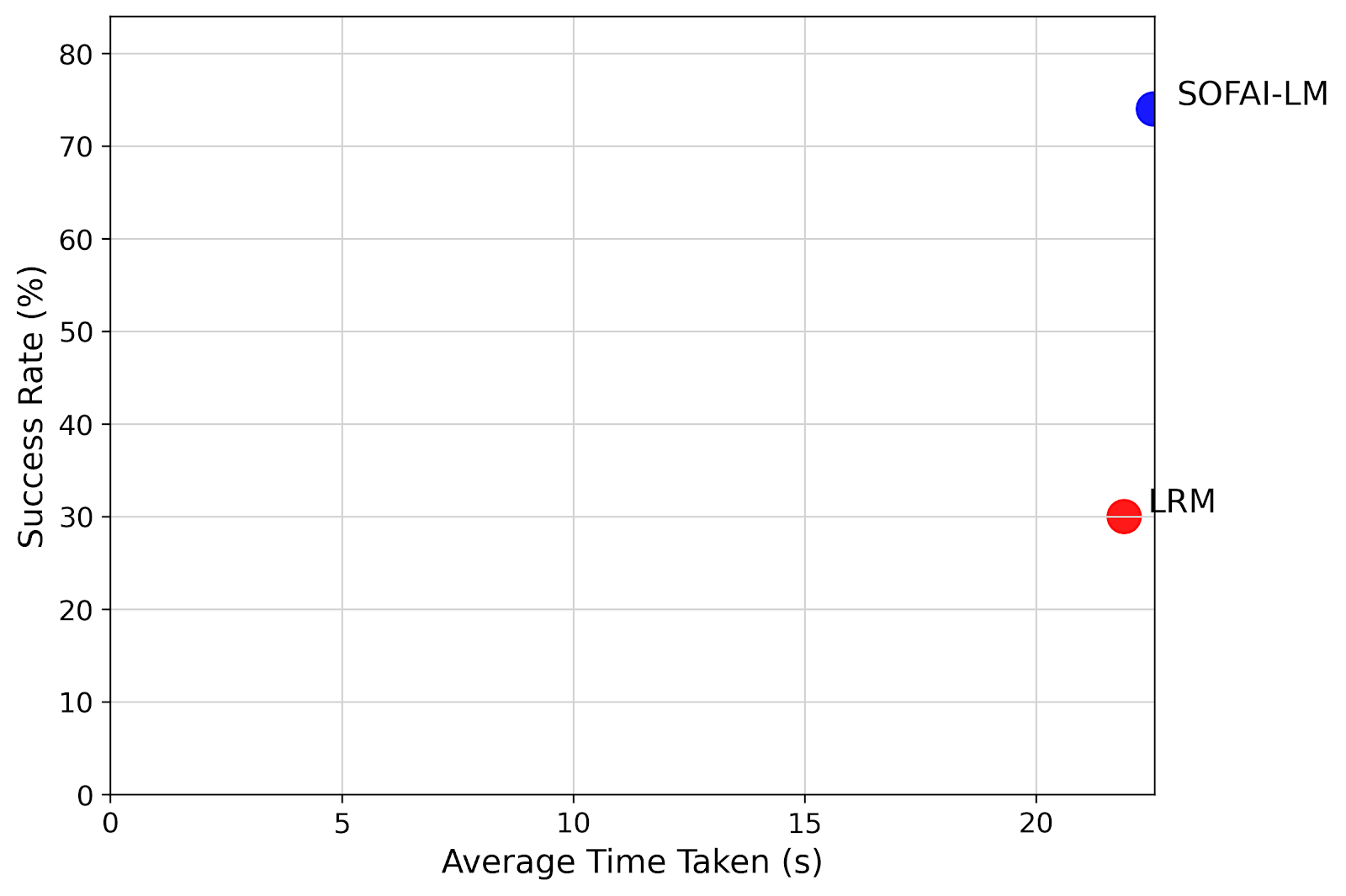}
    \caption{RQ4 - Combination 2: SOFAI-LM (Granite \(\rightarrow\) Granite) vs. LRM (Granite), Size 15}
    \label{fig:rq4_mc2_15}
\end{figure}

\begin{figure}[h!]
    \centering
    \includegraphics[width=0.95\linewidth]{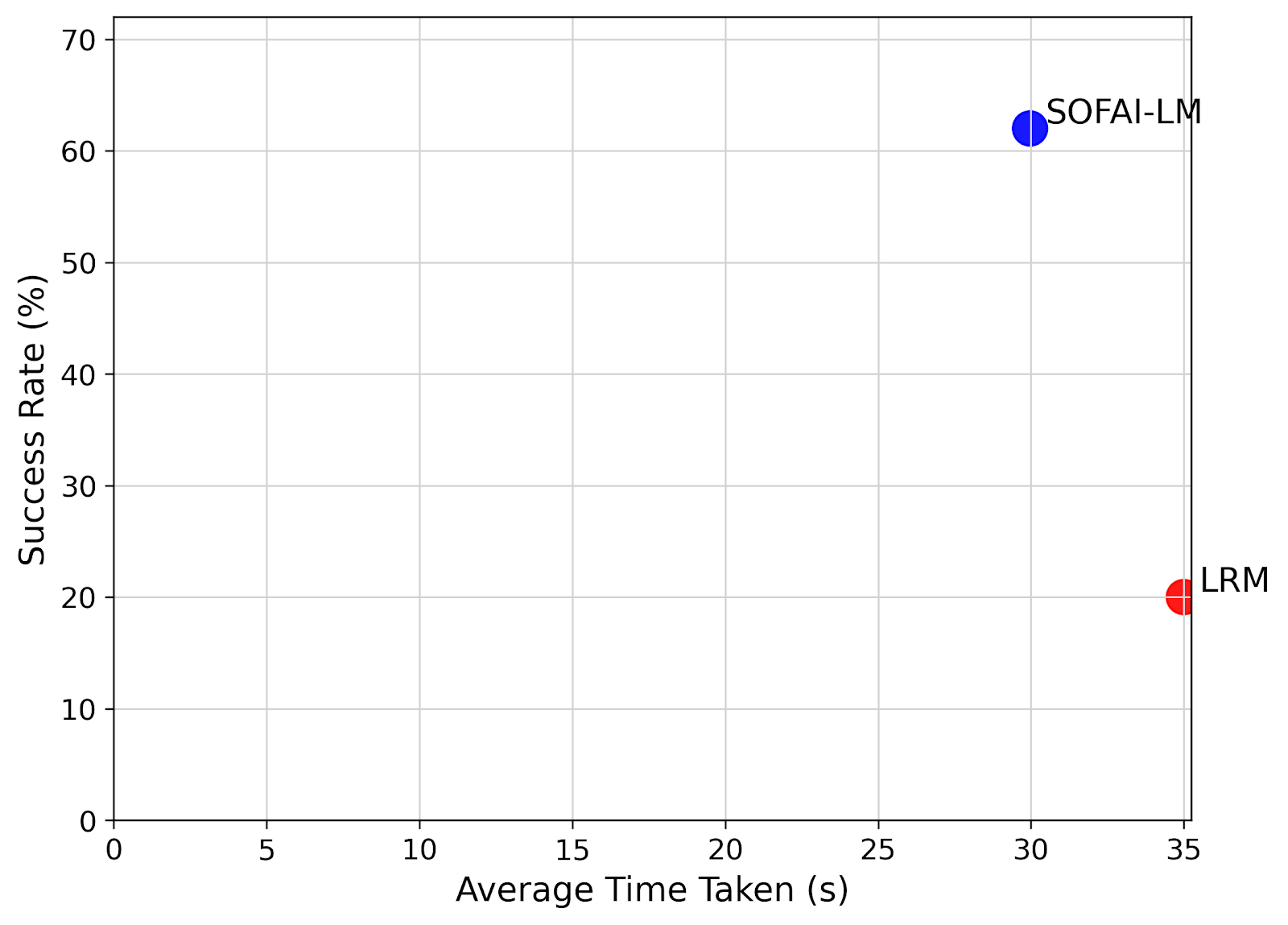}
    \caption{RQ4 - Combination 2: SOFAI-LM (Granite \(\rightarrow\) Granite) vs. LRM (Granite), Size 20}
    \label{fig:rq4_mc2_20}
\end{figure}

\begin{figure}[h!]
    \centering
    \includegraphics[width=0.95\linewidth]{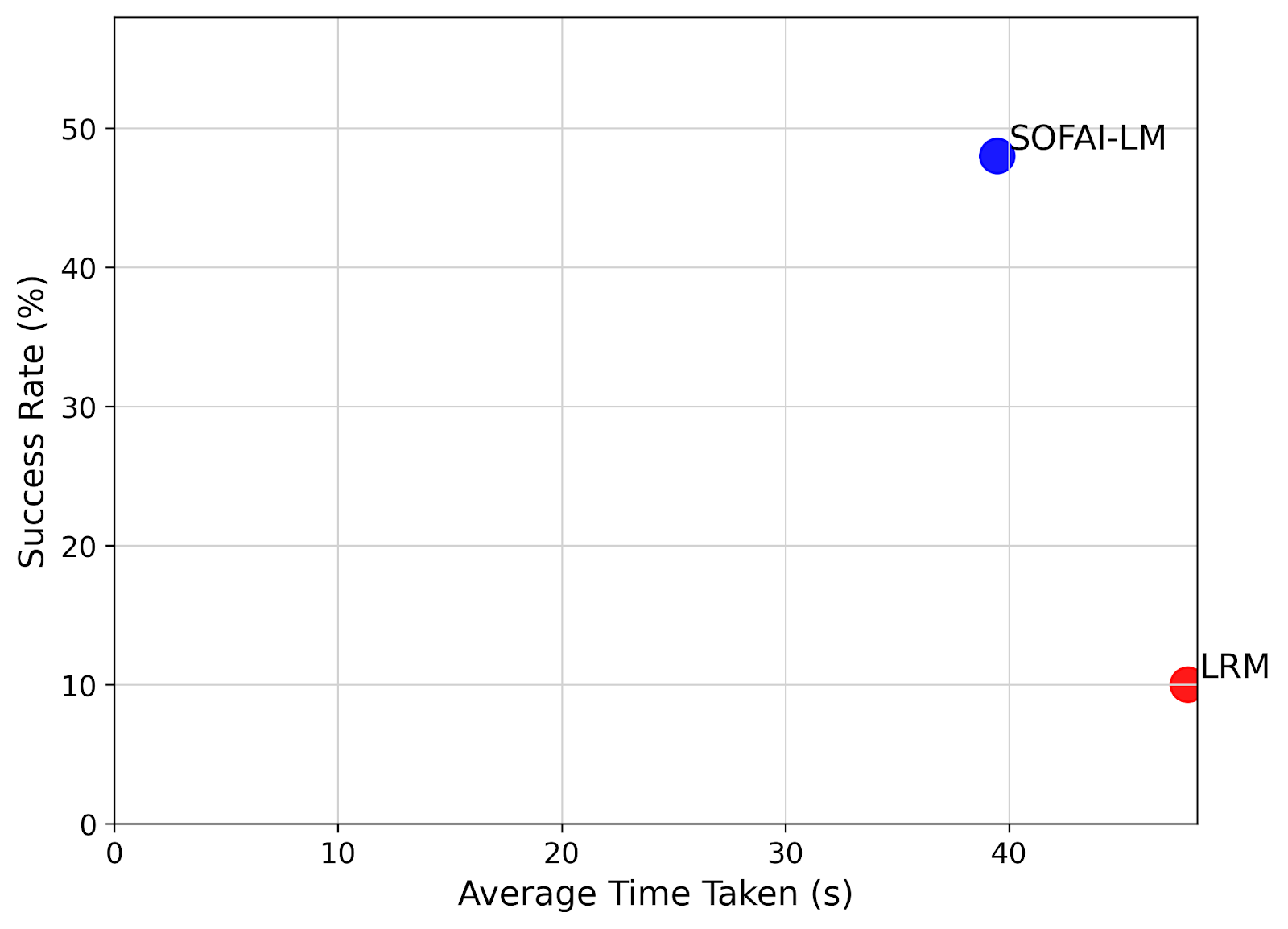}
    \caption{RQ4 - Combination 2: SOFAI-LM (Granite \(\rightarrow\) Granite) vs. LRM (Granite), Size 25}
    \label{fig:rq4_mc2_25}
\end{figure}

\paragraph{Description}
This set of figures (Figures \ref{fig:rq4_mc2_5} through \ref{fig:rq4_mc2_25}) compares the performance of the full SOFAI-LM pipeline against its LRM component, using Granite 3.3B for both roles. The results reinforce the value of the SOFAI-LM architecture even in a homogeneous model setting. The complete SOFAI-LM system consistently achieves a much higher success rate than using the Granite model as a standalone LRM. This demonstrates that the architectural design itself—using an iterative feedback loop with selective fallback—is the key driver of performance.

\paragraph{Summary Finding}
The SOFAI-LM architecture provides a substantial performance benefit over a standalone LRM, even when the same base model is used for both the fast and slow solver roles.


\clearpage

\subsubsection{Combination 3: SOFAI-LM (Granite \(\rightarrow\) Qwen) vs. LRM (Qwen)}
\begin{figure}[h!]
    \centering
    \includegraphics[width=0.95\linewidth]{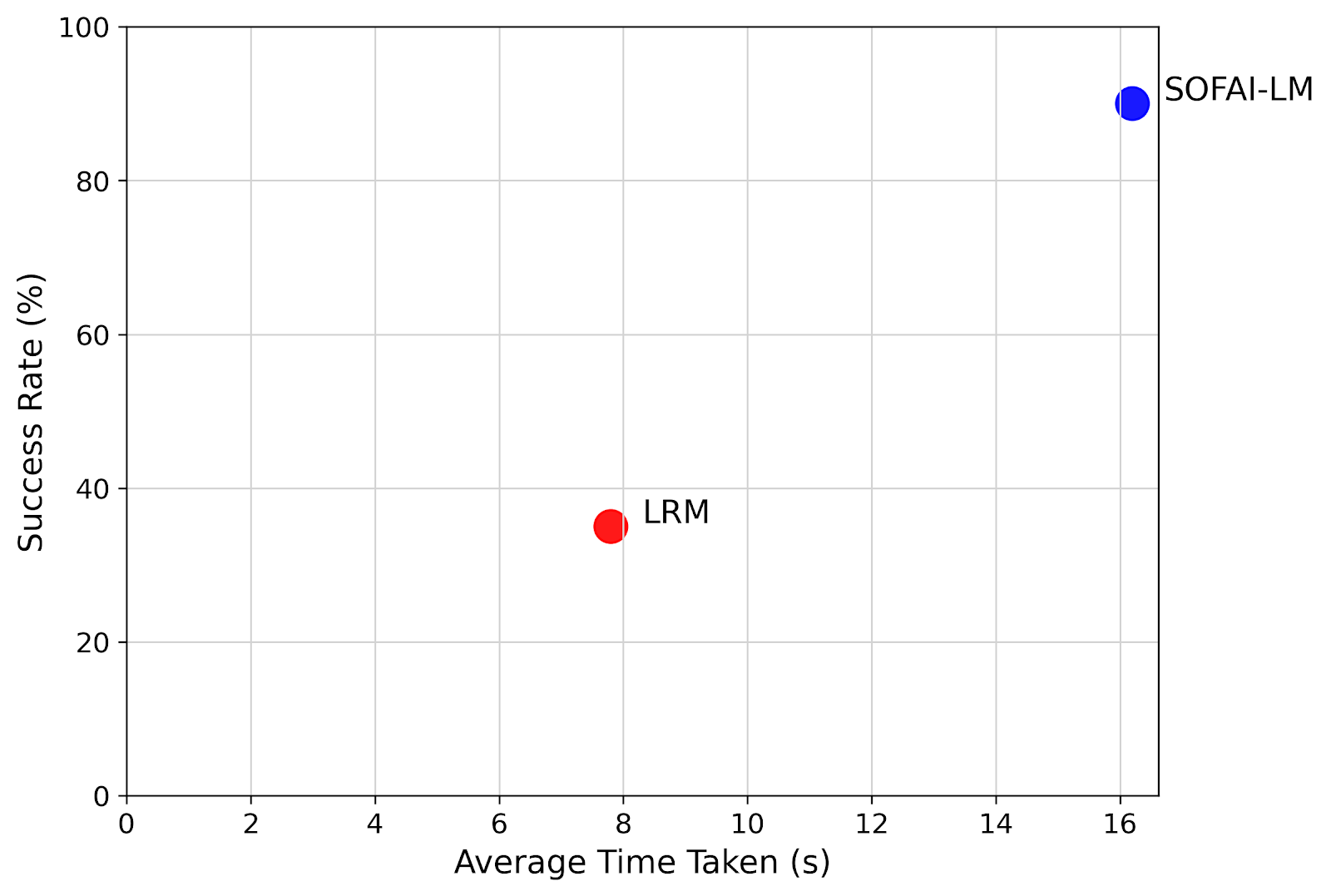}
    \caption{RQ4 - Combination 3: SOFAI-LM (Granite \(\rightarrow\) Qwen) vs. LRM (Qwen), Size 5}
    \label{fig:rq4_mc3_5}
\end{figure}

\begin{figure}[h!]
    \centering
    \includegraphics[width=0.95\linewidth]{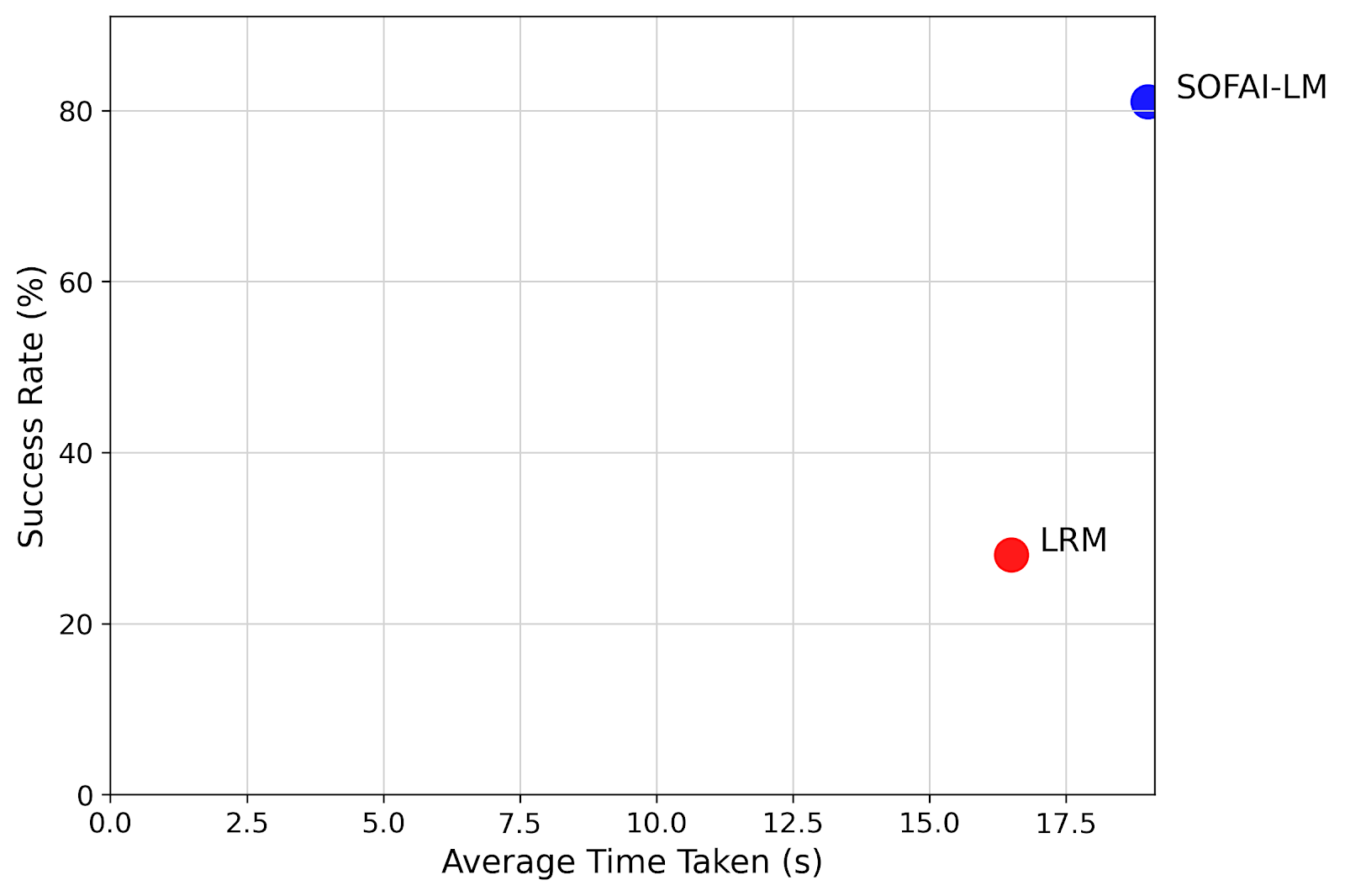}
    \caption{RQ4 - Combination 3: SOFAI-LM (Granite \(\rightarrow\) Qwen) vs. LRM (Qwen), Size 10}
    \label{fig:rq4_mc3_10}
\end{figure}

\begin{figure}[h!]
    \centering
    \includegraphics[width=0.95\linewidth]{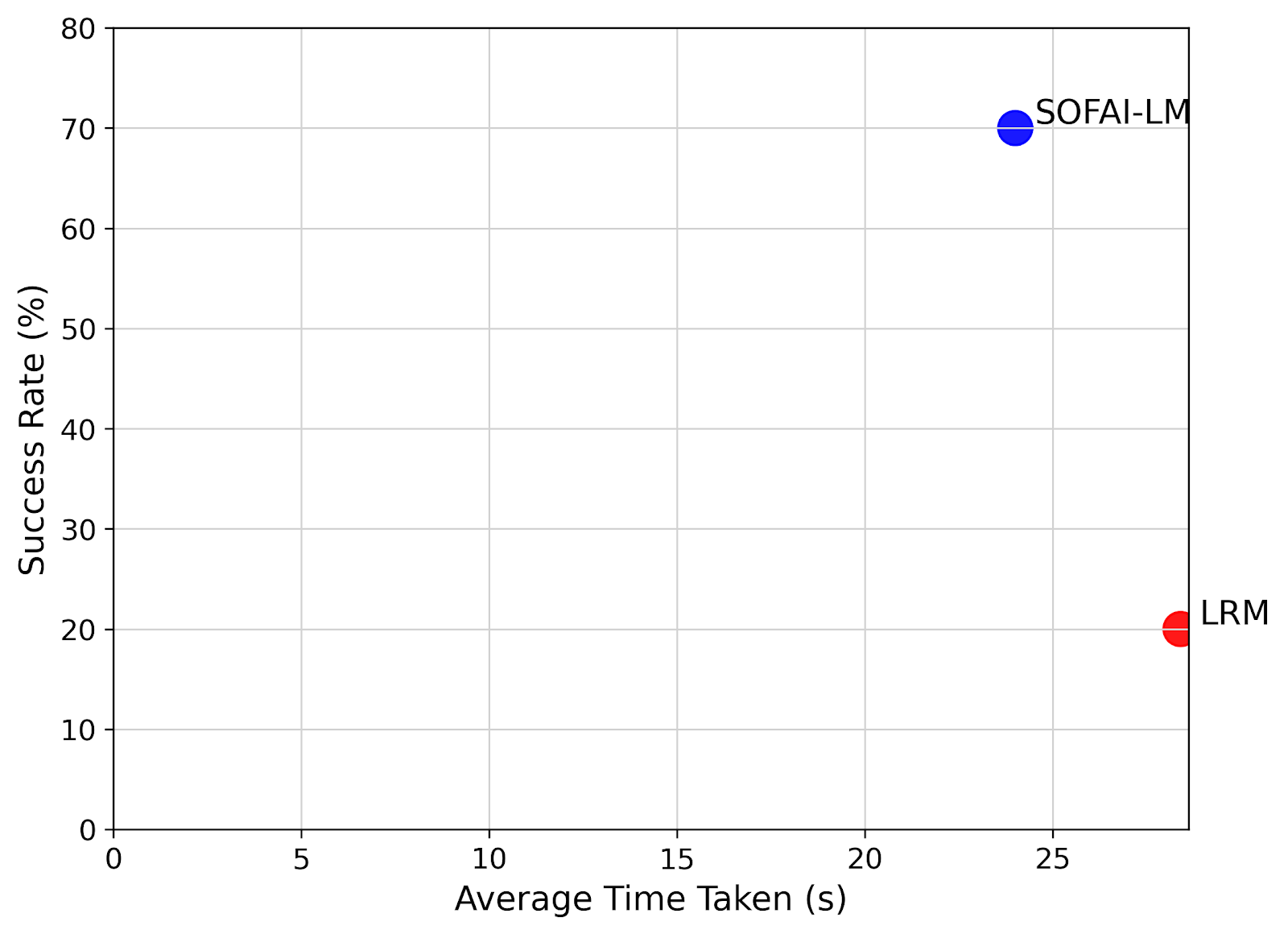}
    \caption{RQ4 - Combination 3: SOFAI-LM (Granite \(\rightarrow\) Qwen) vs. LRM (Qwen), Size 15}
    \label{fig:rq4_mc3_15}
\end{figure}

\begin{figure}[h!]
    \centering
    \includegraphics[width=0.95\linewidth]{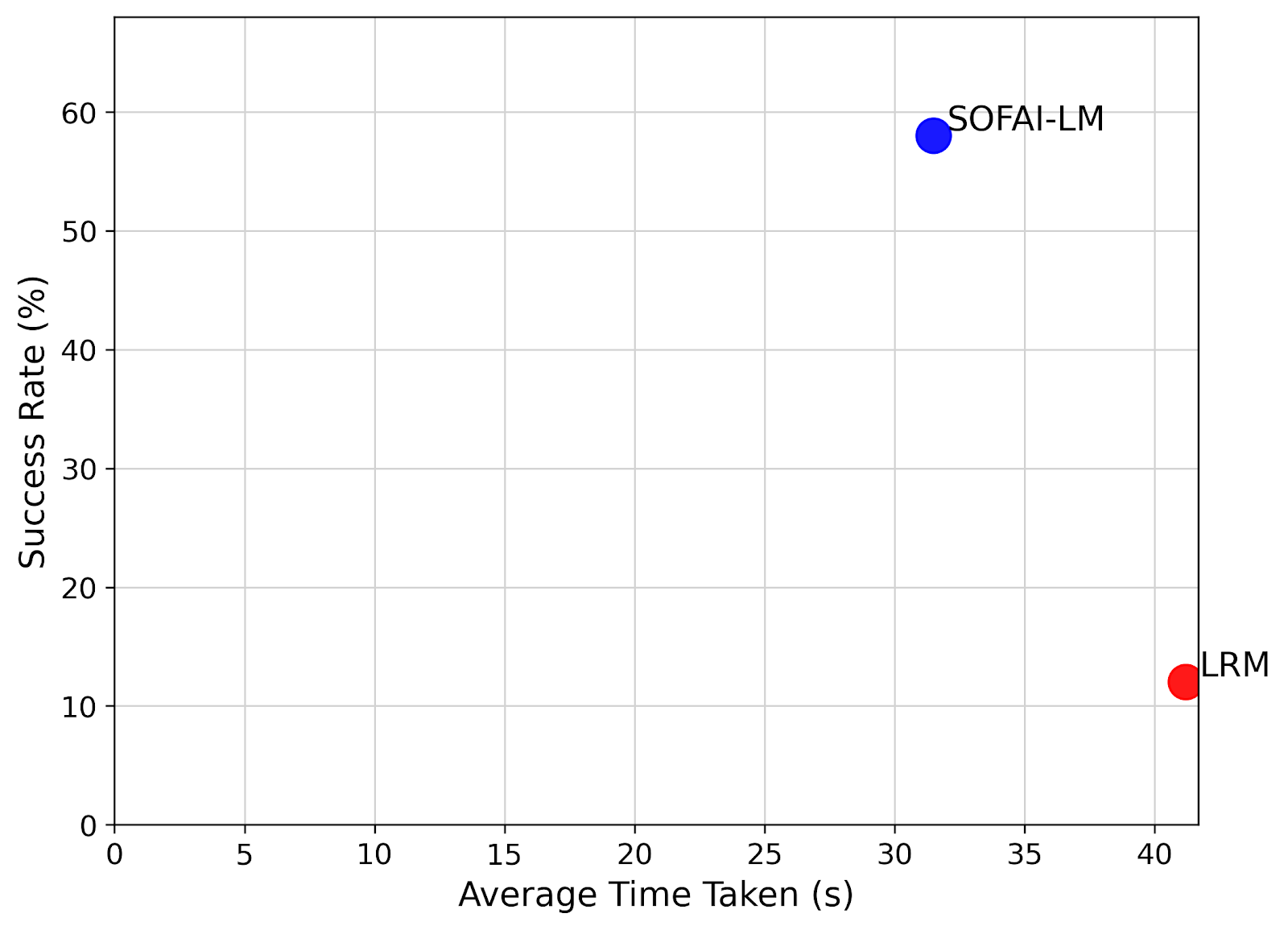}
    \caption{RQ4 - Combination 3: SOFAI-LM (Granite \(\rightarrow\) Qwen) vs. LRM (Qwen), Size 20}
    \label{fig:rq4_mc3_20}
\end{figure}

\begin{figure}[h!]
    \centering
    \includegraphics[width=0.95\linewidth]{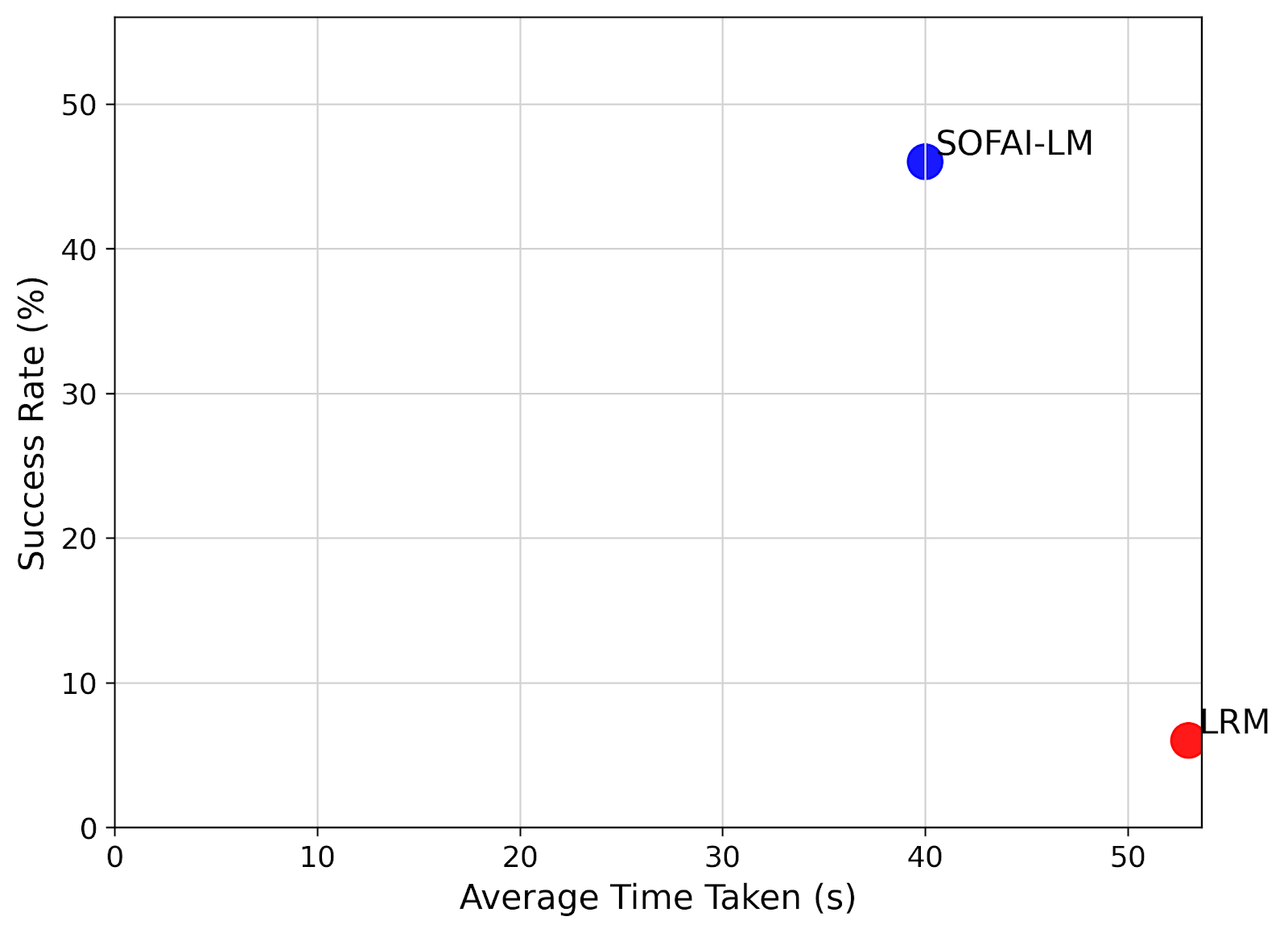}
    \caption{RQ4 - Combination 3: SOFAI-LM (Granite \(\rightarrow\) Qwen) vs. LRM (Qwen), Size 25}
    \label{fig:rq4_mc3_25}
\end{figure}

\paragraph{Description}
This set of figures (Figures \ref{fig:rq4_mc3_5} through \ref{fig:rq4_mc3_25}) compares the full SOFAI-LM pipeline, using Granite 3.3B as the LLM and Qwen 2.5 Pro as the LRM, against the standalone Qwen LRM. The results robustly confirm the paper's central thesis. The integrated SOFAI-LM system consistently and significantly outperforms the standalone LRM on success rate. As the problem complexity scales up, SOFAI-LM also becomes more time-efficient.

\paragraph{Summary Finding}
The SOFAI-LM architecture provides a decisive advantage in both accuracy and efficiency over a powerful standalone LRM like Qwen 2.5 Pro, confirming the generalizability of the approach.


\clearpage

\subsubsection{Combination 4: SOFAI-LM (Llama \(\rightarrow\) DeepSeek) vs. LRM (DeepSeek)}
\begin{figure}[h!]
    \centering
    \includegraphics[width=0.95\linewidth]{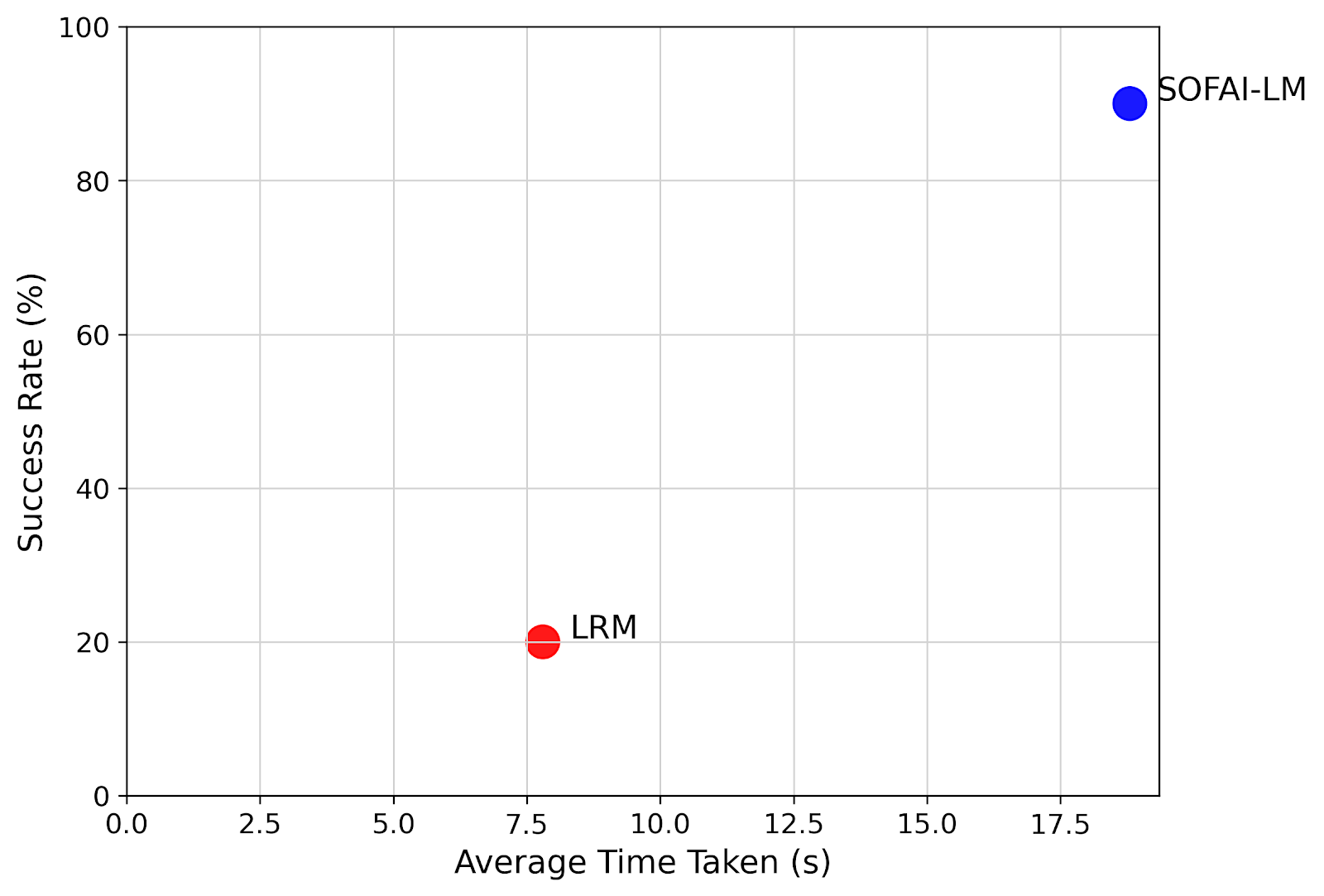}
    \caption{RQ4 - Combination 4: SOFAI-LM (Llama \(\rightarrow\) DeepSeek) vs. LRM (DeepSeek), Size 5}
    \label{fig:rq4_mc4_5}
\end{figure}

\begin{figure}[h!]
    \centering
    \includegraphics[width=0.95\linewidth]{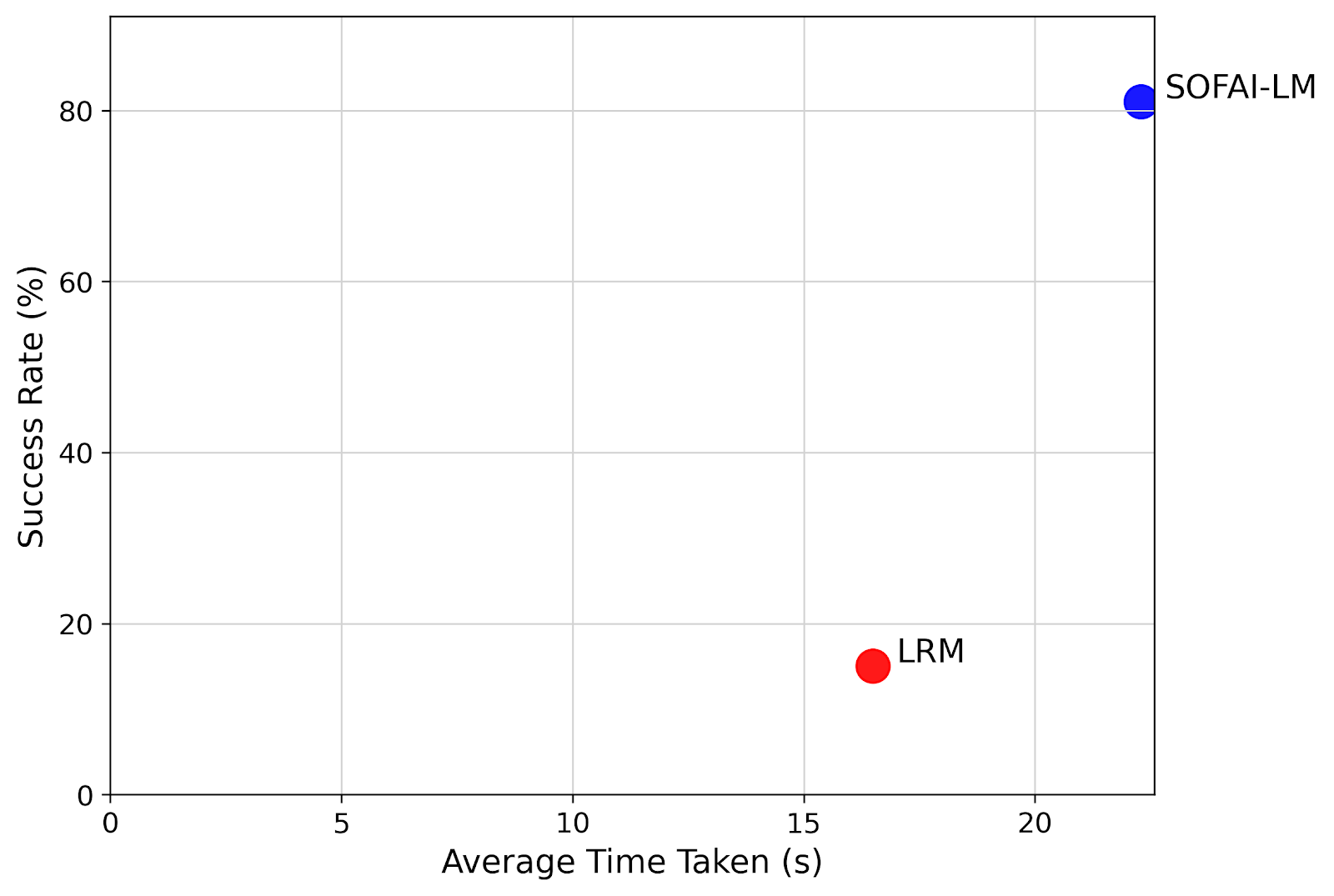}
    \caption{RQ4 - Combination 4: SOFAI-LM (Llama \(\rightarrow\) DeepSeek) vs. LRM (DeepSeek), Size 10}
    \label{fig:rq4_mc4_10}
\end{figure}

\begin{figure}[h!]
    \centering
    \includegraphics[width=0.95\linewidth]{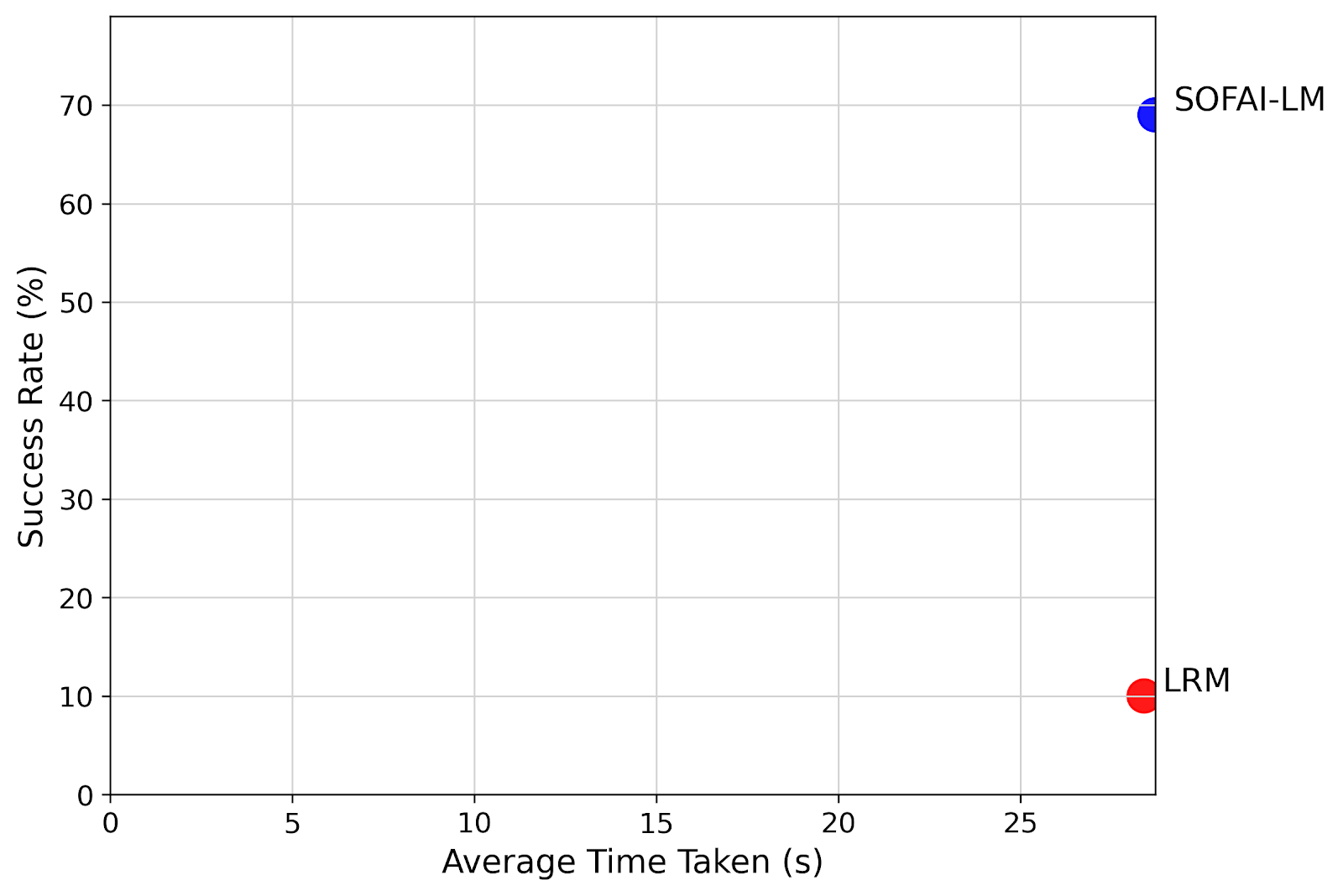}
    \caption{RQ4 - Combination 4: SOFAI-LM (Llama \(\rightarrow\) DeepSeek) vs. LRM (DeepSeek), Size 15}
    \label{fig:rq4_mc4_15}
\end{figure}

\begin{figure}[h!]
    \centering
    \includegraphics[width=0.95\linewidth]{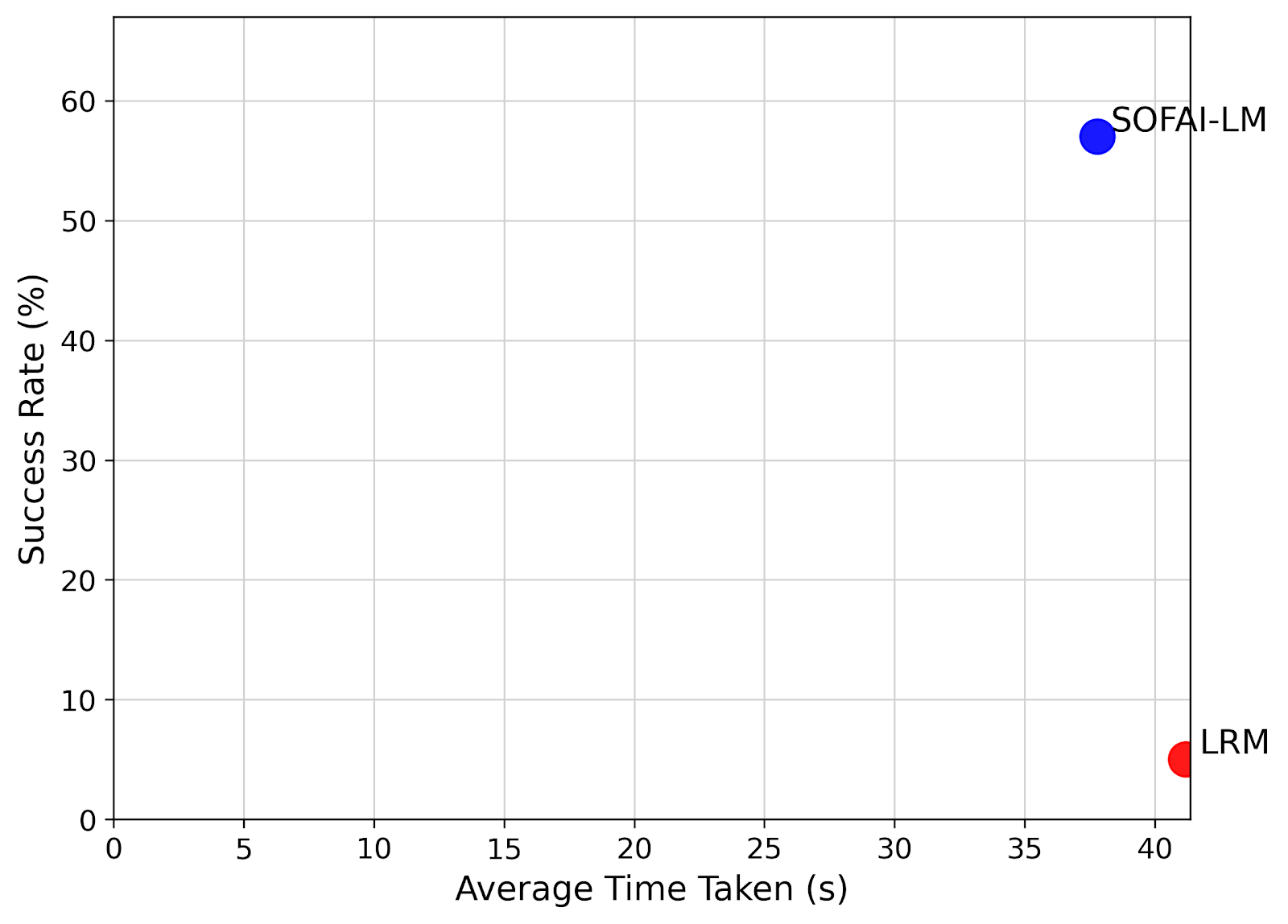}
    \caption{RQ4 - Combination 4: SOFAI-LM (Llama \(\rightarrow\) DeepSeek) vs. LRM (DeepSeek), Size 20}
    \label{fig:rq4_mc4_20}
\end{figure}

\begin{figure}[h!]
    \centering
    \includegraphics[width=0.95\linewidth]{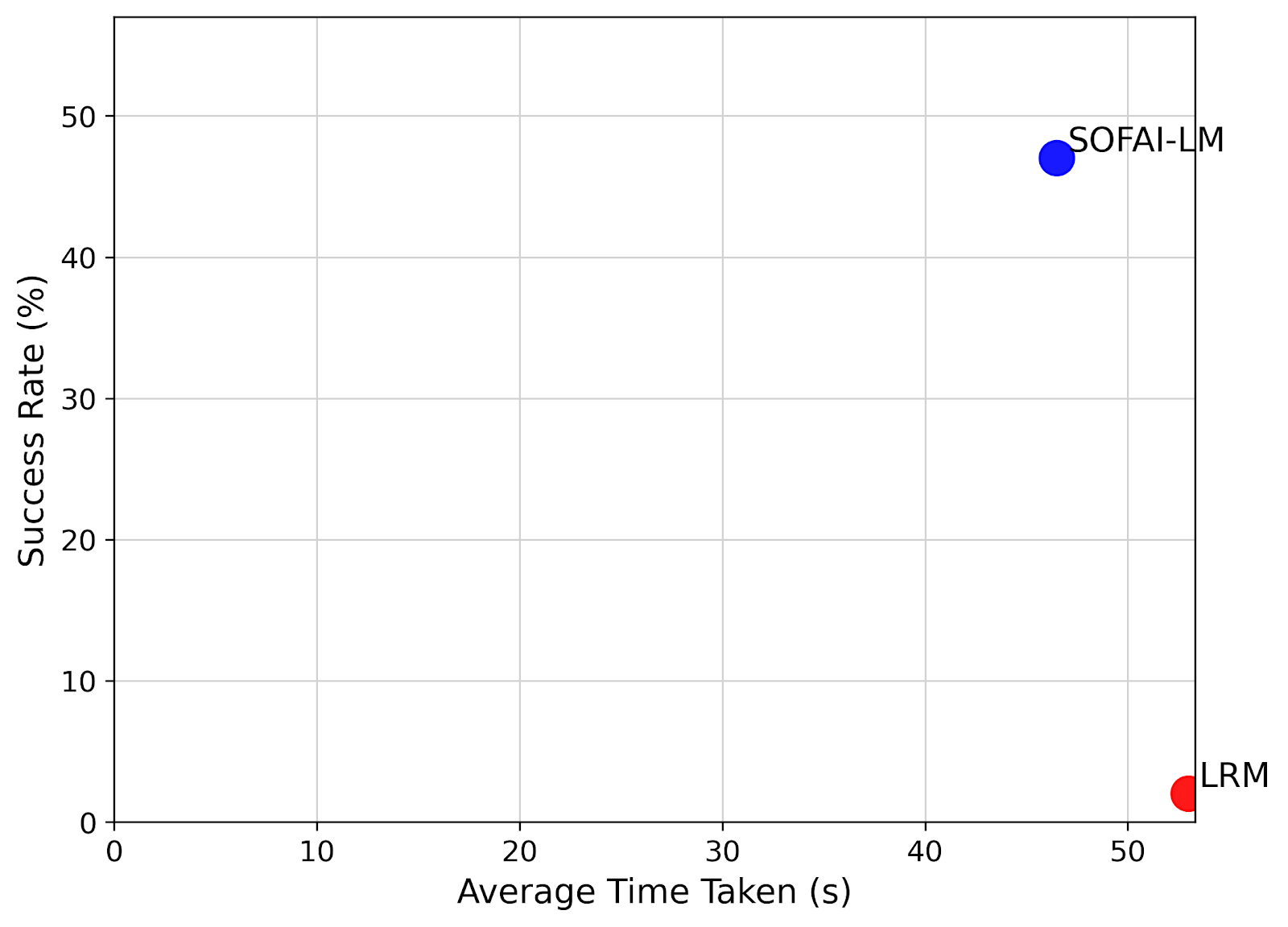}
    \caption{RQ4 - Combination 4: SOFAI-LM (Llama \(\rightarrow\) DeepSeek) vs. LRM (DeepSeek), Size 25}
    \label{fig:rq4_mc4_25}
\end{figure}

\paragraph{Description}
This final set of plots for RQ4 (Figures \ref{fig:rq4_mc4_5} through \ref{fig:rq4_mc4_25}) uses Llama 3.1 as the S1 solver and DeepSeek R1 8B as the S2 solver. The results compellingly demonstrate the power of the SOFAI-LM architecture. The integrated system achieves a vastly superior success rate compared to the standalone DeepSeek R1 8B LRM across all problem complexities. This confirms that the architectural benefits are not dependent on a specific S1 model.

\paragraph{Summary Finding}
The SOFAI-LM framework effectively leverages the strengths of a powerful LLM like Llama 3.1, leading to a system that is overwhelmingly more accurate and ultimately more efficient than its standalone LRM component.


\clearpage

\subsection{Experimental Results :  Code Debugging}

 We validate our core claims across several combinations of Large Language Models (LLMs) and Large Reasoning Models (LRMs) on the code debugging domain (DebugBench), demonstrating the robustness and generalizability of the SOFAI-LM architecture.

\subsection{RQ1: Can a feedback-driven LLM outperform an LRM?}
\label{subsec:supp_rq1}

The main paper demonstrated that a feedback-driven LLM (Granite 3.3 8B) could outperform a standalone LRM (DeepSeek R1 8B). Here, we confirm this finding across three additional model combinations.

\begin{figure}[h!]
    \centering
    \includegraphics[width=0.95\columnwidth]{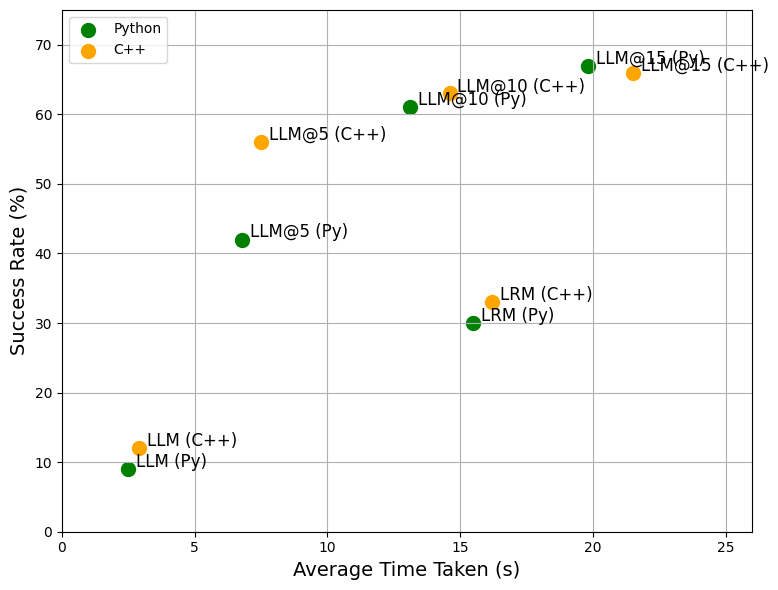}
    \caption{Comparison of success rate versus average time for a feedback-driven LLM and a standalone LRM on the DebugBench dataset (Python and C++). The configurations shown are a single-pass LLM, the LLM with 5, 10, and 15 feedback iterations (LLM@5, LLM@10, LLM@15), and the LRM. For this figure, the LLM (S1 solver) is Granite 3.3 8B (without thinking) and the LRM (S2 solver) is Granite 3.3 8B (with thinking).}
    \label{fig:rq1_c1}
\end{figure}

\paragraph{Combination 1: LLM = Granite 3.3 8B (without thinking), LRM = Granite 3.3 8B (with thinking).}
Figure~\ref{fig:rq1_c1} illustrates the performance trade-off when both the S1 and S2 solvers are from the same model family. The results clearly show that the feedback-driven LLM with 5 iterations (LLM@5) surpasses the performance of its more deliberative ``with thinking'' counterpart used as a standalone LRM. Increasing the number of iterations to 10 and 15 further widens this performance gap, establishing that the iterative feedback mechanism is more effective than relying on the LRM's built-in reasoning capabilities alone.

\begin{figure}[h!]
    \centering
    \includegraphics[width=0.95\columnwidth]{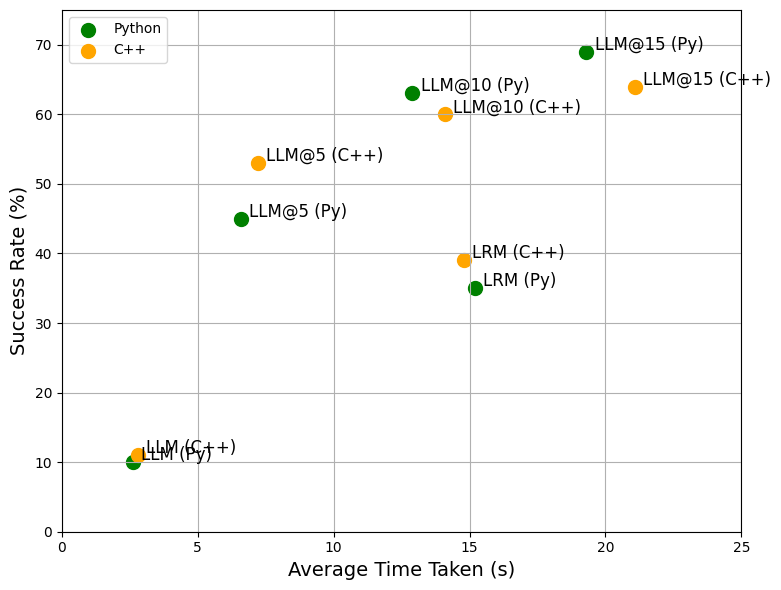}
    \caption{Comparison of success rate versus average time for a feedback-driven LLM and a standalone LRM on the DebugBench dataset (Python and C++ subsets). The configurations shown are a single-pass LLM, the LLM with 5, 10, and 15 feedback iterations (LLM@5, LLM@10, LLM@15), and the LRM. For this figure, the LLM (S1 solver) is Granite 3.3 8B (without thinking) and the LRM (S2 solver) is Qwen 3 8B.}
    \label{fig:rq1_c2}
\end{figure}

\paragraph{Combination 2: LLM = Granite 3.3 8B (without thinking), LRM = Qwen 3 8B.}
In Figure~\ref{fig:rq1_c2}, we pair the Granite 3.3 8B LLM with the Qwen 3 8B LRM. The trend remains consistent: the iterative LLM configurations (LLM@5, LLM@10, and LLM@15) form a clear Pareto frontier, significantly outperforming the standalone Qwen 3 8B LRM in both success rate and average time. This result underscores that the advantage of the SOFAI-LM feedback loop is not limited to a specific LRM but holds against other capable reasoning models.

\begin{figure}[h!]
    \centering
    \includegraphics[width=0.95\columnwidth]{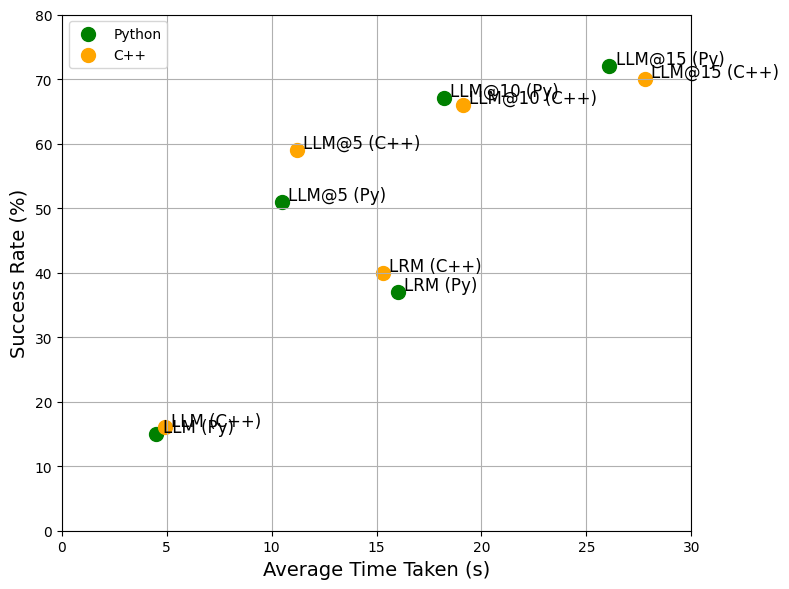}
    \caption{Comparison of success rate versus average time for a feedback-driven LLM and a standalone LRM on the DebugBench dataset (Python and C++ subsets). The configurations shown are a single-pass LLM, the LLM with 5, 10, and 15 feedback iterations (LLM@5, LLM@10, LLM@15), and the LRM. For this figure, the LLM (S1 solver) is Llama 3.1 8B and the LRM (S2 solver) is DeepSeek R1 8B.}
    \label{fig:rq1_c3}
\end{figure}
\paragraph{Combination 3: LLM = Llama 3.1 8B, LRM = DeepSeek R1 8B.}
Figure~\ref{fig:rq1_c3} validates our findings with a different base LLM, Llama 3.1 8B, against the powerful DeepSeek R1 8B LRM. Despite Llama 3.1 being slightly slower per iteration than Granite, the feedback-driven approach remains vastly superior. The LLM@5 configuration again proves more effective and efficient than the standalone LRM, confirming that the principles of our architecture are model-agnostic and apply to different state-of-the-art LLMs.

\clearpage

\subsection{RQ3: Can the information gathered by SOFAI-LM, when used iteratively with an LLM, enhance the performance of an LRM?}
\label{subsec:supp_rq3}

This section explores the impact of different LRM prompting strategies: Problem-Only (PO), Best Attempt (BA), and Full History (FH), within the SOFAI-LM pipeline for the code debugging domain.

\begin{figure}[h!]
    \centering
    \includegraphics[width=0.95\columnwidth]{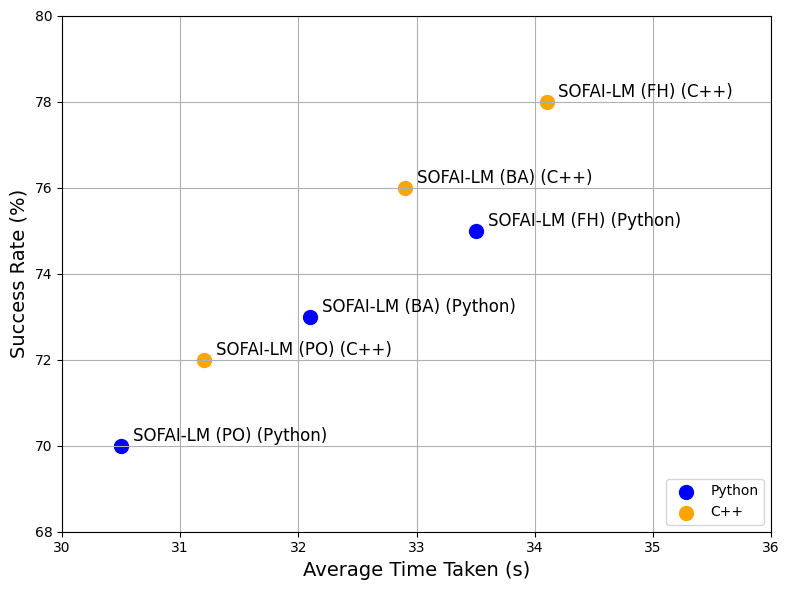}
    \caption{Comparison of SOFAI-LM performance using three different LRM prompting strategies: Problem-Only (PO), Best Attempt (BA), and Full History (FH). The plot shows success rate versus average time on the DebugBench dataset. For this figure, the LLM (S1 solver) is Granite 3.3 8B (without thinking) and the LRM (S2 solver) is Granite 3.3 8B (with thinking).}
    \label{fig:rq3_c1}
\end{figure}

\paragraph{Combination 1: LLM = Granite 3.3 8B (without thinking), LRM = Granite 3.3 8B (with thinking).}
As shown in Figure~\ref{fig:rq3_c1}, providing the LRM with more context from the LLM's attempts is beneficial. For both Python and C++, the success rate increases progressively from PO to BA and again to FH. This confirms that for code debugging, where fixes are often localized, the history of failed attempts provides valuable negative constraints that help the LRM converge on a correct solution.

\begin{figure}[h!]
    \centering
    \includegraphics[width=0.95\columnwidth]{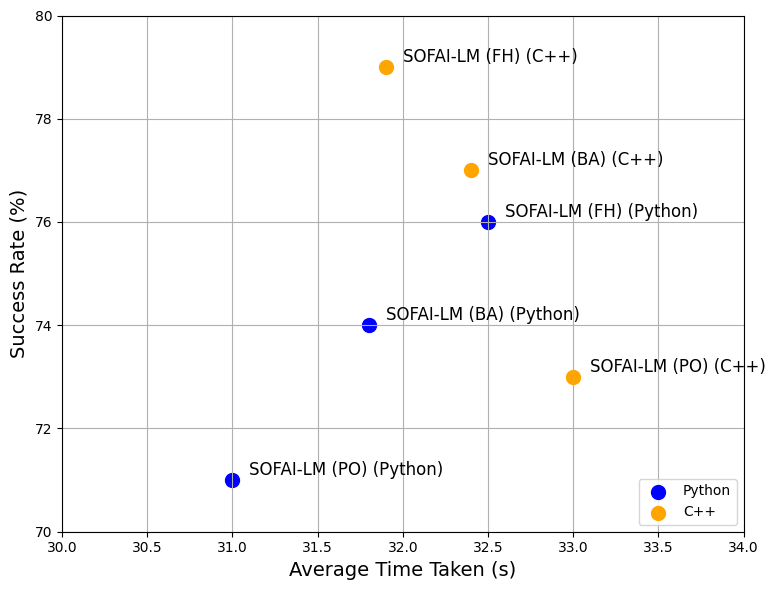}
    \caption{Comparison of SOFAI-LM performance using three different LRM prompting strategies: Problem-Only (PO), Best Attempt (BA), and Full History (FH). The plot shows success rate versus average time on the DebugBench dataset. For this figure, the LLM (S1 solver) is Granite 3.3 8B (without thinking) and the LRM (S2 solver) is Qwen 3 8B.}
    \label{fig:rq3_c2}
\end{figure}

\paragraph{Combination 2: LLM = Granite 3.3 8B (without thinking), LRM = Qwen 3 8B.}
The results in Figure~\ref{fig:rq3_c2} reinforce this finding with the Qwen 3 8B LRM. The Full History (FH) strategy yields the highest success rate. Notably, for the C++ subset, providing more context via BA and FH also slightly reduces the average inference time, suggesting the information helps the LRM find a solution more efficiently, not just more accurately.

\begin{figure}[h!]
    \centering
    \includegraphics[width=0.95\columnwidth]{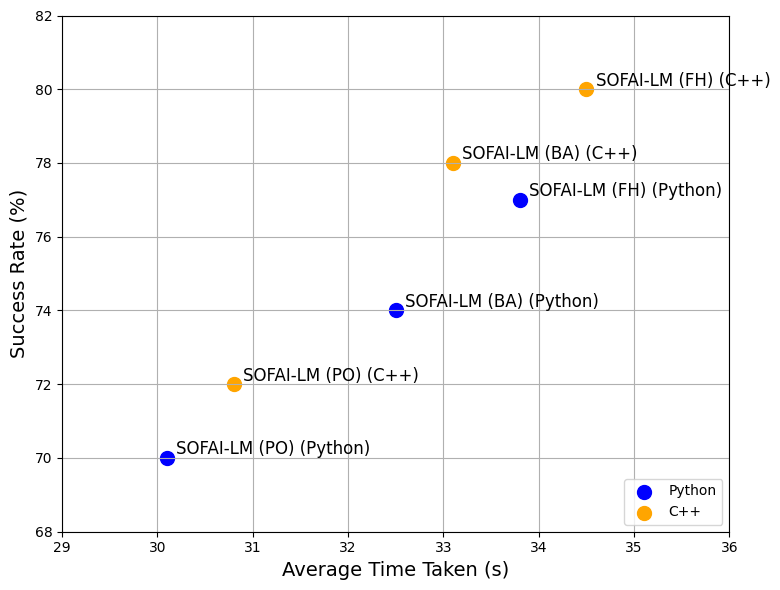}
    \caption{Comparison of SOFAI-LM performance using three different LRM prompting strategies: Problem-Only (PO), Best Attempt (BA), and Full History (FH). The plot shows success rate versus average time on the DebugBench dataset. For this figure, the LLM (S1 solver) is Llama 3.1 8B and the LRM (S2 solver) is DeepSeek R1 8B.}
    \label{fig:rq3_c3}
\end{figure}

\paragraph{Combination 3: LLM = Llama 3.1 8B, LRM = DeepSeek R1 8B.}
Figure~\ref{fig:rq3_c3} shows a distinct stepwise improvement in success rate as more context is provided to the DeepSeek R1 8B LRM. The hierarchy of $FH > BA > PO$ is clear, demonstrating that even when using a powerful LLM and LRM, enriching the LRM's prompt with the full interaction history is the optimal strategy for maximizing performance on code debugging tasks.

\clearpage

\subsection{RQ4: Does SOFAI-LM perform better than its LRM counterpart?}
\label{subsec:supp_rq4}

We present a direct comparison between the complete SOFAI-LM pipeline (using the best-performing FH strategy) and the standalone LRM for each model combination.

\begin{figure}[h!]
    \centering
    \includegraphics[width=0.95\columnwidth]{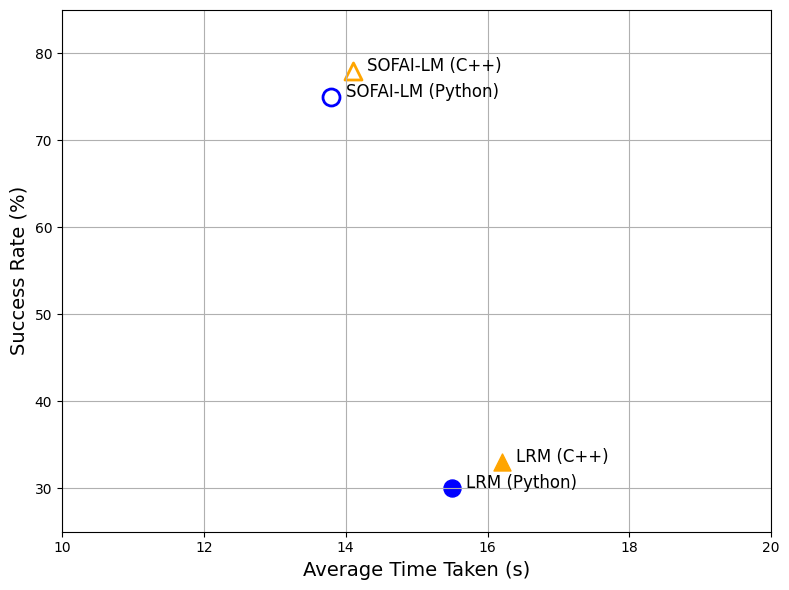}
    \caption{Overall performance comparison between the standalone LRM and the complete SOFAI-LM architecture on the DebugBench dataset. The plot shows success rate versus average time. For this figure, the LLM (S1 solver) is Granite 3.3 8B (without thinking) and the LRM (S2 solver) is Granite 3.3 8B (with thinking).}
    \label{fig:rq4_c1}
\end{figure}

\paragraph{Combination 1: LLM = Granite 3.3 8B (without thinking), LRM = Granite 3.3 8B (with thinking).}
Figure~\ref{fig:rq4_c1} shows the dramatic overall improvement provided by the SOFAI-LM architecture. The pipeline achieves a success rate more than double that of the standalone LRM while also being faster on average. This highlights the synergistic benefit of the architecture, which is far superior to using either of its components in isolation.

\begin{figure}[h!]
    \centering
    \includegraphics[width=0.95\columnwidth]{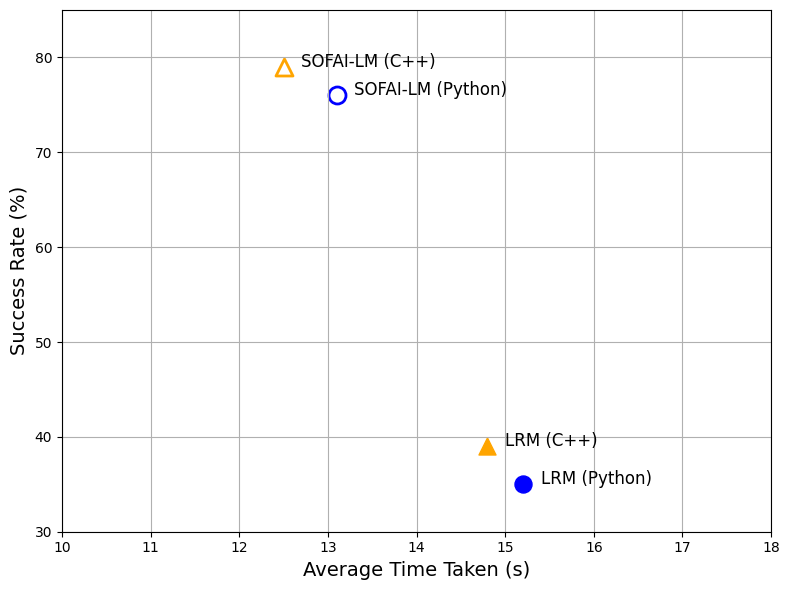}
    \caption{Overall performance comparison between the standalone LRM and the complete SOFAI-LM architecture on the DebugBench dataset. The plot shows success rate versus average time. For this figure, the LLM (S1 solver) is Granite 3.3 8B (without thinking) and the LRM (S2 solver) is Qwen 3 8B.}
    \label{fig:rq4_c2}
\end{figure}

\paragraph{Combination 2: LLM = Granite 3.3 8B (without thinking), LRM = Qwen 3 8B.}
The comparison in Figure~\ref{fig:rq4_c2} confirms the superiority of the SOFAI-LM framework. The integrated system significantly outperforms the standalone Qwen 3 8B LRM, achieving a much higher success rate in less time. This demonstrates the framework's ability to effectively orchestrate different models to achieve a result better than the sum of its parts.

\begin{figure}[h!]
    \centering
    \includegraphics[width=0.95\columnwidth]{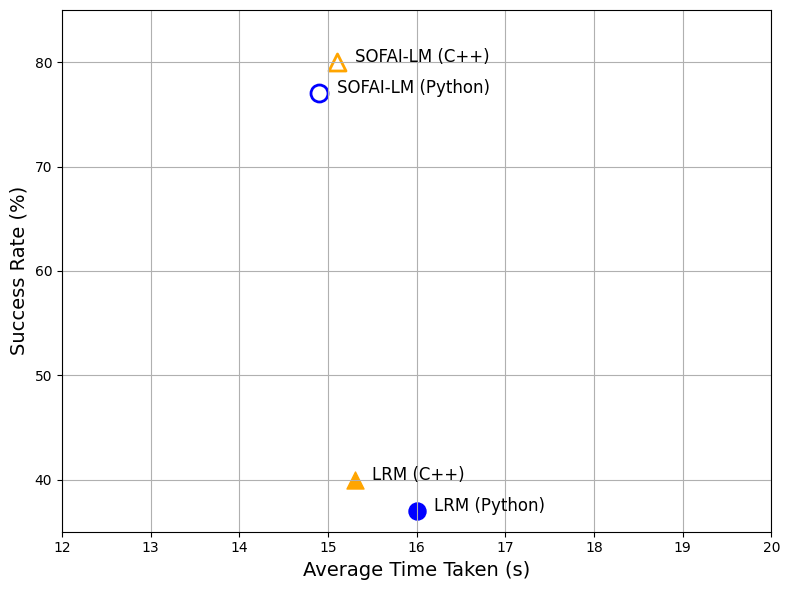}
    \caption{Overall performance comparison between the a standalone LRM and the complete SOFAI-LM architecture on the DebugBench dataset. The plot shows success rate versus average time. For this figure, the LLM (S1 solver) is Llama 3.1 8B and the LRM (S2 solver) is DeepSeek R1 8B.}
    \label{fig:rq4_c3}
\end{figure}

\paragraph{Combination 3: LLM = Llama 3.1 8B, LRM = DeepSeek R1 8B.}
As seen in Figure~\ref{fig:rq4_c3}, the SOFAI-LM architecture's advantage holds even when using a stronger pair of models. The pipeline, which intelligently leverages the fast Llama 3.1 8B LLM and falls back to the powerful DeepSeek R1 8B LRM only when necessary, again doubles the success rate of the standalone LRM while maintaining a slight edge in time efficiency. This result robustly confirms the value of our metacognitive approach.

\end{document}